\documentclass{article} %
\usepackage{main}

\usepackage{booktabs}
\usepackage{graphicx}
\usepackage{enumitem}
\usepackage{wrapfig}
\usepackage{wrapfig}
\usepackage{float}
\usepackage{microtype}
\usepackage{amsmath}
\usepackage{amssymb}
\usepackage{colortbl}
\usepackage[utf8]{inputenc}
\definecolor{lightgray}{rgb}{0.9,0.9,0.9}
\usepackage{caption}
\usepackage{subcaption}
\usepackage{xcolor}
\usepackage{setspace}
\usepackage{url}
\usepackage{multirow}
\usepackage{colortbl}
\usepackage{tabularx}
\usepackage{blindtext}
\usepackage{pgfplots}
\pgfplotsset{compat=1.18} 
\usepackage{tikz}
\usetikzlibrary{er,positioning,bayesnet}
\usepackage{makecell}
\usepackage{siunitx}
\usepackage{nicefrac}
\usepackage{tocloft}
\usepackage{listings}
\usepackage[raster,skins,breakable]{tcolorbox} 
\usepackage{xltabular}
\usepackage{adjustbox}
\usepackage{xurl}

\makeatother

\usepackage{xcolor}
\makeatletter
\@ifpackageloaded{tcolorbox}{}{\PassOptionsToPackage{most}{tcolorbox}}
\makeatother
\usepackage{tcolorbox}
\tcbuselibrary{most}

\definecolor{yuangreen}{HTML}{00CC70}   

\usepackage{xspace}

\newcommand{\method}{\textsc{AdaTutoRanker}\xspace}

\usepackage{dashrule}

\newcommand{\ie}{\textit{i.e., }}
\newcommand{\eg}{\textit{e.g., }}

\usepackage{pifont}

\usepackage{booktabs,soul}
\definecolor{hlblue}{RGB}{198,224,250}
\definecolor{hlgreen}{RGB}{204,236,204}
\definecolor{hlred}{RGB}{252,210,210}
\newcommand{\hlb}[1]{{\sethlcolor{hlblue}\hl{#1}}}
\newcommand{\hlok}[1]{{\setlength{\fboxsep}{1.2pt}\colorbox{hlgreen}{#1}}}
\newcommand{\hlbad}[1]{{\setlength{\fboxsep}{1.2pt}\colorbox{hlred}{#1}}}
\newcommand{\hlbx}[1]{{\setlength{\fboxsep}{1.2pt}\colorbox{hlblue}{#1}}}

\definecolor{systembg}{HTML}{DAECED}
\definecolor{userbg}{HTML}{DAECED}
\definecolor{darkblue}{HTML}{00008B}
\colorlet{insert}{darkblue}

\newtcolorbox{sysuserbox}[1]{%
    colback=#1, colframe=#1,
    boxrule=0pt, arc=0pt,
    left=3pt, right=3pt, top=3pt, bottom=3pt,
    breakable, width=\linewidth,
    before skip=0pt, after skip=0pt}
\newcommand{\systemprompt}[1]{\noindent\begin{sysuserbox}{systembg}%
    \textbf{System:}~#1%
  \end{sysuserbox}}
\newcommand{\userprompt}[1]{\noindent\begin{sysuserbox}{userbg}%
    \textbf{User:}~#1%
  \end{sysuserbox}}
\newcommand{\prompttag}[1]{\hspace{0pt}\texttt{#1}}
\newcommand{\prompt}[2]{\subsection{#1}
    \vspace{-2.5ex}
    {\noindent\\
    \rule[0pt]{\linewidth}{0.7pt}\scriptsize #2
    \noindent\rule[0pt]{\linewidth}{0.6pt}\normalsize}\relax}

\definecolor{CustomImageBlue}{HTML}{4169E1}

\makeatletter
  \newcommand\figcaption{\def\@captype{figure}\caption}
  \newcommand\tabcaption{\def\@captype{table}\caption}
\makeatother

\newcommand{\dalgshifted}{\raisebox{0.5\depth}{$\downarrow$}}

\definecolor{backred}{RGB}{255, 190, 190}
\definecolor{backblue}{RGB}{210, 230, 250}
\definecolor{verylightgray}{gray}{0.95}
\definecolor{backorange}{RGB}{255, 230, 204}
\definecolor{backyellow_soft}{RGB}{255, 252, 51}

\definecolor{myred}{RGB}{192,0,0}
\definecolor{mygreen}{RGB}{88,142,49}

\definecolor{tong}{RGB}{152, 1, 0}
\definecolor{mygray}{gray}{.9}
\definecolor{mylinkcolor}{RGB}{115,194,251}

\usepackage[normalem]{ulem}  

\definecolor{iccvblue}{rgb}{0.21,0.49,0.74}
\definecolor{myblue}{HTML}{486ED7}
\definecolor{deepred}{RGB}{152, 1, 0}
\definecolor{grey1}{RGB}{96, 101, 102}

\definecolor{yuangreendark}{HTML}{009B56}  
\colorlet{citecolor}{yuangreen}       
\colorlet{linkcolor}{yuangreen}       
\colorlet{urlcolor}{yuangreen}        

\makeatletter
\@ifpackageloaded{hyperref}{}{\PassOptionsToPackage{pagebackref}{hyperref}}
\makeatother
\usepackage{hyperref}
\hypersetup{
    breaklinks=true,
    bookmarks=false,
    colorlinks=true,
    allcolors=yuangreen,   
    linkcolor=linkcolor,
    urlcolor=urlcolor,
    citecolor=citecolor,
}

\usepackage{marvosym}

\tcbset{
    challenges/.style={
    colback=gray!5!white,
        colframe=white,
        boxrule=0.5mm,
        left=1mm,
        right=1mm,
        top=1mm,
        bottom=1mm,
        arc=3mm,
        boxsep=3mm,
        drop shadow={black!50!white},
        enhanced,
        overlay={
            \node[fill=cyan!70!black, text=white, rounded corners=1mm, font=\bfseries\scriptsize, inner sep=1mm]
            at (frame.north west)
            [xshift=8mm, yshift=-0mm] {Challenges};
            }
    }
}

\tcbset{
    insights/.style={
    colback=gray!5!white,
        colframe=white,
        boxrule=0.5mm,
        left=1mm,
        right=1mm,
        top=1mm,
        bottom=1mm,
        arc=3mm,
        boxsep=3mm,
        drop shadow={black!50!white},
        enhanced,
        overlay={
            \node[fill=orange!70!black, text=white, rounded corners=1mm, font=\bfseries\scriptsize, inner sep=1mm]
            at (frame.north west)
            [xshift=8mm, yshift=-0mm] {Insights};
            }
    }
}

\usepackage{graphicx}
\usepackage{caption}
\usepackage{subcaption}

\usepackage{epsfig}
\usepackage{inconsolata}
\usepackage{microtype}
\usepackage{physics}   
\usepackage{bbm}
\usepackage{bm}
\usepackage{adjustbox}
\usepackage{multirow}

\usepackage{tikz}
\usetikzlibrary{arrows}

\usepackage{listings}
\usepackage{esdiff}
\usepackage{lastpage}
\usepackage{pgf}

\usepackage{tabularx}
\usepackage{array}
\newcolumntype{H}{>{\setbox0=\hbox\bgroup}c<{\egroup}@{}}
\usepackage{colortbl}
\usepackage{float}
\usepackage{wrapfig}
\usepackage{hhline}
\usepackage{numprint}
\usepackage{makecell}

\usepackage{amsmath,amsfonts,amsthm,amssymb}
\usepackage{mathtools}
\usepackage{nicefrac}
\usepackage{dsfont}

\makeatletter
\@ifpackageloaded{algorithm2e}{}{\PassOptionsToPackage{ruled,linesnumbered}{algorithm2e}}
\makeatother
\usepackage{algorithm2e}

\SetCommentSty{mycommfont}
\SetAlgoNoLine

\SetNlSty{myalgnl}{}{:}
\SetNlSkip{0.6em}
\providecommand{\sg}{\mathrm{sg}}
\SetKwInOut{Require}{Require}
\SetKwInOut{Ensure}{Ensure}

\usepackage{enumitem}
\setlist{nosep}

\usepackage{tocloft}
\usepackage{bbding}   
\usepackage{titlesec}

\usepackage[page]{appendix}
\usepackage[hang,flushmargin]{footmisc}

\newtcbox{\hlredtabA}{on line, box align=base, colback=red!10, colframe=white, size=fbox, arc=3pt,
  before upper=\strut, top=-2pt, bottom=-4pt, left=-2pt, right=-2pt, boxrule=0pt}
\newtcbox{\hlredtabB}{on line, box align=base, colback=green!10, colframe=white, size=fbox, arc=3pt,
  before upper=\strut, top=-2pt, bottom=-4pt, left=-2pt, right=-2pt, boxrule=0pt}
\newtcbox{\hlredtabC}{on line, box align=base, colback=yellow!50, colframe=white, size=fbox, arc=3pt,
  before upper=\strut, top=-2pt, bottom=-4pt, left=-2pt, right=-2pt, boxrule=0pt}
\newtcbox{\hlredtabDA}{on line, box align=base, colback=red!10, colframe=white, size=fbox, arc=3pt,
  before upper=\strut, top=-2pt, bottom=-4pt, left=-2pt, right=-2pt, boxrule=0pt} 
\newtcbox{\hlredtabDB}{on line, box align=base, colback=red!20, colframe=white, size=fbox, arc=3pt,
  before upper=\strut, top=-2pt, bottom=-4pt, left=-2pt, right=-2pt, boxrule=0pt} 
\newtcbox{\hlredtabDC}{on line, box align=base, colback=red!30, colframe=white, size=fbox, arc=3pt,
  before upper=\strut, top=-2pt, bottom=-4pt, left=-2pt, right=-2pt, boxrule=0pt} 
\newtcbox{\hlredtabDD}{on line, box align=base, colback=red!40, colframe=white, size=fbox, arc=3pt,
  before upper=\strut, top=-2pt, bottom=-4pt, left=-2pt, right=-2pt, boxrule=0pt} 
\newtcbox{\hlredtabDE}{on line, box align=base, colback=red!50, colframe=white, size=fbox, arc=3pt,
  before upper=\strut, top=-2pt, bottom=-4pt, left=-2pt, right=-2pt, boxrule=0pt} 
\newtcbox{\hlredtabDF}{on line, box align=base, colback=red!60, colframe=white, size=fbox, arc=3pt,
  before upper=\strut, top=-2pt, bottom=-4pt, left=-2pt, right=-2pt, boxrule=0pt} 

\newcommand{\dabA}[1]{\hlredtabDA{\normalsize \dalgshifted{#1}}} 
\newcommand{\dabB}[1]{\hlredtabDB{\normalsize \dalgshifted{#1}}} 
\newcommand{\dabC}[1]{\hlredtabDC{\normalsize \dalgshifted{#1}}} 
\newcommand{\dabD}[1]{\hlredtabDD{\normalsize \dalgshifted{#1}}} 
\newcommand{\dabE}[1]{\hlredtabDE{\normalsize \dalgshifted{#1}}} 
\newcommand{\dabF}[1]{\hlredtabDF{\normalsize \dalgshifted{#1}}} 

\newcommand{\ColorValue}[1]{
    \pgfmathparse{#1}
    \ifdim\pgfmathresult pt>50pt
        \dabF{#1\%}
    \else
        \ifdim\pgfmathresult pt>40pt
            \dabE{#1\%}
        \else
            \ifdim\pgfmathresult pt>30pt
                \dabD{#1\%}
            \else
                \ifdim\pgfmathresult pt>20pt
                    \dabC{#1\%}
                \else
                    \ifdim\pgfmathresult pt>10pt
                        \dabB{#1\%}
                    \else
                        \ifdim\pgfmathresult pt>0pt
                            \dabA{#1\%}
                        \fi
                    \fi
                \fi
            \fi
        \fi
    \fi
}

\providecommand{\ph}[1]{\textcolor{blue}{\{#1\}}}
\definecolor{promptline}{HTML}{6B2D8E}\definecolor{promptfill}{HTML}{FBF6FE}
\definecolor{promptcap}{HTML}{F1E4FA}
\definecolor{bpromptline}{HTML}{215DB0}\definecolor{bpromptfill}{HTML}{F7FAFF}
\definecolor{bpromptcap}{HTML}{E6EEF9}
\definecolor{hintfill}{HTML}{FFFCF4}\definecolor{hintline}{HTML}{D79A2B}
\definecolor{hinttext}{HTML}{9C5A00}

\newcommand{\hintlabel}[1]{\textcolor{hinttext}{\textbf{{[}#1{]}}}}
\newcommand{\hintsep}{\par\addvspace{4.5pt}\nointerlineskip
  \hbox to \linewidth{\leaders\hbox{\rule{2.5pt}{0.4pt}\hskip2.5pt}\hfil}%
  \nointerlineskip\addvspace{4.2pt}}
\newcommand{\hintsample}[1]{\par\vspace{2.5pt}%
  \begingroup\ttfamily\scriptsize\leftskip=1.2em\noindent #1\par\endgroup\vspace{2.5pt}}

\tcbset{promptstyle/.style={
  enhanced, unbreakable,
  fonttitle=\bfseries\small,
  boxrule=0.8pt, titlerule=0.8pt, arc=2pt, boxsep=1pt,
  left=8pt, right=8pt, top=5pt, bottom=5pt,
  toptitle=2.5pt, bottomtitle=2.5pt, lefttitle=6pt,
  before skip=4pt, after skip=4pt,
  fuzzy shadow={1.4mm}{-1.4mm}{0mm}{0.4mm}{black!32}}}

\newtcolorbox{promptbox}[2][]{promptstyle,
  colback=promptfill, colframe=promptline,
  colbacktitle=promptcap, coltitle=promptline, title={#2}, #1}

\newtcolorbox{bpromptbox}[2][]{promptstyle,
  colback=bpromptfill, colframe=bpromptline,
  colbacktitle=bpromptcap, coltitle=bpromptline, title={#2}, #1}

\newtcolorbox{hintbox}[1][]{enhanced, nobeforeafter, unbreakable,
  colback=hintfill, colframe=hintfill, borderline west={2.2pt}{0pt}{hintline},
  boxrule=0pt, arc=1pt, boxsep=1pt, left=6pt, right=6pt, top=4pt, bottom=4pt,
  before upper={\setlength{\parindent}{0pt}\setlength{\parskip}{0pt}}, #1}

\newcommand{\studentteacherprompts}[2]{%
\begin{figure}[t]
\centering
\begin{promptbox}{Student Prompt}
\rmfamily\footnotesize
\textbf{System prompt.} You are a high-quality document set selector for a deep-research agent.
\ph{task\_description}; \ph{meta\_rubrics}; \ph{output\_format}.\\[4pt]
\textbf{User prompt.} I will provide you with \ph{num\_documents} documents, each indicated by a
numerical identifier {[}{]}. Select and rank the documents based on their joint usefulness for the
search query: \ph{query}; \ph{query\_intent}; \ph{candidate\_documents}.\\[4pt]
\textbf{Documents selection:}
\end{promptbox}
\begin{bpromptbox}{Teacher Prompt}
\rmfamily\footnotesize
\textbf{System prompt \& User prompt.} Identical to the student prompt above.\\[4pt]
\noindent\textbf{Adaptive Tutoring Hints.}\;
One of the three, adaptively matched to the reward of the rollout.\\[3pt]
\begin{hintbox}
\hintlabel{Rubrics as hint}\; Below are the scoring dimensions and key points that the optimal
document subset for this query should cover (a higher weight means more important): \ph{rubric}
\hintsample{{[}Doc-level rubrics{]}\\
- Relevance (weight 5): Does each document give the main role and archetype of the specific
champions listed in the query, rather than discussing general class systems or champion lore?\\
\textcolor{black!70}{\boldmath$\vdots$}\quad
\raisebox{0.5ex}{\textcolor{black!70}{more rubrics across the other dimensions}}}
\noindent Strictly follow the rubrics above and select the optimal document subset.\hintsep
\noindent\hintlabel{Corrective reflection as hint}\; Your current selection is not yet optimal.
Below is a concrete corrective idea for revising it toward the optimal document subset:
\ph{corrective\_reflection}
\hintsample{{[}2{]} and {[}22{]} should be included: {[}2{]} gives the structured class system and
{[}22{]} the specific subclass assignments. {[}1{]} {[}5{]} {[}7{]} {[}19{]} {[}32{]} are less
effective, as they contain unrelated content, repeat one another, or do not directly cover many
champions.}
\noindent Apply the corrective idea above and select the optimal document subset.\hintsep
\noindent\hintlabel{Sibling-set as hint}\; Below is an example optimal document subset for this
query (scored \ph{reward} out of 10 by the judge); use it as a concrete reference:
\ph{sibling\_set}
\hintsample{{[}2{]} {[}22{]}}
\noindent Learn from the reference example above and select the optimal document subset.
\end{hintbox}
\par\vspace{4pt}
\textbf{Documents selection:}
\end{bpromptbox}
\vspace{3pt}
\caption{#1}
\label{#2}
\vspace{-8pt}
\end{figure}}

\usepackage{graphicx}

\usepackage{amsmath, amssymb}
\usepackage{booktabs}
\usepackage{multirow}

\usepackage{pifont} 

\usepackage{bbm}
\usepackage{tikz}
\newlength{\barwidth}
\usepackage{lipsum}

\newcommand\blfootnote[1]{%
  \begingroup
  \renewcommand\thefootnote{}\footnote{#1}%
  \addtocounter{footnote}{-1}%
  \endgroup
}

\usepackage{CJKutf8}

\title{AdaTutoRank: Learning to Rerank Document Sets via Adaptive Tutoring Optimization for RAG and Deep Research}

\author{
Kailin Jiang\textsuperscript{1,{\color{yuangreen}*}} \quad Jian Xi\textsuperscript{1} \quad Lei Liu\textsuperscript{2,{\color{yuangreen}\dag}}  \quad Yangqi Chen\textsuperscript{1} \quad Hui Xu\textsuperscript{1}\\[6pt]
Hongwei Zhao\textsuperscript{2} \quad Bin Li\textsuperscript{2} \quad Yu Lu\textsuperscript{1,{\color{yuangreen}\dag}} \quad Haibo Shi\textsuperscript{1}\\[6pt]
\textsuperscript{1}Yuanbao Team, Tencent \quad \textsuperscript{2}Independent Researcher
}

\begin{document}

\maketitle
\blfootnote{\textsuperscript{{\color{yuangreen}*}}Work done during internship at Tencent Yuanbao. \textsuperscript{{\color{yuangreen}\dag}}Corresponding author: liulei13@ustc.edu.cn.}

\begin{abstract}
Document rerankers determine what evidence reaches the downstream model in RAG and deep research, yet mainstream rerankers select by relevance matching, and individually relevant documents rarely constitute the complete, complementary, non-redundant set a complex information need demands. Prior work rewards a set by its aggregate rubric score, shifting the objective from ranking documents to composing sets. Yet that score is one scalar shared by every document in the set, so the supervision is sparse: a redundant document is rewarded with the rest whenever the set scores well, and a decisive one penalized with the rest whenever it does not; credit assignment leaves \textit{contributors} indistinguishable from \textit{free riders}. On-policy distillation could densify this supervision, but existing methods give every rollout the same fixed guidance, too prescriptive for strong rollouts and too abstract for weak ones. We therefore propose \textbf{AdaTutoRank}, a setwise reranker trained with \textbf{Ada}ptive \textbf{Tuto}ring Optimization (ATO) under a three-level hierarchy of nine rubric dimensions, which supplies silver labels for the cold start, rewards for reinforcement learning, and hints for distillation. ATO draws three hint forms of increasing specificity from the policy's own frozen snapshot: the rubrics alone, a \textit{self-selector}'s sibling-set chosen under rubrics, and a \textit{self-reflector}'s reflection contrasting the rollout with that sibling-set; each rollout receives the form matched to its quality. Re-scoring that rollout under the hint-conditioned frozen teacher and the hint-free snapshot distills the hint's effect into a token-level advantage that complements the group-relative outcome advantage. Across ten benchmarks spanning RAG, deep research, and setwise evaluation, AdaTutoRank attains the best overall performance while issuing fewer retrieval calls.

\vspace{0.5em}
\noindent\raisebox{-0.3em}{\includegraphics[height=1em]{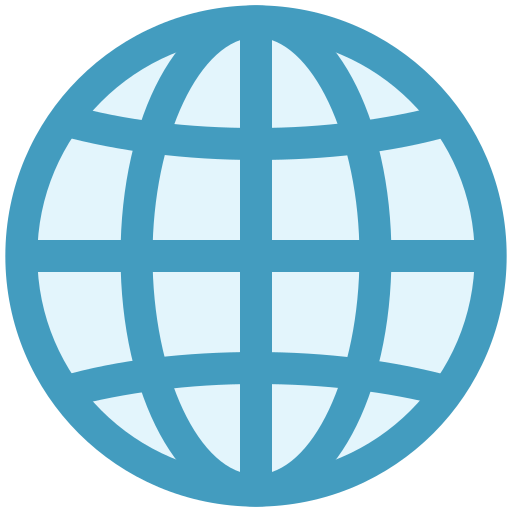}}~~\textbf{Project Page:} \href{https://adatutorank.github.io}{\texttt{https://adatutorank.github.io}}\\[0.3em]
\noindent\raisebox{-0.3em}{\includegraphics[height=1em]{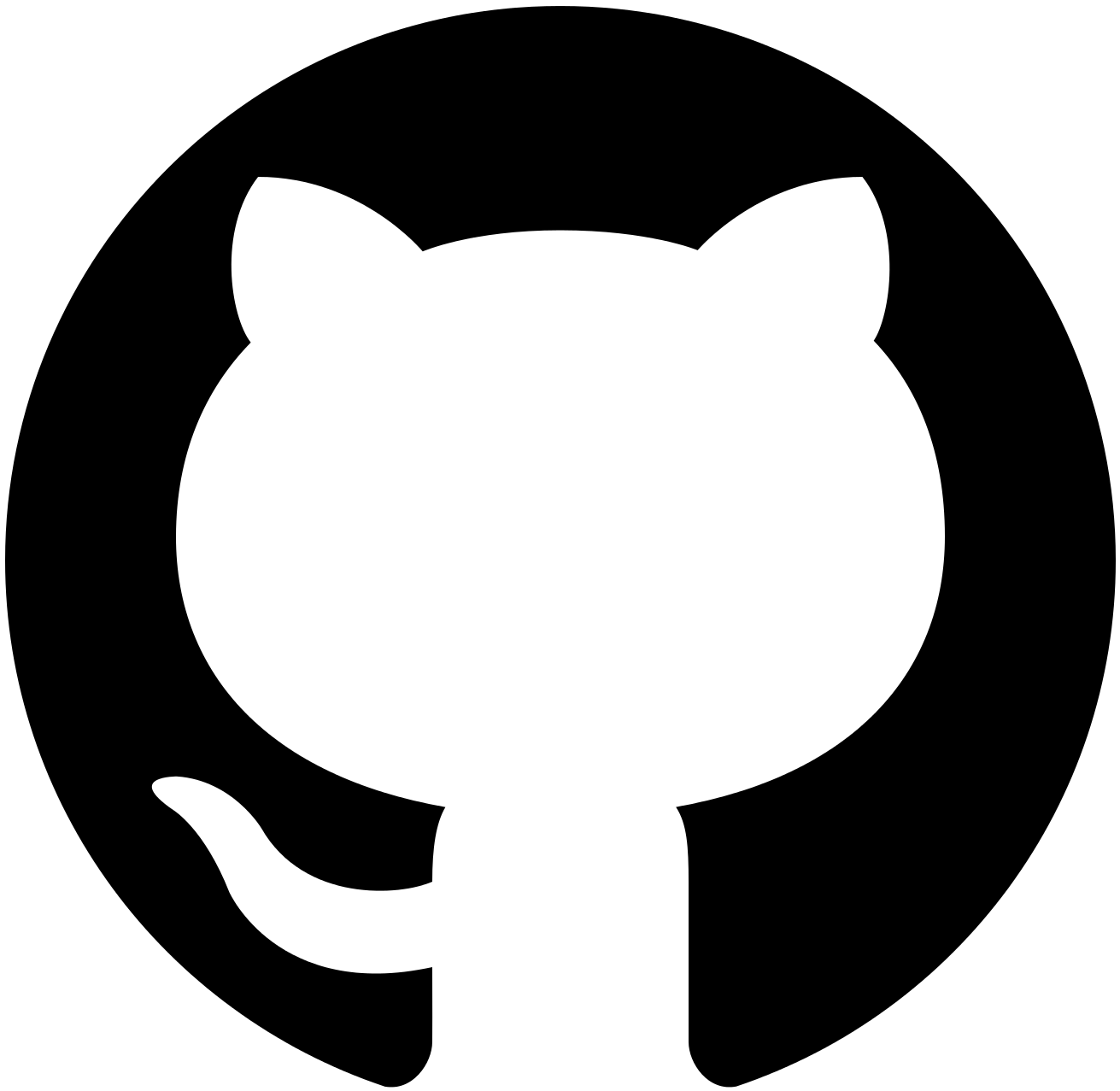}}~~\textbf{GitHub Repo:} \href{https://github.com/AdaTutoRank/AdaTutoRank}{\texttt{https://github.com/AdaTutoRank/AdaTutoRank}}\\[0.3em]
\noindent\raisebox{-0.3em}{\includegraphics[height=1em]{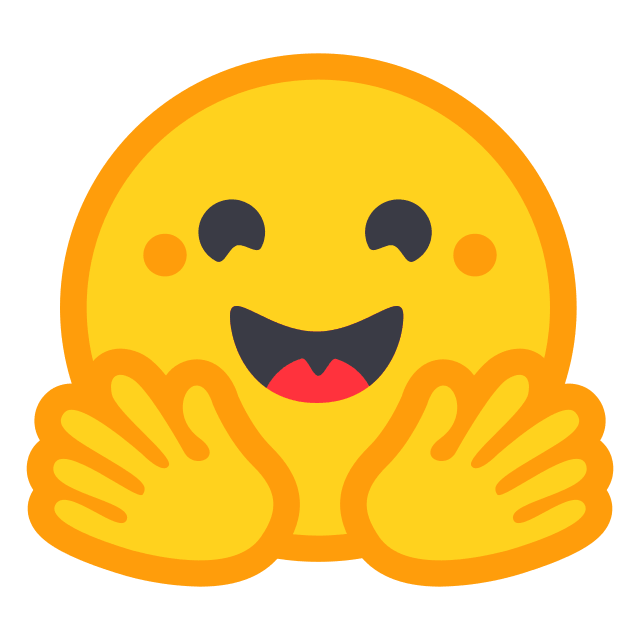}}~~\textbf{Dataset:} \href{https://huggingface.co/datasets/kailinjiang/AdaTutoRank-Train-Data}{\texttt{https://huggingface.co/datasets/kailinjiang/AdaTutoRank-Train-Data}}\\[0.3em]
\noindent\raisebox{-0.3em}{\includegraphics[height=1em]{logo/icon/hf.png}}~~\textbf{Model:} \href{https://huggingface.co/kailinjiang/AdaTutoRank-8B}{\texttt{https://huggingface.co/kailinjiang/AdaTutoRank-8B}}
\end{abstract}

\section{Introduction}\label{sec:intro}

Document rerankers determine the quality of the evidence passed to the downstream LLM in RAG and deep research, and thus the accuracy of the generated answer~\citep{rag_survey,Liu2023LostIT,Yoran2023MakingRL,tongyideepresearch,deepresearch_survey,deepresearch_survey1}. A deep research agent further interacts with the search system over multiple steps, reasoning about its current information need, issuing a sub-query, observing the returned documents, and deciding whether to keep searching before producing a long-form answer~\citep{react,search-r1,Dr-tulu,qi2025context,jiang2025mmke,jia2026benchmarking}. Retrieval quality therefore governs not only the current observation but also the reasoning and search decisions that follow, so a single poor observation propagates and compounds over the trajectory.

\begin{figure*}[t!]
  \centering
\includegraphics[width=1\linewidth]{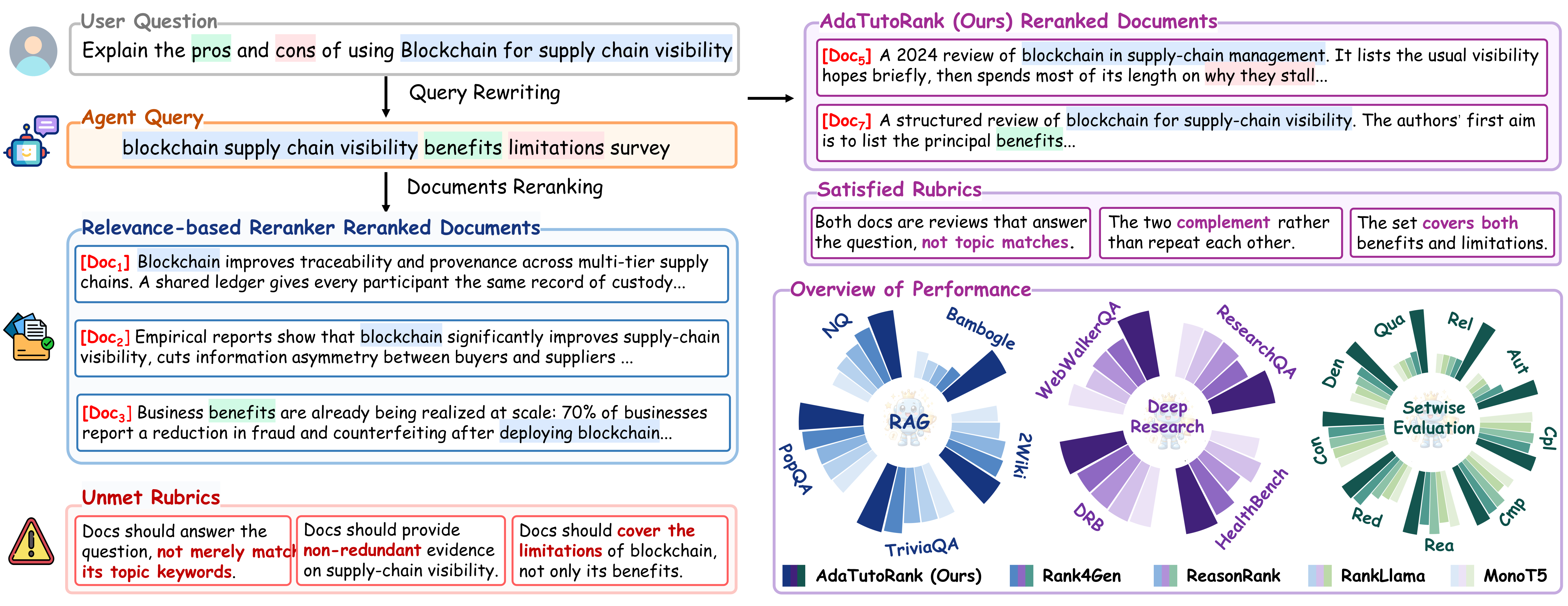}
  \caption{\textbf{(i)} Relevance-based reranking returns individually on-topic documents whose set is superficially relevant, redundant, and incomplete. \textbf{(ii)} AdaTutoRank instead selects a set that is genuinely relevant, complementary, and comprehensive, while consistently performing best on RAG, deep research, and setwise evaluation.}
  \label{fig:teaser}
\end{figure*}

Mainstream rerankers still ground their supervision in relevance annotations and return the top-$k$ individually relevant documents~\citep{RankT5,BGE-Reranker,jiang2025kore,peng2025can}. Yet the downstream model needs not a collection of such documents, but a set that jointly supports answering the query, covering it comprehensively without redundancy or conflict. As shown in Figure~\ref{fig:teaser}, for the question ``Explain the pros and cons of using Blockchain for supply chain visibility'', a relevance-based reranker returns documents that are all on topic yet form a poor set. \ding{182} They merely match the topic keywords rather than answer the question. \ding{183} Doc$_1$ and Doc$_2$ restate the same point, consuming context budget without adding information. \ding{184} All three speak only to the benefits, leaving the limitations uncovered.

Recent work has shown the promise of rubrics for guiding document-set selection. RubricRanker~\citep{rubricranker} drives GRPO training with a rubric-based reward, while Rubric4Setwise employs rubrics as a training-free prompting signal~\citep{SetWiseEvalKit}. Instantiating a rubric as an RL reward, however, collapses the supervision into a set-level scalar spread uniformly over every document. Three problems follow. \ding{182} Document-level credit assignment is absent, as redundant documents free-riding on a high-reward set go unpunished, while decisive documents in a low-reward set are penalized with the rest. \ding{183} Without dense process supervision, the scalar invites reward hacking: selecting a single document trivially incurs zero redundancy and zero conflict, scoring highly on both. \ding{184} Its rubrics cover mainly relevance, conflict, and redundancy, leaving density, completeness, and complementarity unmeasured, which compounds the two problems above.

The first two problems both stem from the sparsity of the reward. On-policy distillation is well suited to remedying it, since rescoring the same rollout under a context augmented with privileged information yields dense supervision resolved to individual documents. Prior work has instantiated such information as environment feedback, reference solutions, demonstration trajectories, or transient experience, yet all share one limitation~\citep{SDPO,OPSD,SDFT,OPCD,jiang2026large,jiang2026mined}. Although its content varies across samples, its type is fixed for the entire training set, which teaches every sample with one teacher and one recipe and cannot adapt to rollouts that fail in different ways. Effective supervision over document sets should therefore satisfy three requirements.

First, the supervision should be \textit{\textbf{on-policy}}, because useful corrections depend on the failure modes the current policy exhibits, whereas fixed demonstrations, static reference sets, and one-time distillation cannot track a policy whose capability and outputs evolve. Second, it should be \textit{\textbf{dense}}, because a scalar reward says only whether the whole set is good, not which document to keep and which to drop. Third, it should be \textit{\textbf{adaptive tutoring}}, because rollouts differ in why they fail and how much correction they can absorb, so the privileged information should vary in form with rollout quality.

We propose \textbf{AdaTutoRank}, a setwise reranker trained under rubric supervision with \textit{Adaptive Tutoring Optimization} (ATO). We introduce rubrics organized as a three-level hierarchy of nine dimensions and carry them through the entire training process, supplying silver labels for supervised fine-tuning and both the scalar reward and the privileged information for ATO. Training proceeds in two stages. \textit{Setwise Supervised Fine-Tuning} fine-tunes the policy on those silver labels, giving a stable cold start that selects sets in the required form. \textit{Adaptive Tutoring Optimization} then supplies the dense signal that the scalar reward lacks. At each step the frozen policy generates a hint for every rollout, matched to its reward. \ding{182} A high-reward rollout receives the rubrics alone, encouraging closer conformance; \ding{183} a medium-reward one receives a reflection contrasting its own set with the rubric-selected set, correcting erroneous selections; and \ding{184} a low-reward one receives that sibling-set directly as a correction. Re-scoring a rollout under its hint turns the hint's effect into a token-level advantage, which is combined with the group-relative outcome advantage so that the quality of the set and that of each document within it are optimized together. Hints and query-specific rubrics are used only in training, and at inference the policy sees only the query, meta-rubric and candidates.

The contributions of this paper are summarized as follows:

\begin{itemize}[leftmargin=1.4em]
    \item We introduce rubrics organized as a three-level hierarchy of nine dimensions and carry them through the training pipeline, where they supply silver labels, reinforcement rewards, and distillation hints, making multi-dimensional set quality explicitly optimizable at every stage.
    \item We propose adaptive tutoring optimization, which matches the form of the hint to each rollout's quality and distills its effect into a token-level advantage combined with the group-relative outcome advantage, turning set-level utility into token-level credit.
    \item Across ten benchmarks spanning answer-level and setwise-level evaluation, AdaTutoRank achieves the best overall performance while reducing the agent's retrieval calls.
\end{itemize}

\section{Related Work}

\noindent \textbf{RAG and Deep Research.} Retrieval-augmented generation (RAG) lets LLMs draw on knowledge beyond their parameters~\citep{rag_survey,Deng2026DataCentricPO,liu2026proprietary,li2026knowhal,fu2026can}, yet a single retrieval round rarely suffices for complex information-seeking tasks. This has motivated agentic search~\citep{react} and deep research agents that interleave reasoning with web search for open-ended queries~\citep{deepresearch_survey1,tongyideepresearch}. Such agents change what a reranker serves: it is invoked once per self-issued sub-query rather than per user question, and its output becomes the observation conditioning the next reasoning step, so any defect in that set propagates through every step that follows.

\noindent \textbf{Document Reranking.} Rerankers have evolved from ordering documents toward composing sets. Ad hoc methods rank candidates by relevance with trained scorers~\citep{BGE-Reranker,MonoT5,RankLlama} or LLM prompting~\citep{RankVicuna,RankZephyr}, and reasoning-enhanced methods add explicit reasoning~\citep{Rearank,ReasonRank}. Both optimize pointwise ordering, leaving set composition to a fixed cutoff. Setwise methods instead predict the subset directly, supervised by answer-generation quality~\citep{Rank4Gen} or a strong teacher~\citep{SetR}, yet neither exists for open-ended queries with unverifiable answers. Rubrics fill this gap with structured criteria, yet RubricRanker~\citep{rubricranker} trains on a set's aggregate rubric score, a single set-level scalar from which the gradient cannot separate contributors from free riders.

Prior rubric-based training shares one scalar across the whole set, crediting no document individually. AdaTutoRank turns the rubrics into hints, matches their form to rollout quality, and distills each into token-level shaping over the identifier sequence.



\section{Methodology}
\label{sec:method}

We propose \textbf{AdaTutoRank}, a setwise reranker for RAG and deep research, trained under rubric supervision throughout the pipeline with \textbf{Ada}ptive \textbf{Tuto}ring Optimization (ATO). Its design follows two observations: a rubric-based reward is a single scalar spread uniformly over the set and thus carries no document-level credit; and one fixed type of privileged information cannot serve rollouts that fail for different reasons. ATO therefore supplements the sparse reward with adaptive tutoring hints, dense supervision whose form varies with rollout quality.

As shown in Figure~\ref{fig:method}, this yields two training stages. We first synthesize query-specific rubrics, use them to produce silver labels, and fine-tune the policy into a stable cold start. We then distill the hints into token-level supervision and combine it with group-relative outcome advantages.

\subsection{Preliminary}
\label{sec:preliminary}

Given a query $q$ and a candidate pool $\mathcal{C}=\{d_1,\ldots,d_m\}$ returned by a retriever, a reranker decides which documents serve as evidence for answering $q$. Most rerankers learn a scoring function $\sigma(q,d)$ measuring the relevance of $d$ to $q$, and truncate the ranking it induces over $\mathcal{C}$ at a fixed cutoff $k$, \ie $\mathcal{S}=\mathrm{Top}_k\big(\sigma(q,\cdot)\big)$, so a hyperparameter rather than the model determines what the evidence contains. We therefore adopt the setwise formulation, in which the subset itself is the prediction,
\begin{equation}
    \mathcal{S}^{\star}=\operatorname*{arg\,max}_{\mathcal{S}\subseteq\mathcal{C}}
    u(\mathcal{S}\mid q),
    \label{eq:setwise}
\end{equation}
where the set utility $u(\cdot\mid q)$ is non-additive over documents and $|\mathcal{S}^{\star}|$ is set by the information need of $q$ rather than by a hyperparameter. The reranker thereby takes on two decisions that ranking leaves out, \ding{182} \textit{which documents belong together} and \ding{183} \textit{how many are enough}.

Since $u$ is not directly observable, we instantiate it with the query-specific rubrics: a reward model scores $\mathcal{S}$ against each rubric in $R_q$ and aggregates them into the training-time value of $u(\mathcal{S}\mid q)$ (Eq.~\ref{eq:utility}). The reranker is an autoregressive policy that outputs the selected identifiers directly,
\begin{equation}
y\sim\pi_\theta(\cdot\mid q,\mathcal{C}),\qquad \mathcal{S}=\mathrm{Parse}(y),
\label{eq:policy}
\end{equation}
where $y$ is a bracketed identifier sequence such as \texttt{"[2] [5] [8]"}.

The formulation covers both scenarios, which differ only in what $q$ is and how often the policy runs. In RAG, it runs once on the user question, so $\mathcal{S}$ is the generator's entire evidence and $u(\mathcal{S}\mid q)$ bounds answer quality. In deep research, the agent~\citep{Dr-tulu} follows ReAct~\citep{react} and interleaves reasoning with search; the policy runs after every search action, with $q=q_t$ a self-issued sub-query and $\mathcal{S}_t\subseteq\mathcal{C}_t$ becoming that step's observation.

\subsection{Rubrics Construction}\label{sec:rubrics}

Following prior works~\citep{rubricranker,SetWiseEvalKit}, we construct query-specific rubrics specifying the properties a selected document set should satisfy at the doc, set, and global levels.

 \begin{figure*}[t!]
  \centering
\includegraphics[width=1\linewidth]{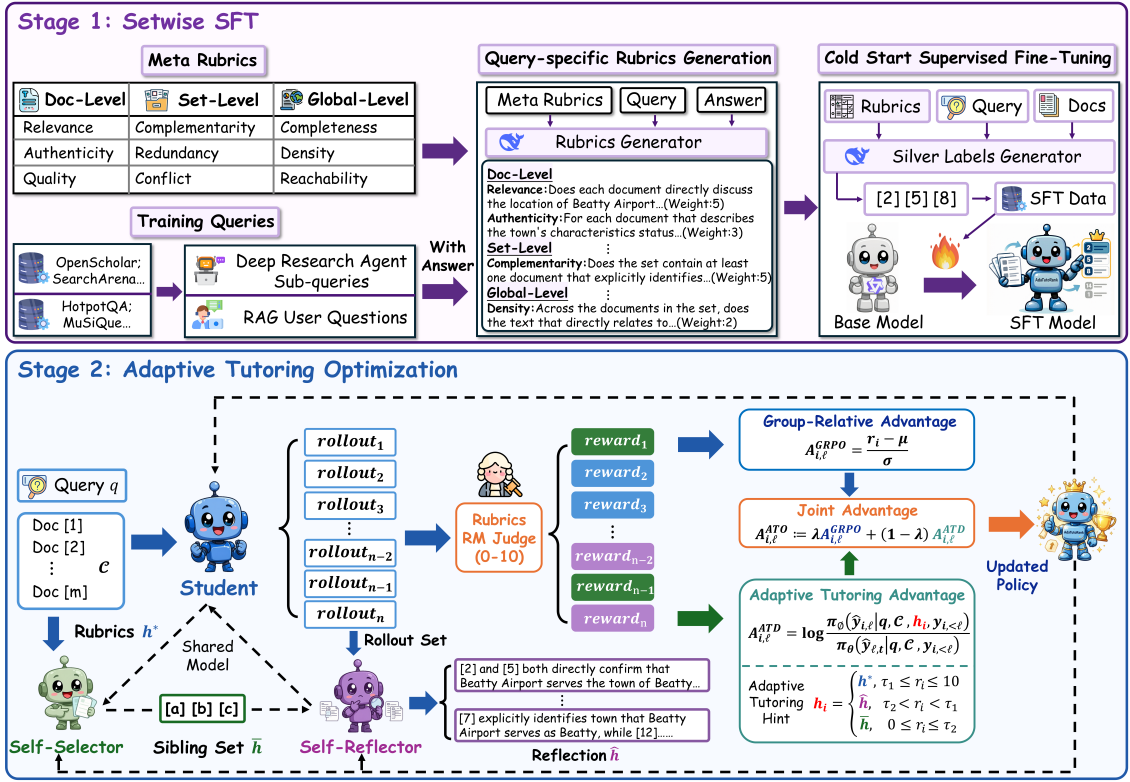}
  \caption{\textbf{Overview of AdaTutoRank.} \textbf{Stage 1} (Setwise SFT) generates silver labels from query-specific rubrics and fine-tunes the policy into a cold start for set selection. \textbf{Stage 2} (Adaptive Tutoring Optimization) routes to each rollout a hint matched to its reward, and re-scoring the rollout with and without that hint yields a dense advantage complementing the group-relative outcome signal.}
  \label{fig:method}
\end{figure*}

\noindent \textbf{Meta Rubrics Design.} Prompting an LLM for rubrics directly tends to yield limited coverage, repetitive items, or conflated dimensions. RubricRanker mitigates this with a fixed schema, but its five dimensions are too narrow to characterize a document set. We therefore adopt meta rubrics organized as a three-level hierarchy of nine dimensions. The document level focuses on each document in isolation, through \ding{182}~\textit{Relevance} {\small(Rel.)}, \ding{183}~\textit{Authenticity} {\small(Aut.)}, and \ding{184}~\textit{Quality} {\small(Qua.)}; the set level focuses on the synergy between documents, through \ding{185}~\textit{Complementarity} {\small(Cmp.)}, \ding{186}~\textit{Redundancy} {\small(Red.)}, and \ding{187}~\textit{Conflict} {\small(Con.)}; and the global level focuses on the set as LLM input context, through \ding{188}~\textit{Completeness} {\small(Cpl.)}, \ding{189}~\textit{Density} {\small(Den.)}, and \ding{190}~\textit{Reachability} {\small(Rea.)}, with details in Appendix~\ref{appendix:meta_rubrics}.

\noindent \textbf{Training Query Collection.} For RAG, we take short closed-ended questions from HotpotQA~\citep{HotpotQA}, NQ~\citep{nq}, 2WikiMultihopQA~\citep{2wikimultihopqa}, and MuSiQue~\citep{MuSiQue}. In deep research, a reranker serves agent-issued sub-queries, not user questions, which open-source datasets rarely release. We therefore reuse those from RubricRanker, covering OpenScholar~\citep{openscholar}, SearchArena~\citep{searcharena}, GlaiveAI-Reasoning-v1-20M, and WebWalker-Silver~\citep{webwalkerqa}, with details in Appendix~\ref{appendix:train_query}.

\noindent \textbf{Query-Specific Rubrics Generation.} We instantiate $R_q$ by prompting DeepSeek-V4 Pro with the meta rubrics, the query $q$, and its reference answer $a$, producing one or more rubrics per dimension. Each rubric is an evaluation question on set quality that must cite specific entities, facts, or values from $q$ and $a$; vague phrasing such as ``relevant content'' is prohibited.

\subsection{Setwise Supervised Fine-Tuning}\label{sec:sft}

The first stage cold-starts the policy to select document sets that jointly satisfy the multi-dimensional criteria in $R_q$, supervised by silver labels that a frontier LLM produces conditioned on $q$ and $R_q$.

\noindent \textbf{Silver Label Generation.} Since a base model rarely assembles a high-quality set on its own, its rollouts would impede the subsequent optimization stage; we thus cold-start the policy on silver labels from DeepSeek-V4-Pro. Conditioned on $q$, $\mathcal{C}$, and $R_q$, it emits $y^{*}$, the identifiers of the retained documents, with $|\mathcal{C}|$ sampled from $10$ to $40$ per query to expose it to varying pool sizes.

\noindent \textbf{Setwise SFT.} We then fine-tune the policy to predict $y^{*}$ from $q$ and $\mathcal{C}$ alone. $R_q$ is withheld, since it depends on a reference answer unavailable at inference, so training and inference see identical inputs. Each example $(q,\mathcal{C},y^{*})\in\mathcal{D}_{\mathrm{sft}}$ is optimized with the negative log-likelihood,
\begin{equation}
\mathcal{L}_{\mathrm{sft}}(\theta)=
-\mathbb{E}_{(q,\mathcal{C},y^{*})\sim\mathcal{D}_{\mathrm{sft}}}
\left[\sum_{\ell=1}^{|y^{*}|}
\log \pi_{\theta}\!\left(y^{*}_{\ell}\mid q,\mathcal{C},y^{*}_{<\ell}\right)\right],
\label{eq:sft}
\end{equation}
where $\ell$ indexes the tokens of $y^{*}$. The resulting $\theta_{\mathrm{sft}}$ initializes the next-stage policy and serves as its frozen distillation teacher (Eq.~\ref{eq:adv_atd}).

\subsection{Adaptive Tutoring Optimization}\label{sec:ato}

The second stage performs adaptive tutoring optimization. At each update, we freeze the current policy as $\pi_{\theta_{\mathrm{old}}}$, which both samples the rollout group $\{y_i\}_{i=1}^{G}$ and parameterizes the two hint generators of Figure~\ref{fig:method}, a \textit{self-selector} and a \textit{self-reflector}. Each rollout is then paired with the hint matched to its reward $r_i$, and $\pi_\theta$ is optimized jointly with the reinforcement learning and distillation objectives before becoming the next snapshot. Supervision therefore stays on-policy: hints derive from the policy's own rollouts, and their effect is distilled back into a policy that needs no hint at inference.

\noindent \textbf{Hierarchical Rubric-based Reward.} Given the query-specific rubrics $R_q = \{R^{\mathrm{doc}}_1, \ldots, R^{\mathrm{doc}}_x\} \cup \{R^{\mathrm{set}}_1, \ldots, R^{\mathrm{set}}_y\} \cup \{R^{\mathrm{global}}_1, \ldots, R^{\mathrm{global}}_z\}$, where $R^{\mathrm{doc}}_i$, $R^{\mathrm{set}}_j$, and $R^{\mathrm{global}}_k$ denote the $i$-th doc-level, $j$-th set-level, and $k$-th global-level rubric with weights $w^{\mathrm{doc}}_i$, $w^{\mathrm{set}}_j$, and $w^{\mathrm{global}}_k$ assigned during rubric construction, we compute the reward through a hierarchical aggregation with these weights.

The reward is provided by a rubric-based judge $J$, instantiated with DeepSeek-V4-Flash, which scores the selected set $\mathcal{S}$ against each rubric in $R_q$ on a $0$--$10$ scale. Set and global-level rubrics are rated once for the whole set, as $J(\cdot, \mathcal{S})$, whereas doc-level rubrics are rated per doc and averaged,
\begin{equation}
F(R^{\mathrm{doc}}_i, \mathcal{S}) = \frac{1}{|\mathcal{S}|} \sum\nolimits_{d \in \mathcal{S}} J(R^{\mathrm{doc}}_i, d).
\label{eq:doc_score}
\end{equation}
The three levels are then aggregated by their weights, instantiating the set utility of Eq.~\ref{eq:setwise},
\begin{equation}
u(\mathcal{S} \mid q) = \frac{\sum\nolimits_i w^{\mathrm{doc}}_i \cdot F(R^{\mathrm{doc}}_i, \mathcal{S}) + \sum\nolimits_j w^{\mathrm{set}}_j \cdot J(R^{\mathrm{set}}_j, \mathcal{S}) + \sum\nolimits_k w^{\mathrm{global}}_k \cdot J(R^{\mathrm{global}}_k, \mathcal{S})}{\sum\nolimits_i w^{\mathrm{doc}}_i + \sum\nolimits_j w^{\mathrm{set}}_j + \sum\nolimits_k w^{\mathrm{global}}_k},
\label{eq:utility}
\end{equation}

We further require the output to be a bracketed identifier sequence, e.g., \texttt{[2] [5] [8]}. The reward of rollout $y_i$ is then the utility of its parsed set $\mathcal{S}_i = \mathrm{Parse}(y_i)$ if the format is valid, and $-1$ otherwise,
\begin{equation}
r_i = \begin{cases} u(\mathcal{S}_i \mid q) \in [0,10], & \text{if the output format is correct},\\[4pt] -1, & \text{otherwise}. \end{cases}
\label{eq:reward}
\end{equation}
Following GRPO~\citep{grpo}, we then obtain the reinforcement advantage by standardizing $r_i$ within the group of $G$ rollouts sharing the same query,
\begin{equation}
A^{\mathrm{GRPO}}_{i,\ell} := \frac{r_i - \mathrm{mean}\{r_{i'}\}_{i'=1}^{G}}{\mathrm{std}\{r_{i'}\}_{i'=1}^{G}} \quad (\text{constant in } \ell),
\label{eq:adv_grpo}
\end{equation}
where $\ell$ indexes tokens of $y_i$. This advantage reflects the outcome but is uniform across the rollout, providing no token-level supervision. We thus introduce a distillation signal adapted to rollout quality.

\noindent \textbf{Adaptive Tutoring Hints.} A hint is privileged information used only in the distillation branch, where re-scoring $y_i$ under it yields the missing token-level signal. Prior methods use one hint type throughout training, so identical guidance is \ding{182} \textit{too prescriptive for strong rollouts} or \ding{183} \textit{too abstract for poor ones}. We therefore match hint form to rollout quality.

At the beginning of each training step, the frozen snapshot $\pi_{\theta_{\mathrm{old}}}$ drives two hint generators. The \textit{self-selector} maps $(q,\mathcal{C},R_q)$ to a sibling-set $\bar{h}$, an alternative document set chosen from $\mathcal{C}$ under the rubrics; the \textit{self-reflector} maps $(q,\mathcal{C},y_i,\bar{h})$ to a reflection $\hat{h}$ diagnosing which documents $y_i$ wrongly retained or missed relative to $\bar{h}$. With the rubric hint $h^{*}=R_q$, which states the criteria alone, these three forms constitute the hint space, and Figure~\ref{fig:prompts} shows how each is injected into the teacher prompt. Which form a rollout receives follows from its reward $r_i$ through two thresholds $\tau_1>\tau_2$,
\begin{equation}
\bm{h}_i=\begin{cases}h^{*}, & \tau_1\le r_i\le 10,\\[2pt]\hat{h}, & \tau_2<r_i<\tau_1,\\[2pt]\bar{h}, & 0\le r_i\le\tau_2.\end{cases}
\label{eq:hint}
\end{equation}
The tiers differ in how much correction a rollout can absorb. A high-reward rollout is already near a qualifying set, so $h^{*}$ refines it without overriding its justified choices; a mid-reward one is partly right, so $\hat{h}$ corrects only its errors; a low-reward one offers little to build on and is replaced outright by $\bar{h}$.

\studentteacherprompts{Prompt example for the student and teacher policies.}{fig:prompts}

\textbf{Sibling-set Gate.} Both $\hat{h}$ and $\bar{h}$ build on the sibling-set, so we gate it with the same judge: $\bar{h}$ is delivered only when it outscores the rollout it supervises, $u(\bar{h}\mid q)>r_i$; otherwise the rollout falls back to $h^{*}$, and no hint ever teaches toward a worse reference. We set $\tau_1=6$ and $\tau_2=3$ throughout.

\textbf{Joint Training Objective.} Beyond the outcome-driven GRPO advantage, we supervise the policy with a token-level \emph{adaptive tutoring distillation} (ATD) advantage derived from the hint. Without regenerating $y_i$, re-scoring it under $(q,\mathcal{C},\bm{h}_i)$ and $(q,\mathcal{C})$ yields, for its $\ell$-th token $\hat{y}_{i,\ell}$,
\begin{equation}
A^{\mathrm{ATD}}_{i,\ell}:=\log\frac{\pi_{\phi}\!\bigl(\hat{y}_{i,\ell}\mid q,\mathcal{C},\bm{h}_i,y_{i,<\ell}\bigr)}{\pi_{\theta_{\mathrm{old}}}\!\bigl(\hat{y}_{i,\ell}\mid q,\mathcal{C},y_{i,<\ell}\bigr)},
\label{eq:adv_atd}
\end{equation}
where the hint-conditioned teacher $\pi_{\phi}$ is the cold-start checkpoint $\theta_{\mathrm{sft}}$, frozen throughout the stage. Were it to move with the policy, the distillation branch would lose its external reference and risk reinforcing the policy's own mistakes. Since $\phi$ and $\theta_{\mathrm{old}}$ are both fixed within the update, $A^{\mathrm{ATD}}_{i,\ell}$ is constant in $\theta$ and enters purely as an advantage, positive where the hint makes a token more likely under the teacher; its sign anchors the policy to $\theta_{\mathrm{sft}}$, obviating an explicit KL penalty.

The final ATO advantage combines group-relative outcome feedback with token-level supervision,
\begin{equation}
A^{\mathrm{ATO}}_{i,\ell} := \lambda A^{\mathrm{GRPO}}_{i,\ell} + (1-\lambda) A^{\mathrm{ATD}}_{i,\ell}, \qquad \lambda \in [0,1],
\label{eq:convex}
\end{equation}
where $\lambda$ balances the two signals and is set to $0.7$ in our experiment.

This formulation keeps the outcome reward as the primary RL signal while adding token-level shaping. We optimize the standard clipped policy objective on the combined advantage,
\begin{equation}
\mathcal{L}_{\mathrm{policy}}(\theta)=-\mathbb{E}_{i,\ell}\Bigl[\min\bigl(\rho_{i,\ell}(\theta)\,A^{\mathrm{ATO}}_{i,\ell},\;\mathrm{clip}(\rho_{i,\ell}(\theta),1-\epsilon,1+\epsilon)\,A^{\mathrm{ATO}}_{i,\ell}\bigr)\Bigr],
\label{eq:ato_loss}
\end{equation}
where $\rho_{i,\ell}(\theta)$ denotes the token-level importance ratio, defined as
\begin{equation}
\rho_{i,\ell}(\theta)=\exp\bigl(\log\pi_{\theta}(\hat{y}_{i,\ell}\mid q,\mathcal{C},y_{i,<\ell})-\log\pi_{\theta_{\mathrm{old}}}(\hat{y}_{i,\ell}\mid q,\mathcal{C},y_{i,<\ell})\bigr).
\label{eq:ratio}
\end{equation}
Here $\mathrm{clip}$ bounds the importance ratio to $[1-\epsilon,1+\epsilon]$, with $\epsilon$ limiting the deviation from the old policy. As noted above, the anchoring in Eq.~\ref{eq:adv_atd} lets us omit the KL penalty.

\noindent \textbf{Training-inference boundary.} The self-selector, self-reflector, and hint serve only to construct the training advantage. At inference the policy acts from $(q,\mathcal{C})$ and the fixed meta-rubric instruction, plus the query intent for deep research, without the query-specific $R_q$ or its reference answer.

\section{Experiments}
\subsection{Settings}\label{sec:setup}

\noindent \textbf{Benchmarks.} We evaluate AdaTutoRank at two granularities. \ding{182} \textbf{Answer-level evaluation} assesses the final response under two scenarios. For \textbf{RAG}, we use Natural Questions~\citep{nq}, TriviaQA~\citep{triviaqa}, PopQA~\citep{popqa}, 2WikiMultihopQA~\citep{2wikimultihopqa}, and Bamboogle~\citep{bamboogle}, whose short closed-ended answers are measured by exact match (EM). For \textbf{deep research}, we use WebWalkerQA~\citep{webwalkerqa}, HealthBench~\citep{healthbench}, DeepResearchBench~\citep{deepresearchbench}, and ResearchQA~\citep{researchqa}, whose open-ended responses are scored by LLM judges. \ding{183} \textbf{Setwise-level evaluation} assesses the document set itself with SetwiseEvalKit~\citep{SetWiseEvalKit}, a rubric-based benchmark scoring a set along nine dimensions beyond relevance (\eg completeness, redundancy, conflict).

\noindent \textbf{Baselines.} We compare AdaTutoRank with 11 rerankers in three categories. \ding{182} \textbf{Adhoc reranking}, which orders documents by pointwise relevance: BGE-Reranker-Large~\citep{BGE-Reranker}, MonoT5~\citep{MonoT5}, RankT5~\citep{RankT5}, RankLlama~\citep{RankLlama}, RankVicuna~\citep{RankVicuna}, and RankZephyr~\citep{RankZephyr}. \ding{183} \textbf{Reasoning-enhanced reranking}, which adds explicit reasoning to the decision: Rearank~\citep{Rearank} and ReasonRank~\citep{ReasonRank}. \ding{184} \textbf{Setwise reranking}, which selects a set rather than a ranked list: Rank4Gen~\citep{Rank4Gen}, SetR~\citep{SetR}, and RubricRanker~\citep{rubricranker}. We also report Initial Retrieval, the unreranked top-5 from BM25~\citep{BM25} or the Google Search API, as a lower bound.

\noindent \textbf{Implementation Details.} AdaTutoRank uses Qwen3-8B~\citep{Qwen3} as its backbone. Documents come from the Google Search API on deep research benchmarks and from BM25 over the Wikipedia dump~\citep{dpr} on RAG benchmarks. Every reranker operates on the top-20 retrieved documents and returns a set. More details are provided in Appendix~\ref{appendix:setup} and~\ref{appendix:implementation_details}.

\begin{table*}[t!]
  \centering

    \caption{\textbf{Answer-Level Performance Comparison.} All rerankers rerank the top-$20$ documents, and we report exact match and LLM-judge scores on RAG and deep research respectively. The \colorbox{backred!50}{\textbf{best}} and \colorbox{backyellow_soft!40}{\underline{second-best}} results are highlighted.
All metrics are reported such that higher is better ($\uparrow$).}
  \label{tab:answer_level}
    \vspace{-6pt}
    
  \renewcommand{\arraystretch}{1.1}
  \resizebox{\textwidth}{!}{%
    \begin{tabular}{l|cccccc|ccccc|c}
      \toprule
      \multirow{2.5}{*}{\textbf{Ranker}}
        & \multicolumn{6}{c|}{\textbf{RAG}}
        & \multicolumn{5}{c|}{\textbf{Deep Research}}
        & \multicolumn{1}{c}{\cellcolor{gray!12}} \\
      \noalign{\global\aboverulesep=0pt \global\belowrulesep=0pt}
      \cmidrule(l{2pt}r{2pt}){2-7} \cmidrule(l{2pt}r{2pt}){8-12}
      \noalign{\global\aboverulesep=.65ex \global\belowrulesep=.65ex}
        & \cellcolor{gray!12}\rule{0pt}{2.6ex}\textbf{Avg.}
        & \textbf{NQ}
        & \textbf{Triv}
        & \textbf{PopQA}
        & \textbf{2Wiki}
        & \textbf{Bambo}
        & \cellcolor{gray!12}\textbf{Avg.}
        & \textbf{WebW}
        & \textbf{HealthB}
        & \textbf{DRB}
        & \textbf{RQA}
        & \cellcolor{gray!12}\raisebox{4.5pt}[0pt][0pt]{\textbf{Overall}} \\[-0.3ex]
      \midrule

      \rowcolor{gray!10}
      \multicolumn{13}{c}{\fontsize{10}{12}\selectfont \textit{\textbf{Only Retrieval (BM25 \& Google Search API)}}} \\

      Initial Retrieval & \cellcolor{gray!12}31.95 & 28.45 & 60.05 & 32.72 & 27.35 & 11.20 & \cellcolor{gray!12}47.98 & 32.00 & 45.40 & 48.62 & 65.89 & \cellcolor{gray!12}39.08 \\

      \midrule
      \rowcolor{gray!10}
      \multicolumn{13}{c}{\fontsize{10}{12}\selectfont \textit{\textbf{Adhoc Reranking}}} \\

      BGE-Reranker$_{L.}$ & \cellcolor{gray!12}35.16 & 32.94 & 62.80 & 35.85 & 29.82 & 14.40 & \cellcolor{gray!12}\colorbox{backyellow_soft!40}{\underline{51.01}} & \colorbox{backred!50}{\textbf{42.00}} & 45.60 & 48.84 & 67.59 & \cellcolor{gray!12}42.20 \\

      MonoT5 {\small(3B)} & \cellcolor{gray!12}34.42 & 31.25 & 62.38 & 35.19 & 28.87 & 14.40 & \cellcolor{gray!12}49.73 & 39.00 & 44.09 & 48.01 & 67.81 & \cellcolor{gray!12}41.22 \\

      RankT5 {\small(3B)} & \cellcolor{gray!12}34.13 & 31.75 & 62.34 & 35.47 & 29.07 & 12.00 & \cellcolor{gray!12}49.70 & 40.00 & 43.57 & 48.15 & 67.07 & \cellcolor{gray!12}41.05 \\

      RankLlama {\small(7B)} & \cellcolor{gray!12}33.77 & 32.38 & 62.17 & 35.40 & 29.32 & 9.60 & \cellcolor{gray!12}49.55 & 36.00 & 46.44 & 48.35 & 67.39 & \cellcolor{gray!12}40.78 \\

      RankVicuna {\small(7B)} & \cellcolor{gray!12}34.61 & 32.58 & 61.85 & 35.10 & 28.30 & 15.20 & \cellcolor{gray!12}50.34 & 35.00 & \colorbox{backyellow_soft!40}{\underline{46.92}} & 48.39 & \colorbox{backyellow_soft!40}{\underline{71.04}} & \cellcolor{gray!12}41.60 \\

      RankZephyr {\small(7B)} & \cellcolor{gray!12}34.89 & 33.68 & 62.19 & 35.32 & 28.08 & 15.20 & \cellcolor{gray!12}50.55 & \colorbox{backyellow_soft!40}{\underline{41.00}} & 44.80 & 47.91 & 68.48 & \cellcolor{gray!12}41.85 \\

      \midrule
      \rowcolor{gray!10}
      \multicolumn{13}{c}{\fontsize{10}{12}\selectfont \textit{\textbf{Reasoning-Enhanced Reranking}}} \\

      Rearank {\small(7B)} & \cellcolor{gray!12}35.11 & 32.41 & 62.31 & 35.31 & 30.32 & 15.20 & \cellcolor{gray!12}49.16 & 36.00 & 44.70 & 47.64 & 68.29 & \cellcolor{gray!12}41.35 \\

      ReasonRank {\small(7B)} & \cellcolor{gray!12}34.98 & 33.02 & 62.09 & 35.49 & 30.72 & 13.60 & \cellcolor{gray!12}48.83 & 37.00 & 43.00 & 48.12 & 67.18 & \cellcolor{gray!12}41.14 \\

      \midrule
      \rowcolor{gray!10}
      \multicolumn{13}{c}{\fontsize{10}{12}\selectfont \textit{\textbf{Setwise Reranking}}} \\

      Rank4Gen {\small(8B)} & \cellcolor{gray!12}37.00 & 34.93 & 65.76 & \colorbox{backyellow_soft!40}{\underline{37.87}} & 31.23 & 15.20 & \cellcolor{gray!12}49.26 & 37.00 & 46.60 & 48.15 & 65.30 & \cellcolor{gray!12}42.45 \\

      SetR {\small(8B)} & \cellcolor{gray!12}36.87 & 34.35 & 65.64 & 37.32 & \colorbox{backyellow_soft!40}{\underline{31.85}} & 15.20 & \cellcolor{gray!12}50.96 & 40.00 & 46.00 & 49.31 & 68.54 & \cellcolor{gray!12}43.13 \\



      {RubricRanker {\small(8B)}} & \cellcolor{gray!12}\colorbox{backyellow_soft!40}{\underline{38.25}} & \colorbox{backyellow_soft!40}{\underline{35.68}} & \colorbox{backyellow_soft!40}{\underline{66.30}} & 37.65 & 31.61 & \colorbox{backyellow_soft!40}{\underline{20.00}} & \cellcolor{gray!12}50.83 & 37.00 & 45.90 & \colorbox{backred!50}{\textbf{49.95}} & 70.48 & \cellcolor{gray!12}\colorbox{backyellow_soft!40}{\underline{43.84}} \\

      \midrule

      {AdaTutoRank {\small(8B)}} & \cellcolor{gray!12}\colorbox{backred!50}{\textbf{39.05}} & \colorbox{backred!50}{\textbf{36.87}} & \colorbox{backred!50}{\textbf{66.54}} & \colorbox{backred!50}{\textbf{38.52}} & \colorbox{backred!50}{\textbf{32.52}} & \colorbox{backred!50}{\textbf{20.80}} & \cellcolor{gray!12}\colorbox{backred!50}{\textbf{53.06}} & \colorbox{backyellow_soft!40}{\underline{41.00}} & \colorbox{backred!50}{\textbf{48.57}} & \colorbox{backyellow_soft!40}{\underline{49.89}} & \colorbox{backred!50}{\textbf{72.78}} & \cellcolor{gray!12}\colorbox{backred!50}{\textbf{45.28}} \\

      \bottomrule
    \end{tabular}%
  }
\end{table*}

\subsection{Analysis of Experimental Results}\label{sec:analysis}

\noindent \textbf{Analysis of Main Results.} Tables~\ref{tab:answer_level} and~\ref{tab:setwise_level_short} report AdaTutoRank against baselines at two granularities. Answer-level results show whether the downstream task improves, but confound the generator with the evidence it receives; setwise-level results isolate the evidence, revealing whether gains originate from a better selected set. We draw three observations: \ding{182} \textbf{\textit{AdaTutoRank improves downstream answer performance across the board.}} In Table~\ref{tab:answer_level}, it attains an overall score of $45.28$, exceeding the strongest baseline RubricRanker by $1.44\uparrow$ points and Initial Retrieval by $6.20\uparrow$, and it ranks first on seven of the nine benchmarks and second on the remaining two. \ding{183} \textbf{\textit{AdaTutoRank yields larger gains in deep research than in RAG.}} In Table~\ref{tab:answer_level}, it raises the scenario average by $0.80\uparrow$ on RAG ($39.05$ vs. $38.25$) and by $2.05\uparrow$ on deep research ($53.06$ vs. $51.01$). This follows from how far the evidence acts: in RAG the reranker runs once, its effect confined to a single generation step, whereas in deep research each set becomes the observation for the next sub-query, so better evidence compounds over the trajectory. \ding{184} \textbf{\textit{AdaTutoRank attains superior evidence quality on set selection.}} In Table~\ref{tab:setwise_level_short}, it attains the highest average at all three rubric levels, ranks first on seven of the nine fine-grained dimensions, and lifts the overall score by $2.17$ points over the strongest baseline RubricRanker. As these sets are what the generator and the agent consume, the gains in Table~\ref{tab:answer_level} trace to the evidence rather than its consumer, resolving the confound above.

\begin{table*}[t!]
  \centering
  \caption{\textbf{Setwise-Level Performance Comparison on SetwiseEvalKit (Short-form Scenario).}
All metrics are reported such that higher is better ($\uparrow$).}
  \label{tab:setwise_level_short}
  \vspace{-6pt}
  \renewcommand{\arraystretch}{1.1}
  \resizebox{\textwidth}{!}{%
    \begin{tabular}{l|cccc|cccc|cccc|c}
      \toprule
      \multirow{2.5}{*}{\textbf{Ranker}}
        & \multicolumn{4}{c|}{\textbf{Doc-Level}}
        & \multicolumn{4}{c|}{\textbf{Set-Level}}
        & \multicolumn{4}{c|}{\textbf{Global-Level}}
        & \cellcolor{gray!12} \\
      \noalign{\global\aboverulesep=0pt \global\belowrulesep=0pt}
      \cmidrule(l{2pt}r{2pt}){2-5} \cmidrule(l{2pt}r{2pt}){6-9} \cmidrule(l{2pt}r{2pt}){10-13}
      \noalign{\global\aboverulesep=.65ex \global\belowrulesep=.65ex}
        &
        \cellcolor{gray!12}\textbf{Avg} & \textbf{Rel.} & \textbf{Aut.} & \textbf{Qua.}
        & \cellcolor{gray!12}\textbf{Avg} & \textbf{Cmp.} & \textbf{Red.} & \textbf{Con.}
        & \cellcolor{gray!12}\textbf{Avg} & \textbf{Cpl.} & \textbf{Den.} & \textbf{Rea.}
        & \cellcolor{gray!12}\raisebox{4.5pt}[0pt][0pt]{\textbf{Overall}} \\
      \midrule
Initial Retrieval& \cellcolor{gray!12}16.30 & 14.67 & 17.18 & 17.05 & \cellcolor{gray!12}60.91 & 18.29 & 71.57 & 92.88 & \cellcolor{gray!12}32.08 & 46.59 & 20.05 & 29.60 & \cellcolor{gray!12}36.43 \\
BGE-Reranker$_{L.}$& \cellcolor{gray!12}20.30 & 18.89 & 21.32 & 20.70 & \cellcolor{gray!12}63.05 & 28.04 & 67.86 & 93.26 & \cellcolor{gray!12}38.34 & 54.35 & 21.03 & 39.64 & \cellcolor{gray!12}40.57 \\
MonoT5 {\small(3B)}& \cellcolor{gray!12}19.90 & 19.00 & 20.35 & 20.36 & \cellcolor{gray!12}62.76 & 26.92 & 68.74 & 92.61 & \cellcolor{gray!12}38.17 & 53.97 & 21.29 & 39.26 & \cellcolor{gray!12}40.28 \\
RankT5 {\small(3B)}& \cellcolor{gray!12}20.12 & 19.47 & 20.43 & 20.47 & \cellcolor{gray!12}62.86 & 26.86 & 68.61 & 93.12 & \cellcolor{gray!12}37.97 & 54.08 & 21.17 & 38.67 & \cellcolor{gray!12}40.32 \\
RankLlama {\small(7B)}& \cellcolor{gray!12}19.63 & 19.13 & 19.94 & 19.81 & \cellcolor{gray!12}63.08 & 26.07 & 69.51 & 93.66 & \cellcolor{gray!12}37.38 & 53.46 & 20.75 & 37.94 & \cellcolor{gray!12}40.03 \\
{Rearank {\small(7B)}}& \cellcolor{gray!12}19.28 & 18.47 & 19.93 & 19.44 & \cellcolor{gray!12}63.34 & 26.24 & 70.23 & 93.55 & \cellcolor{gray!12}38.00 & 53.83 & 20.99 & 39.17 & \cellcolor{gray!12}40.21 \\
{ReasonRank {\small(7B)}}& \cellcolor{gray!12}19.23 & 18.59 & 19.88 & 19.23 & \cellcolor{gray!12}64.99 & 27.90 & 72.52 & 94.54 & \cellcolor{gray!12}38.82 & 54.62 & 21.39 & 40.45 & \cellcolor{gray!12}41.01 \\
{Rank4Gen {\small(8B)}}& \cellcolor{gray!12}25.85 & 25.07 & 26.08 & 26.39 & \cellcolor{gray!12}\colorbox{backyellow_soft!40}{\underline{67.65}} & 28.04 & \colorbox{backyellow_soft!40}{\underline{80.41}} & 94.49 & \cellcolor{gray!12}38.12 & 54.01 & 21.99 & 38.37 & \cellcolor{gray!12}43.87 \\
{SetR {\small(8B)}}& \cellcolor{gray!12}33.44 & 29.98 & \colorbox{backyellow_soft!40}{\underline{35.57}} & 34.77 & \cellcolor{gray!12}64.04 & 27.82 & 71.41 & 92.88 & \cellcolor{gray!12}40.08 & 55.35 & 24.62 & 40.28 & \cellcolor{gray!12}45.85 \\
{RubricRanker {\small(8B)}}& \cellcolor{gray!12}\colorbox{backyellow_soft!40}{\underline{36.15}} & \colorbox{backyellow_soft!40}{\underline{33.69}} & \colorbox{backred!50}{\textbf{38.07}} & \colorbox{backyellow_soft!40}{\underline{36.69}} & \cellcolor{gray!12}63.31 & \colorbox{backyellow_soft!40}{\underline{28.25}} & 65.10 & \colorbox{backyellow_soft!40}{\underline{96.58}} & \cellcolor{gray!12}\colorbox{backyellow_soft!40}{\underline{40.96}} & \colorbox{backyellow_soft!40}{\underline{56.25}} & \colorbox{backyellow_soft!40}{\underline{24.97}} & \colorbox{backred!50}{\textbf{41.65}} & \cellcolor{gray!12}\colorbox{backyellow_soft!40}{\underline{46.80}} \\
      \midrule
{AdaTutoRank {\small(8B)}}& \cellcolor{gray!12}\colorbox{backred!50}{\textbf{36.59}} & \colorbox{backred!50}{\textbf{39.55}} & 33.25 & \colorbox{backred!50}{\textbf{36.98}} & \cellcolor{gray!12}\colorbox{backred!50}{\textbf{69.28}} & \colorbox{backred!50}{\textbf{28.72}} & \colorbox{backred!50}{\textbf{81.49}} & \colorbox{backred!50}{\textbf{97.64}} & \cellcolor{gray!12}\colorbox{backred!50}{\textbf{41.04}} & \colorbox{backred!50}{\textbf{56.93}} & \colorbox{backred!50}{\textbf{25.32}} & \colorbox{backyellow_soft!40}{\underline{40.88}} & \cellcolor{gray!12}\colorbox{backred!50}{\textbf{48.97}} \\
      \bottomrule
    \end{tabular}%
  }
\end{table*}

\begin{figure*}[h!]
\centering
\begin{minipage}[t]{0.53\textwidth}
  \centering
  \vspace{0pt}
  \setlength{\abovecaptionskip}{2pt}
  \setlength{\belowcaptionskip}{0pt}
  \includegraphics[width=\linewidth]{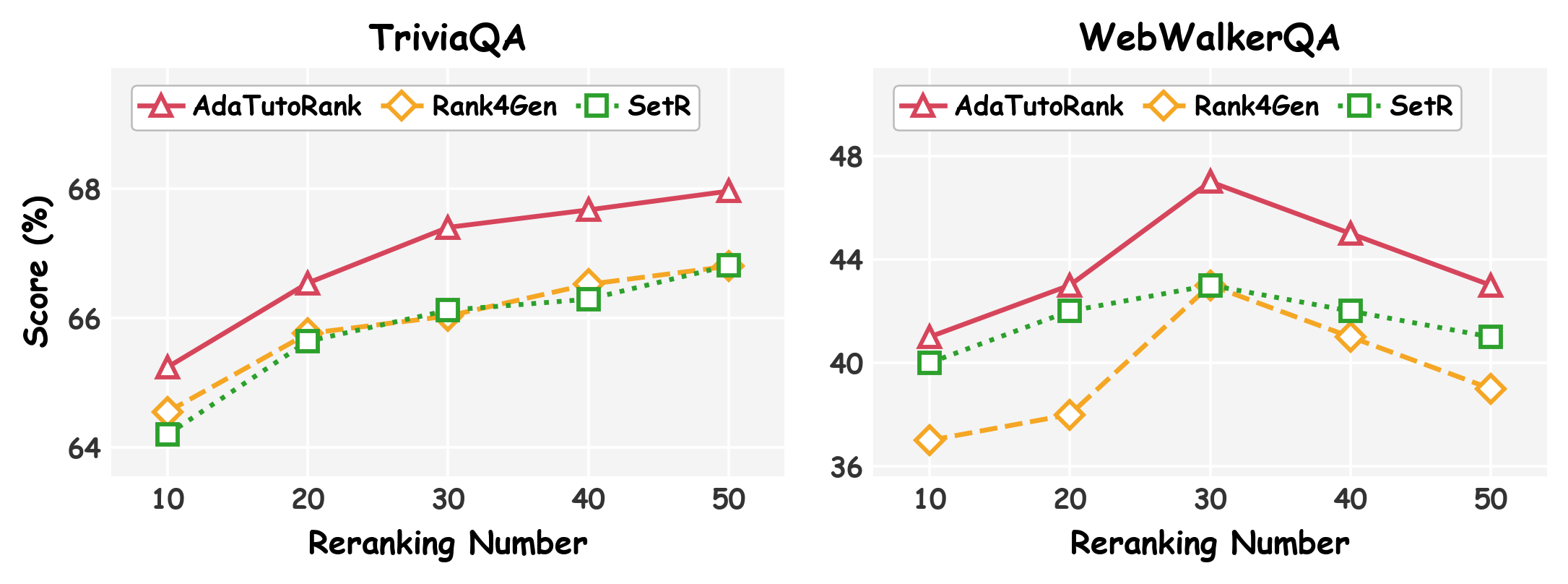}
  \captionof{figure}{Results of different candidate doc numbers.}
  \label{fig:doc_count}

  \vspace{4pt}

  \includegraphics[width=\linewidth]{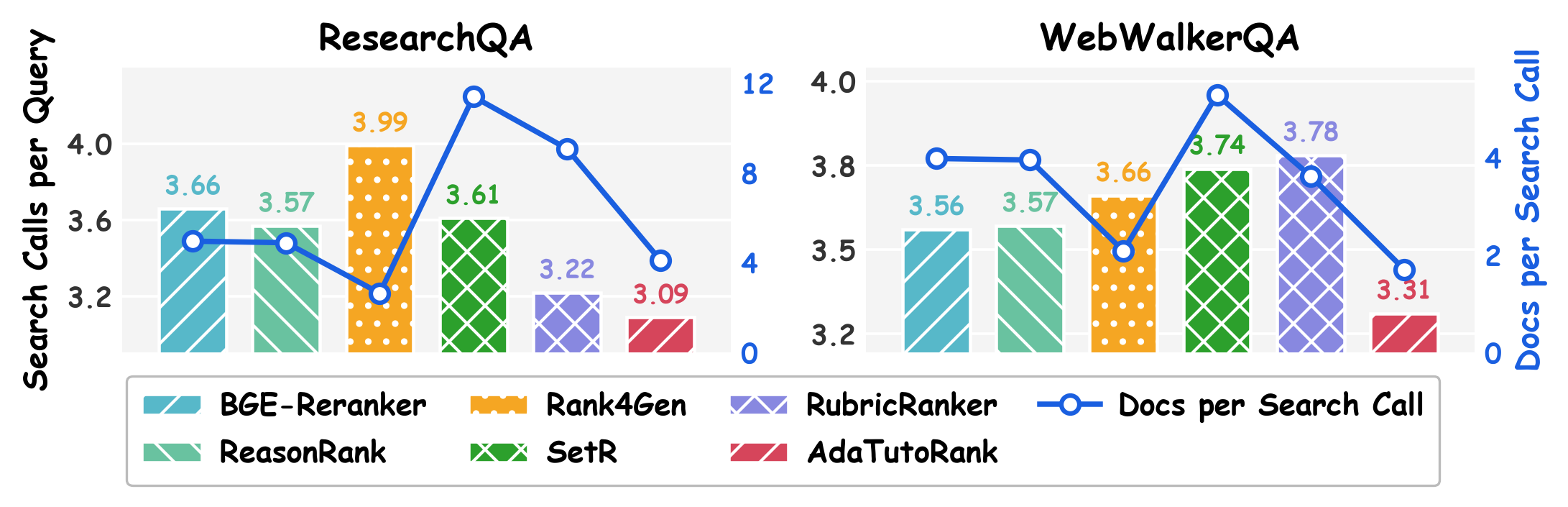}
  \captionof{figure}{Search behaviour of different rerankers.}
  \label{fig:search_call}
\end{minipage}%
\hfill
\begin{minipage}[t]{0.44\textwidth}
  \centering
  \vspace{0pt}
  \captionof{table}{Ablation Study of AdaTutoRank on the RAG and Deep Research Benchmarks.}
  \label{tab:ablation}
  \renewcommand{\arraystretch}{1.1}
  \resizebox{\linewidth}{!}{%
    \begin{tabular}{l|cccc|c}
      \toprule
      \textbf{Setting}
        & \textbf{NQ}
        & \textbf{PopQA}
        & \textbf{HealthB}
        & \textbf{RQA}
        & \cellcolor{gray!12}\textbf{Overall} \\
      \midrule
      AdaTutoRank & 36.87 & 38.52 & 48.57 & 72.78 & \cellcolor{gray!12}\colorbox{backred!50}{\textbf{49.19}} \\
      \midrule
      \rowcolor{gray!10}
      \multicolumn{6}{c}{\fontsize{10}{12}\selectfont \textit{\textbf{Training Stage}}} \\
      w/o ATO & 36.45 & 38.25 & 40.96 & 69.67 & \cellcolor{gray!12}46.33 \\
      w/o SFT & 34.79 & 38.45 & 45.24 & 69.67 & \cellcolor{gray!12}47.04 \\
      \midrule
      \rowcolor{gray!10}
      \multicolumn{6}{c}{\fontsize{10}{12}\selectfont \textit{\textbf{Advantage Superposition}}} \\
      Only RL           & 31.88 & 36.23 & 46.11 & 66.06 & \cellcolor{gray!12}45.07 \\
      Only Distillation & 36.43 & 38.39 & 46.38 & 69.95 & \cellcolor{gray!12}\colorbox{backyellow_soft!40}{\underline{47.79}} \\
      \midrule
      \rowcolor{gray!10}
      \multicolumn{6}{c}{\fontsize{10}{12}\selectfont \textit{\textbf{Interpolation Coefficient}}} \\
      $\lambda = 0.9$ & 33.16 & 36.17 & 47.42 & 69.71 & \cellcolor{gray!12}46.61 \\
      $\lambda = 0.5$ & 36.70 & 38.47 & 45.82 & 70.00 & \cellcolor{gray!12}47.75 \\
      \midrule
      \rowcolor{gray!10}
      \multicolumn{6}{c}{\fontsize{10}{12}\selectfont \textit{\textbf{Privileged Information}}} \\
      Only Rubrics $h^{*}$     & 36.51 & 38.43 & 47.20 & 67.78 & \cellcolor{gray!12}47.48 \\
      Only Reflection $\hat{h}$  & 35.60 & 37.77 & 47.36 & 69.77 & \cellcolor{gray!12}47.62 \\
      Only Sibling-Set $\bar{h}$ & 35.82 & 38.22 & 43.11 & 70.36 & \cellcolor{gray!12}46.88 \\
      \bottomrule
    \end{tabular}%
  }
\end{minipage}
\end{figure*}

\noindent \textbf{Analysis of Ablation Experiment Results.} We ablate four aspects of AdaTutoRank, with results reported in Table~\ref{tab:ablation}, and summarize our observations as follows. \ding{182} \textbf{\textit{Training Stage.}} We ablate the two-stage framework by removing ATO (``w/o ATO'') and cold-start SFT (``w/o SFT''). Dropping ATO lowers the overall score from $49.19$ to $46.33$, while dropping SFT lowers it to $47.04$. Removing either stage costs a comparable amount, indicating that the cold start and ATO are complementary rather than substitutable. \ding{183} \textbf{\textit{Advantage Superposition.}} Optimizing with the outcome advantage alone (``Only RL'') yields $45.07$, even below the $46.33$ of SFT alone, indicating that a scalar reward shared by every selected identifier is too coarse to improve set selection. The distillation advantage alone (``Only Distillation'') reaches $47.79$, and superposing the two as in Eq.~\ref{eq:convex} gives the best $49.19$, confirming that the two signals are complementary. \ding{184} \textbf{\textit{Interpolation Coefficient.}} Performance degrades on both sides of $\lambda=0.7$ ($49.19$), falling to $47.75$ at $\lambda=0.5$ and $46.61$ at $\lambda=0.9$. Each end collapses toward the corresponding single-signal regime: a large $\lambda$ suppresses the tutoring advantage and drifts toward ``Only RL'' ($45.07$), while a small one weakens the outcome grounding and nearly reproduces ``Only Distillation'' ($47.79$), so neither signal can dominate. \ding{185} \textbf{\textit{Privileged Information.}} Restricting the hint to one fixed form degrades performance, yielding $47.48$ (rubrics), $46.88$ (sibling sets), and $47.62$ (reflections). The extremes fail oppositely: abstract rubrics give a badly wrong rollout no foothold, while a sibling-set overrides the equally valid choices of a strong one. The adaptive tutoring hint of Eq.~\ref{eq:hint} is therefore necessary.

\vspace{5pt}

\vspace{5pt}

\noindent \textbf{Varying Number of Candidate Documents and Search Call Rounds.} We further examine how AdaTutoRank behaves as its input and downstream usage vary: Figure~\ref{fig:doc_count} reports performance across candidate pool sizes, and Figure~\ref{fig:search_call} inspects the retrieval behavior it induces in a deep research agent. \ding{182} \textbf{\textit{AdaTutoRank is robust to the candidate pool size.}} In Figure~\ref{fig:doc_count}, AdaTutoRank stays strongest as the pool grows from $10$ to $50$. On TriviaQA, performance rises steadily with pool size; on WebWalkerQA, all three rerankers peak at $30$ and decline thereafter, implicating the retrieval source rather than the reranker: web results for agent-issued sub-queries contain a long tail of low-value pages that inflate the context without adding evidence. \ding{183} \textbf{\textit{AdaTutoRank is more efficient in search agents.}} As shown in Figure~\ref{fig:search_call}, AdaTutoRank incurs the fewest search calls on the two deep research benchmarks while selecting fewer documents per round. Its sets are thus both sufficient and compact: the agent meets its information need without issuing extra queries.





\section{Conclusion and Discussion}

We present AdaTutoRank, a setwise reranker for RAG and deep research that optimizes set-level utility. To make utility supervisable, we represent it as hierarchical query-specific rubrics spanning the document, set, and global levels, which score individual documents, their interaction, and the set as a whole. Our two-stage framework cold-starts the policy on rubric-conditioned silver labels, then runs adaptive tutoring optimization, superposing the outcome reward with a token-level distillation advantage from a hint matched to each rollout's quality. At inference it selects a high-quality subset directly, without query-specific rubrics or hints. Experiments show consistent gains at both the answer and setwise levels, with fewer search calls in deep research agents. More broadly, rubrics can supply not only evaluation but also dense token-level supervision distilled back into the policy.

\bibliography{references}
\bibliographystyle{main}

\clearpage
\appendix

\clearpage

\section{What Adaptive Tutoring Optimization Optimizes}\label{appendix:theory}

\makeatletter
\@ifundefined{assumption}{\newtheorem{assumption}{Assumption}}{}
\@ifundefined{lemma}{\newtheorem{lemma}{Lemma}}{}
\@ifundefined{proposition}{\newtheorem{proposition}{Proposition}}{}
\@ifundefined{corollary}{\newtheorem{corollary}{Corollary}}{}
\makeatother
\providecommand{\KL}{\mathrm{KL}}
\providecommand{\Ex}{\mathbb{E}}

This section identifies the objective behind Eq.~\ref{eq:ato_loss}, shows that the distillation branch already contains a reverse-KL anchor to the cold start, and explains why the teacher is frozen at $\theta_{\mathrm{sft}}$ rather than tracking the policy.

\noindent \textbf{Setup.} Write $x=(q,\mathcal{C})$ for the hint-free context and $x_{h}=(q,\mathcal{C},\bm{h}_i)$ for the hint-augmented one. For a rollout $y_i\sim\pi_{\theta_{\mathrm{old}}}(\cdot\mid x)$ and a position $\ell$, abbreviate the per-token conditionals on the prefix $y_{i,<\ell}$ as $p_{\theta}$, $p_{\mathrm{old}}$, and $p^{h}_{\phi}$ for $\pi_{\theta}(\cdot\mid x,\cdot)$, $\pi_{\theta_{\mathrm{old}}}(\cdot\mid x,\cdot)$, and $\pi_{\phi}(\cdot\mid x_{h},\cdot)$, and let $p_{\phi}:=\pi_{\phi}(\cdot\mid x,\cdot)$ be the same frozen teacher evaluated \emph{without} the hint, a quantity the algorithm never computes but which is well defined and needed below. Every divergence that follows is summed over positions with prefixes drawn from $\pi_{\theta_{\mathrm{old}}}$, the form the token-level advantage implements. With this notation Eq.~\ref{eq:adv_atd} reads $A^{\mathrm{ATD}}_{i,\ell}=\log p^{h}_{\phi}(\hat y_{i,\ell})-\log p_{\mathrm{old}}(\hat y_{i,\ell})$ with $\hat y_{i,\ell}\sim p_{\mathrm{old}}$, and Eq.~\ref{eq:convex} reads $A^{\mathrm{ATO}}_{i,\ell}=\lambda A^{\mathrm{GRPO}}_{i}+(1-\lambda)m_iA^{\mathrm{ATD}}_{i,\ell}$, where $m_i=\mathds{1}[r_i\neq-1]$ is the format mask of Algorithm~\ref{alg:ato}. It excludes malformed rollouts from the distillation branch, so a rollout receiving $r_i=-1$ under Eq.~\ref{eq:reward} is supervised by the outcome advantage alone and the hint that Eq.~\ref{eq:hint} assigns to it never takes effect; every statement below is therefore vacuous for such rollouts.

\begin{assumption}[Detached advantages at the snapshot]\label{asm:detach}
Both advantages are evaluated at parameters fixed within the update, so they are constants with respect to $\theta$, and $\epsilon>0$. Gradients are taken at $\theta=\theta_{\mathrm{old}}$, which is exact for the first inner step of each update.
\end{assumption}

\begin{lemma}[The clip is inactive at the snapshot]\label{lem:clip}
Under Assumption~\ref{asm:detach}, $\nabla_{\theta}\mathcal{L}_{\mathrm{policy}}\big|_{\theta_{\mathrm{old}}}=-\Ex\bigl[A^{\mathrm{ATO}}_{i,\ell}\,\nabla_{\theta}\log p_{\theta}(\hat y_{i,\ell})\bigr]$.
\end{lemma}
\begin{proof}
At $\theta=\theta_{\mathrm{old}}$ we have $\rho_{i,\ell}=1$, interior to $[1-\epsilon,1+\epsilon]$, so in a neighbourhood both arguments of the minimum equal $\rho_{i,\ell}A^{\mathrm{ATO}}_{i,\ell}$. Differentiating and using $\nabla_{\theta}\rho_{i,\ell}=\nabla_{\theta}\log p_{\theta}$ at $\rho_{i,\ell}=1$ gives the claim.
\end{proof}

\begin{lemma}[The snapshot divergence is first-order flat]\label{lem:flat}
$\nabla_{\theta}\KL(p_{\theta}\,\|\,p_{\mathrm{old}})\big|_{\theta=\theta_{\mathrm{old}}}=0$.
\end{lemma}
\begin{proof}
Differentiating $\KL(p_{\theta}\|p_{\mathrm{old}})=\Ex_{p_{\theta}}[\log p_{\theta}-\log p_{\mathrm{old}}]$ gives $\Ex_{p_{\theta}}[\nabla_{\theta}\log p_{\theta}(\log p_{\theta}-\log p_{\mathrm{old}})]+\Ex_{p_{\theta}}[\nabla_{\theta}\log p_{\theta}]$. At $\theta=\theta_{\mathrm{old}}$ the first bracket vanishes pointwise and the second term is zero since $\Ex_{p_{\theta}}[\nabla_{\theta}\log p_{\theta}]=\nabla_{\theta}\sum_{y}p_{\theta}(y)=0$.
\end{proof}

\begin{proposition}[ATO is reinforcement learning with a reverse-KL pull toward the hint-conditioned teacher]\label{prop:ato_objective}
Under Assumption~\ref{asm:detach}, the first inner step of an update follows the ascent direction of
\begin{equation}
J_{\mathrm{ATO}}(\theta)\;=\;\lambda\,J_{\mathrm{GRPO}}(\theta)\;-\;(1-\lambda)\,m_i\,\KL\bigl(p_{\theta}\,\big\|\,p^{h}_{\phi}\bigr),
\qquad
J_{\mathrm{GRPO}}(\theta):=\Ex_{y\sim\pi_{\theta}(\cdot\mid x)}\bigl[A^{\mathrm{GRPO}}(y)\bigr].
\label{eq:thy_ato}
\end{equation}
\end{proposition}
\begin{proof}
Let $J_{\mathrm{ATD}}(\theta):=\Ex_{\hat y\sim p_{\theta}}[\log p^{h}_{\phi}(\hat y)-\log p_{\mathrm{old}}(\hat y)]$. Adding and subtracting $\Ex_{p_{\theta}}[\log p_{\theta}]$ gives $J_{\mathrm{ATD}}=-\KL(p_{\theta}\|p^{h}_{\phi})+\KL(p_{\theta}\|p_{\mathrm{old}})$, and Lemma~\ref{lem:flat} kills the gradient of the second summand, so $\nabla_{\theta}J_{\mathrm{ATD}}\big|_{\theta_{\mathrm{old}}}=-\nabla_{\theta}\KL(p_{\theta}\|p^{h}_{\phi})$. By Lemma~\ref{lem:clip} and the score-function identity $\nabla_{\theta}\Ex_{p_{\theta}}[f]=\Ex_{p_{\theta}}[f\nabla_{\theta}\log p_{\theta}]$, valid for any $\theta$-independent $f$ and applied to $A^{\mathrm{GRPO}}$ at the level of rollouts and to $A^{\mathrm{ATD}}_{i,\ell}$ at the level of tokens, $\nabla_{\theta}\mathcal{L}_{\mathrm{policy}}=-\lambda\nabla_{\theta}J_{\mathrm{GRPO}}-(1-\lambda)m_i\nabla_{\theta}J_{\mathrm{ATD}}$.
\end{proof}

Eq.~\ref{eq:thy_ato} makes the role of $\lambda$ precise: it trades the outcome objective against a regularizer whose target is selected per rollout by Eq.~\ref{eq:hint}. Fixing $\bm{h}_i\equiv h^{*}$ recovers self-distillation from a single privileged context~\citep{OPSD} as the special case in which that target does not depend on $i$.

\begin{proposition}[The distillation branch contains an anchor to the cold start]\label{prop:decomp}
Let the \emph{hint lift} be $\Delta(y):=\log p^{h}_{\phi}(y)-\log p_{\phi}(y)$, which depends only on the frozen $\phi$ and is therefore independent of $\theta$. Then $\KL(p_{\theta}\|p^{h}_{\phi})=\KL(p_{\theta}\|p_{\phi})-\Ex_{p_{\theta}}[\Delta]$, and hence, with $\phi=\theta_{\mathrm{sft}}$,
\begin{equation}
J_{\mathrm{ATO}}(\theta)\;=\;\underbrace{\lambda\,J_{\mathrm{GRPO}}(\theta)}_{\text{outcome objective}}\;+\;\underbrace{(1-\lambda)m_i\,\Ex_{p_{\theta}}\bigl[\Delta\bigr]}_{\text{token-level hint shaping}}\;-\;\underbrace{(1-\lambda)m_i\,\KL\bigl(p_{\theta}\,\big\|\,\pi_{\theta_{\mathrm{sft}}}\bigr)}_{\text{reverse-KL anchor to the cold start}}.
\label{eq:thy_three_terms}
\end{equation}
\end{proposition}
\begin{proof}
Both divergences share the term $\Ex_{p_{\theta}}[\log p_{\theta}]$, so their difference is $\Ex_{p_{\theta}}[\log p_{\phi}-\log p^{h}_{\phi}]=-\Ex_{p_{\theta}}[\Delta]$. Substituting into Eq.~\ref{eq:thy_ato} and using $p_{\phi}=\pi_{\theta_{\mathrm{sft}}}(\cdot\mid x,y_{i,<\ell})$ gives the claim.
\end{proof}

\begin{corollary}[An explicit KL penalty rescales a coefficient rather than adding a constraint]\label{cor:no_kl}
Since $\Delta$ is the only $h_i$-dependent factor in Eq.~\ref{eq:thy_three_terms} while the anchor is common to all rollouts, averaging over a group leaves the anchor with coefficient $(1-\lambda)\bar m$, where $\bar m=\frac{1}{G}\sum_i m_i$. An explicit penalty $\beta\KL(p_{\theta}\|\pi_{\theta_{\mathrm{sft}}})$ enters in the same position and with the same sign, so it changes that coefficient to $(1-\lambda)\bar m+\beta$ and supplies no regularizer of a different kind.
\end{corollary}

This is the precise sense of the claim in Section~\ref{sec:ato} that freezing the teacher lets us omit the KL penalty; the anchor also weakens in proportion to the share of malformed rollouts, which $m_i=0$ excludes. We do not claim that either mechanism bounds the cumulative displacement from $\theta_{\mathrm{sft}}$, only that both are first-order regularizers of the same type.

\noindent \textbf{Why the teacher is frozen.} A natural alternative is to let the teacher track the policy and to restore stability with an explicit KL term. Two results argue against it.

\begin{proposition}[A moving self-teacher annihilates the anchor]\label{prop:selfteacher}
Substituting $\phi:=\theta_{\mathrm{old}}$ into Proposition~\ref{prop:decomp} gives $p_{\phi}=p_{\mathrm{old}}$, so the anchor becomes $\KL(p_{\theta}\|p_{\mathrm{old}})$, whose gradient vanishes at $\theta=\theta_{\mathrm{old}}$ by Lemma~\ref{lem:flat}. The surviving objective $\lambda J_{\mathrm{GRPO}}+(1-\lambda)m_i\Ex_{p_{\theta}}[\Delta^{\mathrm{old}}]$, with $\Delta^{\mathrm{old}}(y):=\log\pi_{\theta_{\mathrm{old}}}(y\mid x_{h})-\log p_{\mathrm{old}}(y)$, contains no divergence to any parameter setting held fixed across updates.
\end{proposition}

An explicit KL term is therefore not an optional addition in that design but a repair of what the substitution removed, and it is the term Eq.~\ref{eq:thy_three_terms} already provides.

\begin{proposition}[Hint-conditioned behaviour is never supervised]\label{prop:no_grad}
Neither stage places a gradient on the policy's hint-conditioned conditionals. The cold-start objective of Eq.~\ref{eq:sft} is defined on hint-free inputs, since $R_q$ is withheld, and by Lemma~\ref{lem:clip} the ATO gradient depends on $\theta$ only through $\nabla_{\theta}\log\pi_{\theta}(\hat y_{i,\ell}\mid x,y_{i,<\ell})$. Hence for every hint the map $\bm{h}\mapsto\pi_{\theta}(\cdot\mid x_{h})$ appears in no objective and evolves only as a side effect of updating shared parameters.
\end{proposition}

This is the decisive difference between the two designs. Exploiting a hint is an in-context ability inherited from pre-training: it is used during training, never trained, and never invoked at inference. A self-teacher therefore draws its targets from an object that drifts with the policy while no term maintains its quality, so $\Delta^{\mathrm{old}}$ degrades in an uncontrolled way. It also admits a shortcut, since a policy that becomes insensitive to hints makes $\Delta^{\mathrm{old}}\equiv0$ and zeroes the distillation gradient regardless of the utility of its selected sets, at no cost under hint-free inference. Freezing $\phi=\theta_{\mathrm{sft}}$ removes both problems: $\Delta$ is a fixed function of $y$, and since $p^{h}_{\phi}$ does not depend on $\theta$, the distillation gradient can vanish only at stationary points of $\KL(p_{\theta}\|p^{h}_{\phi})$. It is also cheaper, requiring one extra forward pass per update rather than two, because $\pi_{\theta_{\mathrm{old}}}(\cdot\mid x)$ is already computed for the ratio of Eq.~\ref{eq:ratio}.

\noindent \textbf{Limitations.} Three caveats delimit the above. The identities are first-order at $\theta=\theta_{\mathrm{old}}$ and exact only for the first inner step, after which the clip may activate. The hint lift $\Delta$ is measured under $\phi$ rather than under the current policy, so it estimates the effect of the hint off-policy; this is the price of a stationary target. Finally, Eq.~\ref{eq:thy_three_terms} is a penalty rather than a constraint, and neither it nor an explicit KL term bounds the cumulative displacement of $\pi_{\theta}$ from $\pi_{\theta_{\mathrm{sft}}}$, since the clipped objective enforces only a per-update trust region.

\section{Details of Experimental Setup}\label{appendix:setup}

\subsection{Details of RAG, Deep Research and Setwise Benchmarks}

We evaluate AdaTutoRank on ten benchmarks: five short-form RAG datasets, four deep research datasets, and one setwise-level benchmark. Brief descriptions and evaluation protocols are given below.

\noindent \textbf{RAG Benchmarks.} All five datasets require short, closed-ended answers and are scored by exact match (EM) against the gold answers, with no LLM judge involved. \ding{182} \textbf{Natural Questions}~\citep{nq} consists of real queries issued to Google search, paired with answers annotated from Wikipedia. Its questions reflect authentic information needs and are mostly single-hop, making it a standard testbed for open-domain retrieval-augmented QA. \noindent \ding{183} \textbf{TriviaQA}~\citep{triviaqa} contains over 95K trivia-style question-answer pairs, each accompanied by independently collected evidence documents. Because the questions are written without reference to a specific passage, answering them requires retrieving evidence rather than matching surface lexical cues. \noindent \ding{184} \textbf{PopQA}~\citep{popqa} comprises 14K entity-centric questions generated from Wikidata triples, with a long-tail distribution of entity popularity. It is designed to expose the limits of parametric knowledge, and thus isolates the contribution of the retrieved evidence. \noindent \ding{185} \textbf{2WikiMultihopQA}~\citep{2wikimultihopqa} is constructed from Wikipedia and Wikidata with approximately 192K samples, providing explicit evidence paths in the form of triples and covering four question types: comparison, inference, compositional, and bridge-comparison. \noindent \ding{186} \textbf{Bamboogle}~\citep{bamboogle} is a curated test set of 125 two-hop questions designed so that no single search query can retrieve the answer, requiring models to identify and use intermediate bridging entities.

\noindent \textbf{Deep Research Benchmarks.} All four datasets require long-form, open-ended responses and are scored by LLM judges following the protocol released with each benchmark. For comparability and efficiency, we evaluate 100 queries per benchmark, sampling randomly where the official set is larger. \noindent \ding{187} \textbf{WebWalkerQA}~\citep{webwalkerqa} focuses on complex web and website-level question answering. Solving its questions requires agents to search, browse, and connect information distributed across web pages, rather than relying on a single retrieved passage. \noindent \ding{188} \textbf{HealthBench}~\citep{healthbench} is a healthcare-oriented benchmark that evaluates whether models can answer medically related user questions in a helpful, accurate, and safe manner. Its queries often require careful interpretation of user intent and reliable synthesis of evidence, making it suitable for evaluating long-form responses in high-stakes domains. \noindent \ding{189} \textbf{DeepResearchBench}~\citep{deepresearchbench} evaluates general-domain deep research capabilities using open-ended questions, emphasizing comprehensiveness, depth of analysis, instruction following, and readability. Its 100 queries are evenly split between English and Chinese; we score the generated articles on the four criteria above and report the macro average. Following the official protocol~\citep{deepresearchbench}, the Jina API is used to scrape URLs for evidence snippets when needed, while for our system outputs we use the URL contents collected by the corresponding search or browsing tools. \noindent \ding{190} \textbf{ResearchQA}~\citep{researchqa} contains research-style questions that require collecting external evidence and producing synthesized long-form answers. It tests whether an agent can identify useful information, organize evidence, and provide a coherent response to complex information needs. We use its official 100-question subset for evaluating deep research systems and report the average rubric score.

\noindent \textbf{Setwise-level Benchmark.} The remaining benchmark scores the retrieved set itself rather than the answer produced from it. \ding{191} \textbf{SetwiseEvalKit}~\citep{SetWiseEvalKit} instantiates query-specific criteria and scores a set along nine dimensions organized at three levels: the document level (relevance, authenticity, quality), the set level (complementarity, redundancy, conflict), and the global level (completeness, density, reachability). All metrics are reported such that higher is better, and we report the per-level averages together with the overall score. Our main results use the short-form scenario, where each query is paired with a candidate pool from which a set must be selected. Note that the benchmark instantiates its own criteria with its own judge, independently of the rubrics and reward judge used in our training pipeline (Section~\ref{sec:ato}); we therefore treat the setwise-level results as a diagnostic that localizes where the answer-level gains come from, rather than as an independent replication of them.

\subsection{Details of Reranker}

We compare AdaTutoRank with 11 rerankers spanning three categories, together with an unreranked lower bound. Each baseline is described below.

\noindent \ding{182} \textbf{Adhoc reranking.} These methods score candidates by pointwise relevance to the query and return a ranked list, from which the top-5 documents are taken as the selected set~\citep{li2026wow,zhang2026unichange,jiang2022attention,li2022fish,lin2021feasibility}. \textbf{BGE-Reranker-Large}~\citep{BGE-Reranker} is a cross-encoder released as part of the C-Pack suite. Query and document are concatenated into a single sequence so that full bidirectional attention can be applied across them, and the resulting representation is mapped to a relevance score. We adopt the \texttt{bge-reranker-large} checkpoint, an XLM-RoBERTa model of roughly 560M parameters. \textbf{MonoT5}~\citep{MonoT5} recasts relevance estimation as sequence-to-sequence generation: conditioned on a query--document pair, the model emits either ``true'' or ``false'', and the normalized probability assigned to ``true'' is used as the score. \textbf{RankT5}~\citep{RankT5} also builds on T5 but reads a scalar directly off the decoder logits instead of going through token probabilities, which allows it to be trained with pointwise, pairwise, or listwise ranking losses. We adopt the \texttt{rankt5-base} checkpoint fine-tuned on MS MARCO. \textbf{RankLLaMA}~\citep{RankLlama} fine-tunes LLaMA-2-7B for pointwise reranking. The hidden state of the final token serves as the sequence representation and is projected linearly to a scalar, with training driven by a pairwise ranking loss over hard negatives mined from both sparse and dense retrievers. \textbf{RankVicuna}~\citep{RankVicuna} is an early fully open listwise reranker for zero-shot settings. Starting from Vicuna-7B, it is trained by permutation distillation against a GPT-3.5-based teacher, learning to emit a reordered permutation of the candidates within a sliding window. \textbf{RankZephyr}~\citep{RankZephyr} is a 7B listwise reranker built on Zephyr. It combines permutation distillation from GPT-3.5 and GPT-4 with direct preference optimization, narrowing the gap to proprietary rerankers while remaining openly available.

\noindent \ding{183} \textbf{Reasoning-enhanced reranking.} These methods expose an explicit reasoning process before committing to a ranking decision, trading inference cost for accuracy on harder queries. \textbf{Rearank}~\citep{Rearank} treats listwise reranking as a reasoning task and optimizes the reasoning-then-ranking behavior with reinforcement learning, so that competitive rankings can be obtained from a modest amount of labeled supervision. \textbf{ReasonRank}~\citep{ReasonRank} targets reasoning-intensive retrieval. It is first warmed up on synthesized reasoning traces for ranking and then refined with a ranking-oriented reward, which yields substantial gains on benchmarks where relevance cannot be judged from surface overlap.

\noindent \ding{184} \textbf{Setwise reranking.} Rather than producing a full ordering, these methods predict the retained subset directly, and therefore differ in what supervision they use to define a good set. \textbf{Rank4Gen}~\citep{Rank4Gen} selects a subset with the downstream generator in mind, using the quality of the answer produced from a candidate set as the training signal, so that selection is optimized for generation rather than for human-perceived relevance. \textbf{SetR}~\citep{SetR} learns set selection by imitating a strong teacher, distilling the teacher's subset decisions into a smaller policy that can assemble a set in a single pass. \textbf{RubricRanker}~\citep{rubricranker} is the closest baseline to our work. It scores a candidate set against query-specific rubrics and uses the aggregated score as a reinforcement learning reward, which supervises set composition but delivers a single set-level scalar shared by all its documents.

\subsection{Details of Generator and Search Agent}

\noindent \textbf{RAG Generator.} In the RAG scenario, the selected documents are consumed by \textit{Llama-3.1-8B-Instruct}, which conditions on the top-$k$ passages returned by each reranker and produces a short answer. A mid-scale instruction-tuned generator is deliberately chosen here: it is strong enough to exploit good evidence, yet not so strong that it can compensate for a poor set from parametric knowledge alone, so the differences we observe downstream can be attributed to the quality of the selected set rather than to the generator.

\noindent \textbf{Deep Research Agent.} In the deep research scenario, we build on \textit{DR.Tulu-8B}~\citep{Dr-tulu}, an open-source deep research agent initialized from Qwen3-8B and trained end-to-end with reinforcement learning against evolving rubrics. The agent runs an autonomous multi-turn loop that alternates between internal planning (\texttt{think}), tool invocation (\texttt{call\_tool}), and answer synthesis (\texttt{answer}), with web search, page browsing, and paper retrieval available as tools. We keep the agent fixed and swap only the reranker inside its retrieval pipeline, so that any change in the final report can be traced to how the evidence at each search step was composed.

\section{Implementation Details} \label{appendix:implementation_details}

\subsection{Details of Meta Rubrics}\label{appendix:meta_rubrics}

The meta rubrics are dimension-level templates rather than criteria themselves: for each query they are instantiated into concrete, checkable criteria, each carrying an importance weight. The nine dimensions are organized into three levels according to what they examine.

\noindent \textbf{Level 1: Document-level.} These dimensions examine each document in isolation and are rated per document and averaged over the set. \noindent \ding{182} \textbf{Relevance (Rel.)} asks whether a document addresses the information need actually expressed by the query, as opposed to merely sharing surface terms with it. \noindent \ding{183} \textbf{Authenticity (Aut.)} asks whether the factual claims a document makes are consistent with the reference answer, penalizing outdated, misattributed, or fabricated statements. It is the only dimension requiring the reference answer, and is therefore available during training but never at inference. \noindent \ding{184} \textbf{Quality (Qua.)} asks whether the target information can be extracted with little effort, rewarding clear structure and self-contained statements while penalizing fragmented or navigation-heavy text.

\noindent \textbf{Level 2: Set-level.} These dimensions examine how documents interact and are rated once for the whole set. None of them is a property of any single document, which is why a pointwise reranker cannot observe them. \noindent \ding{185} \textbf{Complementarity (Cmp.)} asks whether the members contribute distinct information elements, so that the set is worth more than any one of them. \noindent \ding{186} \textbf{Redundancy (Red.)} asks whether multiple documents convey the same information without incremental contribution. It is not the inverse of complementarity: a set may avoid redundancy yet still fail to be complementary, if its members are distinct but jointly uninformative. \noindent \ding{187} \textbf{Conflict (Con.)} asks whether documents make contradictory statements about the same fact, which forces the consumer to adjudicate between sources.

\noindent \textbf{Level 3: Global-level.} These dimensions evaluate the set in its role as the input context of a downstream model, and are likewise rated once for the whole set. \noindent \ding{188} \textbf{Completeness (Cpl.)} asks whether the set covers every key element a satisfactory answer requires. Unlike complementarity, which compares members against one another, completeness compares the set against the information need. \noindent \ding{189} \textbf{Density (Den.)} asks whether useful content dominates the context rather than irrelevant filler, thereby capturing the trade-off between broader coverage and a concise, undiluted input. \noindent \ding{190} \textbf{Reachability (Rea.)} asks whether a model with no external knowledge could arrive at the correct answer from this set alone, and is thus the dimension most directly aligned with the answer-level metrics.

\noindent \textbf{Instantiation.} Given the meta rubrics above, a query $q$, and its reference answer $a$, DeepSeek-V4-Pro instantiates one or more criteria under each dimension. Each criterion must be phrased as a question about the document set and must name specific entities, facts, or figures from $q$ and $a$; vague formulations such as ``contains relevant content'' are rejected. The resulting criteria and weights form $R_q$, which drives both the silver labels and the reward of Eq.~\ref{eq:utility}. Full prompts are in Appendix~\ref{appendix:prompt}.

\noindent \textbf{Inapplicable criteria.} Several dimensions presuppose a configuration that a given set may not exhibit: conflict needs two comparable claims about the same fact, and complementarity needs two relevant documents carrying different required points. Scoring such a criterion on a $0$--$10$ scale would reward a set for trivially ``not conflicting'' when nothing could conflict in the first place. The judge therefore returns \texttt{null} rather than a score whenever a criterion's applicability prerequisite is unmet, and every \texttt{null} criterion is dropped from both the numerator and the denominator of Eq.~\ref{eq:utility}, so the weights renormalize over the applicable criteria alone and an inapplicable dimension neither raises nor lowers the reward. Relevance is never \texttt{null}, and completeness and reachability take $0$ rather than \texttt{null} when the relevant subset covers nothing, so at least one criterion always survives and $u(\mathcal{S}\mid q)$ stays well defined.

\subsection{Details of Training Query}\label{appendix:train_query}

\begin{table*}[h!]
  \centering
  \caption{\textbf{Training Data Composition.} Number of queries used in each training
  stage, grouped by scenario. OSch: OpenScholar; SA$_\text{s}$/SA$_\text{l}$: SearchArena
  short-/long-form; Glaive: GlaiveAI-Reasoning-v1; WW: WebWalker.}
  \label{tab:train_data}
  \vspace{-6pt}

  \renewcommand{\arraystretch}{1.1}
  \resizebox{\textwidth}{!}{%
    \begin{tabular}{l|ccccc|cccccc|c}
      \toprule
      \multirow{2.5}{*}{\textbf{Stage}}
        & \multicolumn{5}{c|}{\textbf{RAG}}
        & \multicolumn{6}{c|}{\textbf{Deep Research}}
        & \multicolumn{1}{c}{\cellcolor{gray!12}} \\
      \noalign{\global\aboverulesep=0pt \global\belowrulesep=0pt}
      \cmidrule(l{2pt}r{2pt}){2-6} \cmidrule(l{2pt}r{2pt}){7-12}
      \noalign{\global\aboverulesep=.65ex \global\belowrulesep=.65ex}
        & \cellcolor{gray!12}\rule{0pt}{2.6ex}\textbf{Sum}
        & \textbf{HotpotQA}
        & \textbf{NQ}
        & \textbf{2Wiki}
        & \textbf{MuSiQue}
        & \cellcolor{gray!12}\textbf{Sum}
        & \textbf{OSch}
        & \textbf{SA$_\text{s}$}
        & \textbf{Glaive}
        & \textbf{WW}
        & \textbf{SA$_\text{l}$}
        & \cellcolor{gray!12}\raisebox{4.5pt}[0pt][0pt]{\textbf{Total}} \\[-0.3ex]
      \midrule

      SFT & \cellcolor{gray!12}3{,}433 & 1{,}644 & 1{,}209 & 432 & 148
          & \cellcolor{gray!12}5{,}525 & 2{,}439 & 1{,}176 & 831 & 739 & 340
          & \cellcolor{gray!12}8{,}958 \\

      ATO & \cellcolor{gray!12}1{,}065 & 511 & 426 & 83 & 45
          & \cellcolor{gray!12}2{,}492 & 1{,}252 & 527 & 276 & 306 & 131
          & \cellcolor{gray!12}3{,}557 \\

      \midrule

      \textbf{Total} & \cellcolor{gray!12}\textbf{4{,}498} & 2{,}155 & 1{,}635 & 515 & 193
          & \cellcolor{gray!12}\textbf{8{,}017} & 3{,}691 & 1{,}703 & 1{,}107 & 1{,}045 & 471
          & \cellcolor{gray!12}\textbf{12{,}515} \\

      \bottomrule
    \end{tabular}%
  }
\end{table*}

\vspace{8pt}

Training queries are drawn from both scenarios the reranker is meant to serve, since the two differ in what a query looks like.

\noindent \textbf{RAG queries.} We sample user questions from the training splits of four multi-hop question answering datasets, HotpotQA~\citep{HotpotQA}, Natural Questions~\citep{nq}, 2WikiMultihopQA~\citep{2wikimultihopqa}, and MuSiQue~\citep{MuSiQue}. These questions are short and closed-ended, and each comes with a gold answer, which is required because our rubric generation conditions on the reference answer.

\noindent \textbf{Deep research queries.} In this scenario a reranker is invoked on the sub-queries an agent issues to itself, not on the original user question, and such sub-queries are rarely released with open-source datasets. Rather than regenerate them with an agent of our own, which would tie the training distribution to one particular agent, we reuse the agent-issued sub-queries released by RubricRanker~\citep{rubricranker}. They originate from OpenScholar~\citep{openscholar}, SearchArena~\citep{searcharena}, GlaiveAI-Reasoning-v1-20M, and WebWalker-Silver~\citep{webwalkerqa}, and therefore span scientific, general-purpose, reasoning-oriented, and web-navigation information needs.

\noindent \textbf{Candidate pools and filtering.} For every retained query we build a candidate pool with the retriever of the corresponding scenario, and sample the pool size from $10$ to $40$ per query so that the policy is exposed to varying numbers of candidates. Queries for which rubric generation fails to produce a well-formed set of weighted criteria are discarded. Table~\ref{tab:train_data} reports the resulting number of queries per source for the cold-start and the adaptive tutoring optimization stages.

Combining the two sources yields queries that differ in form, length, and difficulty, which in turn yields more diverse rubrics and, empirically, better generalization across the two scenarios.

\subsection{Cold-Start SFT Stage}\label{appendix:sft}

\noindent \textbf{Silver label construction.} Silver labels are produced by DeepSeek-V4-Pro, which is given the query, the candidate pool, and the query-specific rubrics, and returns the identifiers of the documents to retain. The two scenarios use different prompt templates, since an agent-issued sub-query carries an accompanying query intent that a user question does not.

\noindent \textbf{Fine-tuning setup.} We fine-tune Qwen3-8B~\citep{Qwen3} on these labels with LlamaFactory~\citep{llamafactory} on 8 NVIDIA H20 GPUs. The model input follows the same scenario-specific format used for label construction, comprising the query and the candidate documents, plus the query intent in the deep research case, but excluding the query-specific rubrics: they depend on a reference answer that is unavailable at inference, so withholding them keeps the training and inference inputs identical. Optimization uses a learning rate of \texttt{5e-6}, a per-device batch size of $1$, and $8$ gradient accumulation steps, giving an effective batch size of $64$, for one epoch under BF16 mixed precision. Throughput is improved with DeepSpeed ZeRO-2~\citep{deepspeed} and FlashAttention-2~\citep{flashattention}.

\subsection{Adaptive Tutoring Optimization Stage}\label{appendix:ato}

\noindent \textbf{Implementation.} The second stage is implemented in the VERL framework on top of its on-policy self-distillation recipe~\citep{OPSD}, which we extend from a single fixed privileged context to the reward-conditioned hint space of Eq.~\ref{eq:hint}. Rollouts are generated with SGLang under a tensor-parallel degree of $8$. Each update combines the outcome-driven GRPO advantage with the token-level distillation advantage of Eq.~\ref{eq:adv_atd} in the convex form of Eq.~\ref{eq:convex}, so that a single policy-gradient step is taken on the combined advantage. The two terms are not on a common scale: the GRPO advantage is standardized within each rollout group, whereas the distillation advantage is a raw log-probability gap. We therefore read $\lambda$ as a mixing weight tuned empirically, not as a literal ratio of gradient contributions, and select it by the ablation in Table~\ref{tab:ablation} rather than by matching magnitudes. Following the reverse-KL instantiation of on-policy self-distillation, the distillation advantage is computed as the log-probability gap between the hint-conditioned teacher and the rollout snapshot on the sampled tokens, with the teacher distribution truncated to its top $128$ logits. The teacher is held at the cold-start checkpoint $\theta_{\mathrm{sft}}$ throughout the stage and is never updated, which also supplies the implicit anchor that lets us disable the explicit KL penalty. Rollouts whose output violates the required identifier format are excluded from the distillation branch and supervised by the outcome reward alone, so that malformed sequences are never used as distillation targets.

\IncMargin{2em}
\begin{algorithm}[t]
\DontPrintSemicolon
\SetAlgoNoLine
\caption{AdaTutoRank: Adaptive Tutoring Optimization}
\label{alg:ato}
\Require{Cold-start checkpoint $\theta_{\mathrm{sft}}$; query set $\mathcal{Q}$ with candidate pools $\mathcal{C}$ and rubrics $R_q$; judge $J$; self-selector $\mathcal{S}$; self-reflector $\mathcal{F}$; thresholds $\tau_1 > \tau_2$; group size $G$; convex weight $\lambda$; clip parameter $\epsilon$; learning rate $\eta$}
\Ensure{Hint-free reranking policy $\pi_\theta$}
$\theta \leftarrow \theta_{\mathrm{sft}}$, \ $\phi \leftarrow \theta_{\mathrm{sft}}$ \tcp*{teacher $\pi_\phi$ stays frozen throughout}
\For{each policy update}{
  $\theta_{\mathrm{old}} \leftarrow \theta$; \ sample a query batch $\mathcal{B} \subset \mathcal{Q}$\;
  \tcc{On-policy rollouts, rubric rewards, and outcome advantages}
  \ForEach{query $q \in \mathcal{B}$}{
    Sample $\mathcal{G}_q = \{y_i\}_{i=1}^{G}$ with $y_i \sim \pi_{\theta_{\mathrm{old}}}(\cdot \mid q, \mathcal{C})$\;
    \ForEach{rollout $y_i \in \mathcal{G}_q$}{
      \uIf{$y_i$ violates the identifier format}{
        $r_i \leftarrow -1$, \ $m_i \leftarrow 0$ \tcp*{excluded from distillation}
      }
      \Else{
        $\mathcal{S}_i \leftarrow \mathrm{Parse}(y_i)$, \ $r_i \leftarrow u(\mathcal{S}_i \mid q)$ via $J$ and $R_q$, \ $m_i \leftarrow 1$\;
      }
    }
    $A^{\mathrm{GRPO}}_i \leftarrow \big(r_i - \mathrm{mean}(\{r_j\})\big) / \mathrm{std}(\{r_j\})$ for all $i$\;
    \tcc{Sibling set from the self-selector, scored by the same judge}
    $\bar{h}_q \leftarrow \mathcal{S}\big(q, \mathcal{C}, R_q; \pi_{\theta_{\mathrm{old}}}\big)$, \ $r_{\bar h} \leftarrow u(\bar{h}_q \mid q)$\;
    \tcc{Reward-conditioned hint assignment with admission gate}
    \ForEach{rollout $y_i \in \mathcal{G}_q$}{
      \uIf{$r_i \geq \tau_1$}{
        $\bm{h}_i \leftarrow h^{\ast} = R_q$ \tcp*{strong rollout: criteria only}
      }
      \uElseIf{$r_{\bar h} \leq r_i$}{
        $\bm{h}_i \leftarrow h^{\ast} = R_q$ \tcp*{gate rejects: fall back}
      }
      \uElseIf{$\tau_2 < r_i < \tau_1$}{
        $\bm{h}_i \leftarrow \hat{h}_i = \mathcal{F}\big(q, \mathcal{C}, y_i, \bar{h}_q; \pi_{\theta_{\mathrm{old}}}\big)$ \tcp*{corrective reflection}
      }
      \Else{
        $\bm{h}_i \leftarrow \bar{h}_q$ \tcp*{weak rollout ($r_i \le \tau_2$): sibling set}
      }
    }
  }
  \tcc{Paired re-scoring and joint optimization}
  \ForEach{minibatch of sampled tokens $(i, \ell)$}{
    $A^{\mathrm{ATD}}_{i,\ell} \leftarrow \sg\!\big[\log \pi_\phi(\hat y_{i,\ell} \mid q, \mathcal{C}, \bm{h}_i, y_{i,<\ell}) - \log \pi_{\theta_{\mathrm{old}}}(\hat y_{i,\ell} \mid q, \mathcal{C}, y_{i,<\ell})\big]$ \tcp*{no regeneration}\;
    $A_{i,\ell} \leftarrow \lambda A^{\mathrm{GRPO}}_i + (1-\lambda)\, m_i A^{\mathrm{ATD}}_{i,\ell}$\;
    $\rho_{i,\ell}(\theta) \leftarrow \pi_\theta(\hat y_{i,\ell} \mid q, \mathcal{C}, y_{i,<\ell}) \big/ \pi_{\theta_{\mathrm{old}}}(\hat y_{i,\ell} \mid q, \mathcal{C}, y_{i,<\ell})$\;
    $\mathcal{L}(\theta) \leftarrow -\mathbb{E}_{i,\ell}\big[\min\big(\rho_{i,\ell} A_{i,\ell}, \ \mathrm{clip}(\rho_{i,\ell}, 1-\epsilon, 1+\epsilon) A_{i,\ell}\big)\big]$\;
    $\theta \leftarrow \theta - \eta \nabla_\theta \mathcal{L}(\theta)$ \tcp*{no explicit KL; $\pi_\phi$ anchors}
  }
}
\end{algorithm}
\DecMargin{2em}

\noindent \textbf{Hint assignment.} Each rollout is routed to one of three hints by its rubric reward $r_i$, with bucket boundaries at $\tau_2 = 3$ and $\tau_1 = 6$ on the judge's $0$--$10$ scale: rollouts scoring at least $6$ receive the rubrics, those in $(3, 6)$ receive a corrective reflection, and those scoring at most $3$ receive the sibling set. The two hints that build on the sibling set are additionally gated in relative mode, i.e. they are admitted only when the sibling set outscores the rollout it supervises ($r_{\bar h} > r_i$, with zero margin), and otherwise the rollout falls back to the rubric hint. 

\noindent \textbf{Reward service.} Rubric rewards are produced by a locally deployed DeepSeek-V4-Flash judge, served in FP8 with SGLang behind an internal endpoint, which avoids the cost and rate limits of a commercial API and keeps the reward model fixed across all runs. To reduce the number of judge calls, all set- and global-level criteria of a rollout are scored in a single request that returns one score per criterion, and document-level criteria are likewise batched per document. Requests are issued by a parallel reward manager with $512$ concurrent workers.

\noindent \textbf{Hyperparameters.} Training runs on $32$ NVIDIA H20 GPUs ($4$ nodes $\times$ $8$ GPUs). We use a training batch of $64$ prompts per step with a PPO mini-batch of $64$ and $8$ rollouts per prompt, a learning rate of \texttt{5e-6} with $10$ warmup steps and a cosine schedule decaying to $0.1$ of its peak, a maximum prompt length of $16{,}000$ tokens, and the interpolation coefficient $\lambda = 0.7$ selected by the ablation in Table~\ref{tab:ablation}. The maximum response length is $64$ tokens, since the policy emits the bracketed identifier sequence of Eq.~\ref{eq:reward} directly, with no reasoning trace before the selection. Queries are shuffled once per epoch and drawn without replacement, and the trailing incomplete batch is dropped, so one epoch amounts to $\lfloor 3{,}557/64\rfloor=55$ steps over the ATO queries of Table~\ref{tab:train_data}, leaving $37$ queries unused in that epoch. Validation runs every $5$ steps and checkpointing every $15$. All results we report come from the checkpoint at step $30$, which attains the best validation reward; it has therefore been optimized on $30\times64=1{,}920$ distinct queries, slightly over half of a single epoch, so the reported policy is obtained well before the training queries are exhausted.

\begin{table}[t]
\centering
\small
\caption{Representative adaptive tutoring hints for two training queries, one from the RAG
pool and one from the deep research pool. For each query we show three rollouts of the same
policy that fall into different reward bands. $\hat{y}_i$ is the subset selected by that
rollout itself, and $r_i$ its judge reward; the hint is chosen according to $r_i$ and is
shown only to the teacher. The rubric hint states the criteria only, the corrective
reflection names the documents to add or drop, and the sibling set gives a concrete
reference subset.}
\label{tab:hint_examples}
\renewcommand{\arraystretch}{1.15}
\begin{tabular}{@{}p{1.9cm}p{0.6cm}p{1.3cm}p{\dimexpr\textwidth-3.8cm-6\tabcolsep\relax}@{}}
\toprule
\textbf{Hint type} & \textbf{$r_i$} & \textbf{Rollout $\hat{y}_i$} & \textbf{Privileged information to the teacher} \\
\midrule
\multicolumn{4}{@{}p{\textwidth}@{}}{\textit{\textbf{RAG.}} \ Do both films
\textit{The Garden of Eden} (1998) and \textit{One Night in Yoshiwara} have directors
that share the same nationality?} \\
\addlinespace[3pt]
Rubrics & $9.25$ & [2] [14] &
\textit{\textup{[doc-level]} Relevance (weight 5): does each document directly discuss the
director (by name) and nationality of The Garden of Eden (1998) and/or One Night in Yoshiwara?}
\newline
\textit{\textup{[set-level]} Complementarity (weight 5): does the set contain at least one
document identifying the director and nationality of each of the two films, so that together
they enable the comparison?}
\newline
\textit{$\dots$ seven further dimensions across the three tiers.} \\
\addlinespace[3pt]
Corrective reflection & $4.00$ & [1] [2] &
\textit{The chosen documents [1] and [2] are insufficient. [1] is irrelevant, and [2] only
covers The Garden of Eden, giving its director Alessandro D'Alatri and his Italian nationality.
Document [14] supplies the director of One Night in Yoshiwara, Max Oph\"uls, and his French
nationality. Only [2] and [14] are needed.} \\
\addlinespace[3pt]
Sibling-set & $3.67$ & [1] [2] &
\textit{[2] [14]} \ (scored $9.25$ by the judge) \\
\midrule
\multicolumn{4}{@{}p{\textwidth}@{}}{\textit{\textbf{Deep Research.}} \ Illuvium Pre-season 2
rewards 10k ILV} \\
\addlinespace[3pt]
Rubrics & $8.84$ & [1] [3] &
\textit{\textup{[doc-level]} Relevance (weight 5): does each document address the Pre-season 2
reward allocation in terms of the total ILV amount, rather than merely mentioning Illuvium or
Pre-season 2 without the pool size?}
\newline
\textit{\textup{[set-level]} Conflict (weight 5): are all statements about the Pre-season 2 ILV
amount mutually consistent, with no document claiming a different total (e.g.\ 5,000 or 15,000
ILV)?}
\newline
\textit{$\dots$ seven further dimensions across the three tiers.} \\
\addlinespace[3pt]
Corrective reflection & $4.83$ & [1] [6] &
\textit{Document [3] should be included: it states that ``an additional 10,000 ILV awaits in
Pre-Season 2'' and comes from the official Illuvium account. Document [6] should be excluded:
it refers to 10,000 ILV added to an existing 20,000 ILV pool, which is ambiguous and conflicts
with the figure in [3].} \\
\addlinespace[3pt]
Sibling-set & $0.00$ & [2] [6] &
\textit{[1] [3]} \ (scored $9.32$ by the judge) \\
\bottomrule
\end{tabular}
\end{table}

\subsection{Algorithm and Hint Examples}\label{app:algorithm_hints}
Algorithm~\ref{alg:ato} summarizes the adaptive tutoring loop. At each update, the frozen snapshot $\pi_{\theta_{\mathrm{old}}}$ serves three roles: it samples the rollout group for every query, it assembles the sibling set under the query-specific rubrics, and it contrasts each rollout against that sibling set to produce a corrective reflection. The rubric judge then scores both the rollouts and the sibling set, and each rollout is routed to one of the three hints by its own reward, falling back to the rubrics whenever the sibling set fails the admission gate. The hint-conditioned teacher $\pi_\phi$, held at the cold-start checkpoint $\theta_{\mathrm{sft}}$, re-scores the sampled tokens under the hint-augmented context, while the same snapshot re-scores them under the ordinary context, so the resulting advantage is constant in $\theta$ and gradients reach the policy only through the importance ratio of Eq.~\ref{eq:ratio}. The resulting ATD advantage is combined convexly with the GRPO advantage and optimized in a single clipped policy-gradient step, after which the updated policy becomes the next snapshot.

Table~\ref{tab:hint_examples} provides representative hints of the three types, drawn from rollouts in the high, middle, and low reward bands of the same queries.

\begin{table*}[t!]
  \centering
  \caption{\textbf{Setwise-Level Performance Comparison on SetwiseEvalKit (Long-form Scenario).}
The \colorbox{backred!50}{\textbf{best}} (1\textsuperscript{st}) and \colorbox{backyellow_soft!40}{\underline{second-best}} (2\textsuperscript{nd}) results are highlighted.
The per-level \textbf{Avg} is the arithmetic mean of the three dimensions within that level, and \textbf{Overall} is the mean of all nine dimensions.
All metrics are reported such that higher is better ($\uparrow$).}
  \label{tab:setwise_level_long}
  \vspace{-6pt}
  \renewcommand{\arraystretch}{1.1}
  \resizebox{\textwidth}{!}{%
    \begin{tabular}{l|cccc|cccc|cccc|c}
      \toprule
      \multirow{2.5}{*}{\textbf{Ranker}}
        & \multicolumn{4}{c|}{\textbf{Doc-Level}}
        & \multicolumn{4}{c|}{\textbf{Set-Level}}
        & \multicolumn{4}{c|}{\textbf{Global-Level}}
        & \cellcolor{gray!12} \\
      \noalign{\global\aboverulesep=0pt \global\belowrulesep=0pt}
      \cmidrule(l{2pt}r{2pt}){2-5} \cmidrule(l{2pt}r{2pt}){6-9} \cmidrule(l{2pt}r{2pt}){10-13}
      \noalign{\global\aboverulesep=.65ex \global\belowrulesep=.65ex}
        &
        \cellcolor{gray!12}\textbf{Avg} & \textbf{Rel.} & \textbf{Aut.} & \textbf{Qua.}
        & \cellcolor{gray!12}\textbf{Avg} & \textbf{Cmp.} & \textbf{Red.} & \textbf{Con.}
        & \cellcolor{gray!12}\textbf{Avg} & \textbf{Cpl.} & \textbf{Den.} & \textbf{Rea.}
        & \cellcolor{gray!12}\raisebox{4.5pt}[0pt][0pt]{\textbf{Overall}} \\
      \midrule
Initial Retrieval& \cellcolor{gray!12}20.20 & 29.28 & 11.38 & 19.93 & \cellcolor{gray!12}\colorbox{backyellow_soft!40}{\underline{64.52}} & 31.90 & \colorbox{backyellow_soft!40}{\underline{71.06}} & 90.61 & \cellcolor{gray!12}23.51 & 22.26 & 31.93 & 16.35 & \cellcolor{gray!12}36.08 \\
BGE-Reranker$_{L.}$& \cellcolor{gray!12}21.57 & 31.66 & 12.76 & 20.30 & \cellcolor{gray!12}62.10 & 33.33 & 65.09 & 87.87 & \cellcolor{gray!12}25.27 & 23.48 & 33.96 & 18.37 & \cellcolor{gray!12}36.31 \\
MonoT5 {\small(3B)}& \cellcolor{gray!12}21.76 & 32.23 & 12.57 & 20.47 & \cellcolor{gray!12}63.85 & 34.70 & 69.61 & 87.25 & \cellcolor{gray!12}25.11 & 24.89 & 32.80 & 17.64 & \cellcolor{gray!12}36.91 \\
RankT5 {\small(3B)}& \cellcolor{gray!12}20.01 & 29.85 & 11.34 & 18.84 & \cellcolor{gray!12}62.31 & 30.83 & 69.62 & 86.49 & \cellcolor{gray!12}23.52 & 22.06 & 32.54 & 15.97 & \cellcolor{gray!12}35.28 \\
RankLlama {\small(7B)}& \cellcolor{gray!12}20.96 & 29.95 & 13.32 & 19.62 & \cellcolor{gray!12}62.69 & 32.53 & 64.76 & \colorbox{backyellow_soft!40}{\underline{90.77}} & \cellcolor{gray!12}23.93 & 22.58 & 31.38 & 17.82 & \cellcolor{gray!12}35.86 \\
{Rearank {\small(7B)}}& \cellcolor{gray!12}22.50 & 32.63 & 13.24 & 21.64 & \cellcolor{gray!12}63.64 & 34.37 & 67.53 & 89.02 & \cellcolor{gray!12}25.41 & 24.41 & 33.64 & 18.18 & \cellcolor{gray!12}37.18 \\
{ReasonRank {\small(7B)}}& \cellcolor{gray!12}22.95 & 33.94 & 13.39 & 21.52 & \cellcolor{gray!12}\colorbox{backred!50}{\textbf{64.82}} & 36.94 & 67.98 & 89.55 & \cellcolor{gray!12}26.80 & 25.92 & 35.34 & 19.14 & \cellcolor{gray!12}38.19 \\
{Rank4Gen {\small(8B)}}& \cellcolor{gray!12}23.57 & \colorbox{backyellow_soft!40}{\underline{35.09}} & 13.45 & 22.18 & \cellcolor{gray!12}62.54 & 28.00 & \colorbox{backred!50}{\textbf{74.64}} & 84.98 & \cellcolor{gray!12}23.45 & 20.34 & \colorbox{backyellow_soft!40}{\underline{36.48}} & 13.53 & \cellcolor{gray!12}36.52 \\
{SetR {\small(8B)}}& \cellcolor{gray!12}21.13 & 30.46 & 12.45 & 20.48 & \cellcolor{gray!12}63.31 & 32.53 & 65.73 & \colorbox{backred!50}{\textbf{91.68}} & \cellcolor{gray!12}24.10 & 24.38 & 29.86 & 18.06 & \cellcolor{gray!12}36.18 \\
{RubricRanker {\small(8B)}}& \cellcolor{gray!12}\colorbox{backyellow_soft!40}{\underline{29.70}} & 34.42 & \colorbox{backyellow_soft!40}{\underline{23.90}} & \colorbox{backyellow_soft!40}{\underline{30.78}} & \cellcolor{gray!12}60.26 & \colorbox{backred!50}{\textbf{41.65}} & 53.75 & 85.39 & \cellcolor{gray!12}\colorbox{backred!50}{\textbf{33.66}} & \colorbox{backred!50}{\textbf{35.13}} & 33.57 & \colorbox{backred!50}{\textbf{32.27}} & \cellcolor{gray!12}\colorbox{backyellow_soft!40}{\underline{41.21}} \\
      \midrule
{AdaTutoRank {\small(8B)}}& \cellcolor{gray!12}\colorbox{backred!50}{\textbf{34.10}} & \colorbox{backred!50}{\textbf{40.97}} & \colorbox{backred!50}{\textbf{26.00}} & \colorbox{backred!50}{\textbf{35.32}} & \cellcolor{gray!12}63.27 & \colorbox{backyellow_soft!40}{\underline{39.09}} & 62.78 & 87.94 & \cellcolor{gray!12}\colorbox{backyellow_soft!40}{\underline{33.17}} & \colorbox{backyellow_soft!40}{\underline{32.00}} & \colorbox{backred!50}{\textbf{37.71}} & \colorbox{backyellow_soft!40}{\underline{29.79}} & \cellcolor{gray!12}\colorbox{backred!50}{\textbf{43.51}} \\
      \bottomrule
    \end{tabular}%
  }
\end{table*}

\begin{table*}[t!]
  \centering
    \caption{Ablation Study of AdaTutoRank on the RAG and Deep Research Benchmarks.}

  \label{tab:ablation_all}

  \vspace{-6pt}
  \renewcommand{\arraystretch}{1.1}
  \resizebox{\textwidth}{!}{%
    \begin{tabular}{l|ccccc|cccc|c}
      \toprule
      \multirow{2.5}{*}{\textbf{Setting}}
        & \multicolumn{5}{c|}{\textbf{RAG}}
        & \multicolumn{4}{c|}{\textbf{Deep Research}}
        & \multicolumn{1}{c}{\cellcolor{gray!12}} \\
      \noalign{\global\aboverulesep=0pt \global\belowrulesep=0pt}
      \cmidrule(l{2pt}r{2pt}){2-6} \cmidrule(l{2pt}r{2pt}){7-10}
      \noalign{\global\aboverulesep=.65ex \global\belowrulesep=.65ex}
        & \rule{0pt}{2.6ex}\textbf{NQ}
        & \textbf{Triv}
        & \textbf{PopQA}
        & \textbf{2Wiki}
        & \textbf{Bambo}
        & \textbf{WebW}
        & \textbf{HealthB}
        & \textbf{DRB}
        & \textbf{RQA}
        & \cellcolor{gray!12}\raisebox{4.5pt}[0pt][0pt]{\textbf{Overall}} \\[-0.3ex]
      \midrule

      AdaTutoRank & 36.87 & 66.54 & 38.52 & 32.52 & 20.80 & 41.00 & 48.57 & 49.89 & 72.78 & \cellcolor{gray!12}\colorbox{backred!50}{\textbf{45.28}} \\

      \midrule
      \rowcolor{gray!10}
      \multicolumn{11}{c}{\fontsize{10}{12}\selectfont \textit{\textbf{Training Stage}}} \\

      w/o ATO  & 36.45 & 66.72 & 38.25 & 33.25 & 20.80 & 38.00 & 40.96 & 48.61 & 69.67 & \cellcolor{gray!12}43.63 \\
      w/o SFT                 & 34.79 & 66.14 & 38.45 & 33.23 & 14.40 & 37.00 & 45.24 & 48.92 & 69.67 & \cellcolor{gray!12}43.09 \\

      \midrule
      \rowcolor{gray!10}
      \multicolumn{11}{c}{\fontsize{10}{12}\selectfont \textit{\textbf{Advantage Superposition}}} \\

      Only RL             & 31.88 & 64.28 & 36.23 & 26.49 & 17.60 & 43.00 & 46.11 & 48.64 & 66.06 & \cellcolor{gray!12}42.25 \\
    Only Distillation   & 36.43 & 66.34 & 38.39 & 32.85 & 16.80 & 46.00 & 46.38 & 49.18 & 69.95 & \cellcolor{gray!12}44.70 \\

      \midrule
      \rowcolor{gray!10}
      \multicolumn{11}{c}{\fontsize{10}{12}\selectfont \textit{\textbf{Interpolation Coefficient}}} \\

      $\lambda = 0.9$     & 33.16 & 65.86 & 36.17 & 28.94 & 18.40 & 41.00 & 47.42 & 48.40 & 69.71 & \cellcolor{gray!12}43.23 \\
      $\lambda = 0.5$ & 36.70 & 66.74 & 38.47 & 32.74 & 17.60 & 39.00 & 45.82 & 48.74 & 70.00 & \cellcolor{gray!12}43.98 \\

      \midrule
      \rowcolor{gray!10}
      \multicolumn{11}{c}{\fontsize{10}{12}\selectfont \textit{\textbf{Privileged Information}}} \\

      Only Rubrics $h^{*}$             & 36.51 & 66.86  & 38.43 & 32.28 & 16.00 & 39.00 & 47.20 & 48.41 & 67.78 & \cellcolor{gray!12}43.61 \\
      Only Reflection $\hat{h}$       & 35.60 & 66.30  & 37.77 & 31.35 & 21.60 & 47.00 & 47.36 & 48.74 & 69.77 & \cellcolor{gray!12}\colorbox{backyellow_soft!40}{\underline{45.05}} \\
    Only Sibling-Set $\bar{h}$   & 35.82  & 66.65  & 38.22  & 32.40 & 16.00 & 43.00 & 43.11 & 47.72 & 70.36 & \cellcolor{gray!12}43.70 \\
      \bottomrule
    \end{tabular}%
  }
\end{table*}

\section{More Experimental Results} \label{appendix:experimental_results}

Beyond the results reported in the main text, this section supplements four groups of experiments. Appendix~\ref{appendix:setwise_long} extends the setwise-level evaluation from the short-form to the long-form scenario, verifying that the gains in evidence quality are not specific to short closed-ended queries. Appendix~\ref{appendix:ablation_all} gives the per-benchmark breakdown of all four ablations, whose overall scores are averaged over the nine answer-level benchmarks rather than the subset shown in the main text. Appendix~\ref{appendix:scaling} completes the two analyses of Section~\ref{sec:analysis} on the benchmarks omitted there, covering candidate pool sizes on NQ and PopQA and search behavior on HealthBench and DRB. Appendix~\ref{appendix:hint_dynamics} then turns from outcomes to the training process itself, tracking how the three hint types are distributed, how often the admission gate admits them, and how strong the sibling-set reference remains as the policy improves, so as to verify that the adaptive assignment is genuinely exercised rather than collapsing into a single form.

\subsection{Setwise-Level Evaluation}\label{appendix:setwise_long}

Table~\ref{tab:setwise_level_long} extends the setwise evaluation to the long-form scenario, where the same pattern holds. AdaTutoRank attains the best overall score of $43.51$, ahead of the strongest baseline RubricRanker by $2.30$ points. It leads on every document-level metric, raising the document-level average from $29.70$ to $34.10$, and also improves the set-level average over RubricRanker ($63.27$ vs.\ $60.26$), so the gains over the closest baseline appear at both the level of individual documents and the level of their interaction. At the global level it leads on density ($37.71$), indicating that the selected context stays informative rather than padded, but trails RubricRanker on completeness ($32.00$ vs.\ $35.13$) and reachability ($29.79$ vs.\ $32.27$), and hence on the global-level average ($33.17$ vs.\ $33.66$); its long-form advantage therefore comes from document- and set-level composition rather than from global coverage. Taken together, the improvement comes from how documents are chosen and combined rather than from retaining more of them.


\begin{figure}[t]
\centering
\begin{minipage}[b]{0.49\textwidth}
  \centering
  \includegraphics[width=\linewidth]{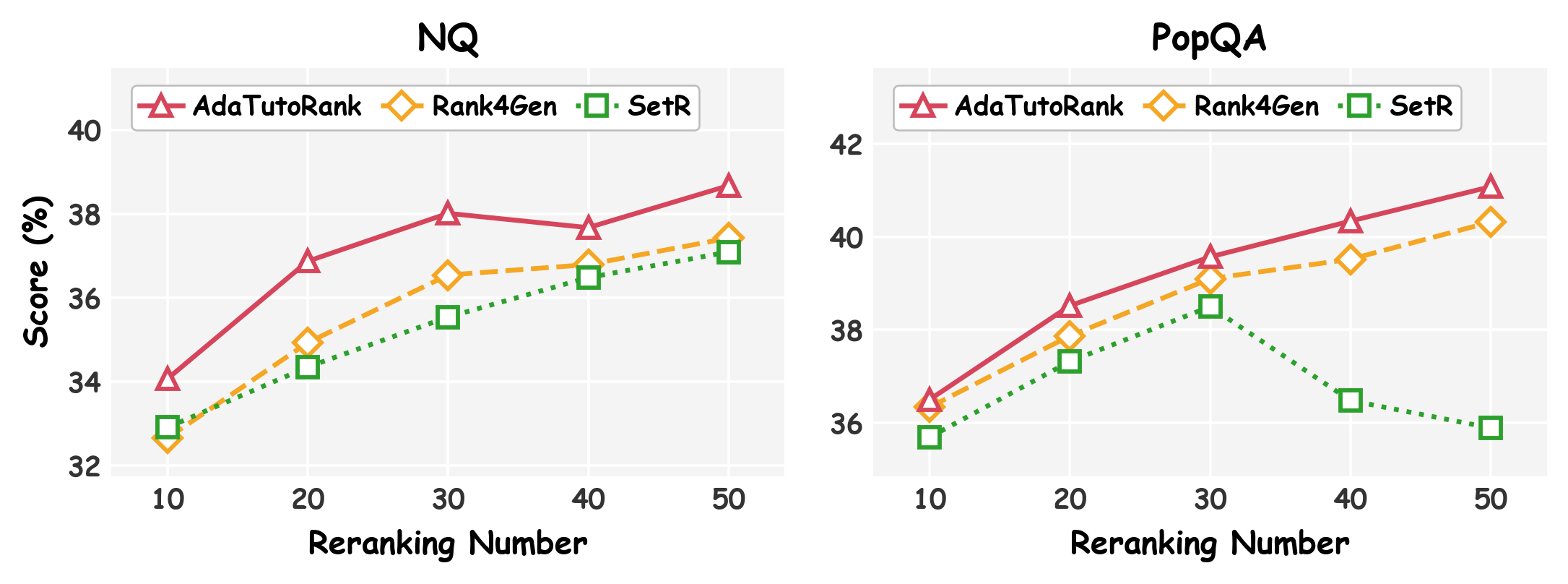}
\end{minipage}
\hfill
\begin{minipage}[b]{0.50\textwidth}
  \centering
  \includegraphics[width=\linewidth]{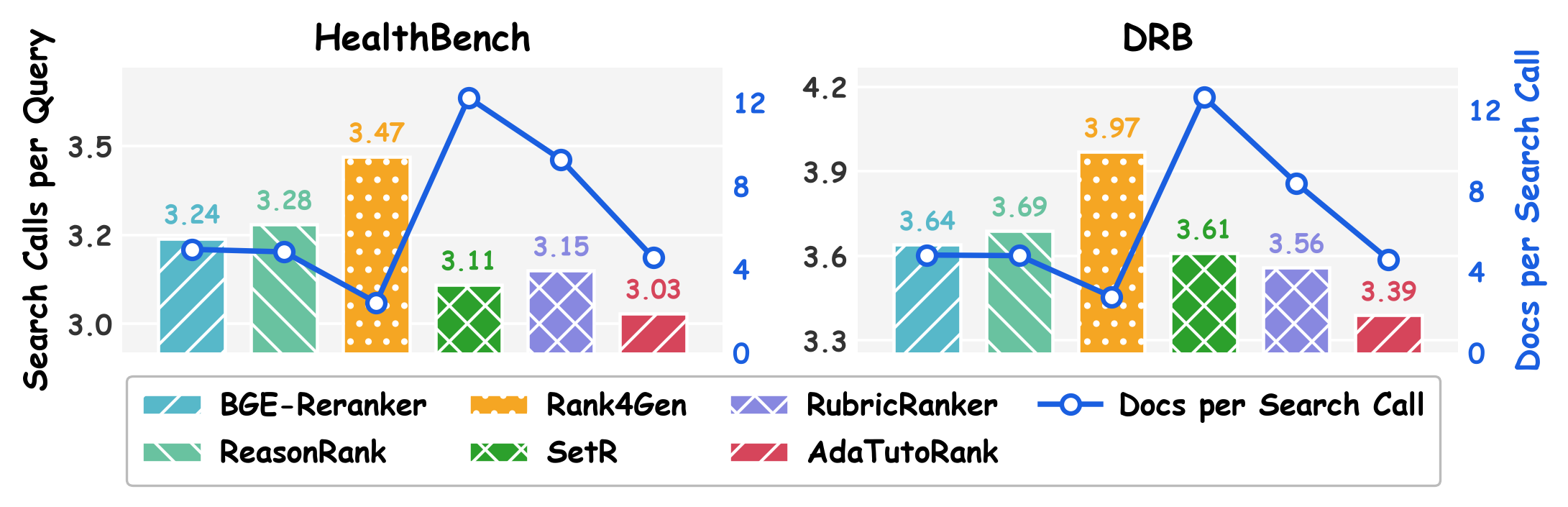}
\end{minipage}
\caption{\textbf{Left:} performance across candidate pool sizes on NQ and PopQA. \textbf{Right:} search calls per query and documents per search call on HealthBench and DRB.}
\label{fig:doc_count_search_call}
\end{figure}

\subsection{Complete Ablation Experiment Results}\label{appendix:ablation_all}

\noindent Table~\ref{tab:ablation_all} reports the per-benchmark breakdown of all four ablations, averaged over the nine benchmarks. Removing either training stage costs a comparable amount, $1.65$ points without ATO and $2.19$ without the cold start, so neither stage is dispensable; the cold start matters most on the short-form benchmarks, where its removal drops Bamboogle from $20.80$ to $14.40$, consistent with its role as a stable initialization for the format the policy must emit. Bamboogle alone contributes $0.71$ of that $2.19$, and over the remaining eight benchmarks the two stages cost $1.85$ and $1.66$ respectively, matching the ordering obtained on the four-benchmark subset of Table~\ref{tab:ablation}.

For the advantage design, the outcome signal alone is the weakest configuration overall ($42.25$) and degrades most sharply on the multi-hop sets, falling from $32.52$ to $26.49$ on 2Wiki, whereas the distillation signal alone already reaches $44.70$; superposing the two recovers the best score on both scenario groups. The interpolation coefficient degrades on both sides of $\lambda = 0.7$, and each end drifts toward the corresponding single-signal regime, $43.23$ at $\lambda = 0.9$ against $42.25$ for the outcome signal alone and $43.98$ at $\lambda = 0.5$ against $44.70$ for the distillation signal alone. Finally, every fixed hint form underperforms the adaptive assignment, and the breakdown shows why no single form suffices: among the three fixed-hint variants, rubrics alone are the strongest on NQ ($36.51$) yet tie with the sibling set for the weakest on Bamboogle ($16.00$), while reflections alone reverse this ordering ($35.60$ on NQ against $21.60$ on Bamboogle), so the form that helps one query type is not the form that helps another.

\subsection{Varying Number of Candidate Documents and Search Call Rounds}\label{appendix:scaling}

\noindent Figure~\ref{fig:doc_count_search_call} reports the two analyses on the remaining benchmarks. On NQ and PopQA, AdaTutoRank stays the strongest reranker at every pool size and improves monotonically as the pool grows from $10$ to $50$, whereas SetR peaks at $30$ candidates on PopQA and then declines, indicating that a larger pool benefits our policy rather than distracting it. Since the training pools are sampled from $10$ to $40$ (Appendix~\ref{appendix:train_query}), the continued gain at $50$ lies outside the range the policy was trained on, so set selection transfers to pools larger than any it saw during training rather than being tuned to a particular pool size. On HealthBench and DRB, it issues the fewest search calls per query ($3.03$ and $3.39$) while also returning the fewest documents per call, so the agent reaches its answer with both fewer rounds and a smaller context.

\begin{figure*}[t]
\centering
\begin{minipage}[t]{0.32\textwidth}
  \centering
  \includegraphics[width=\linewidth]{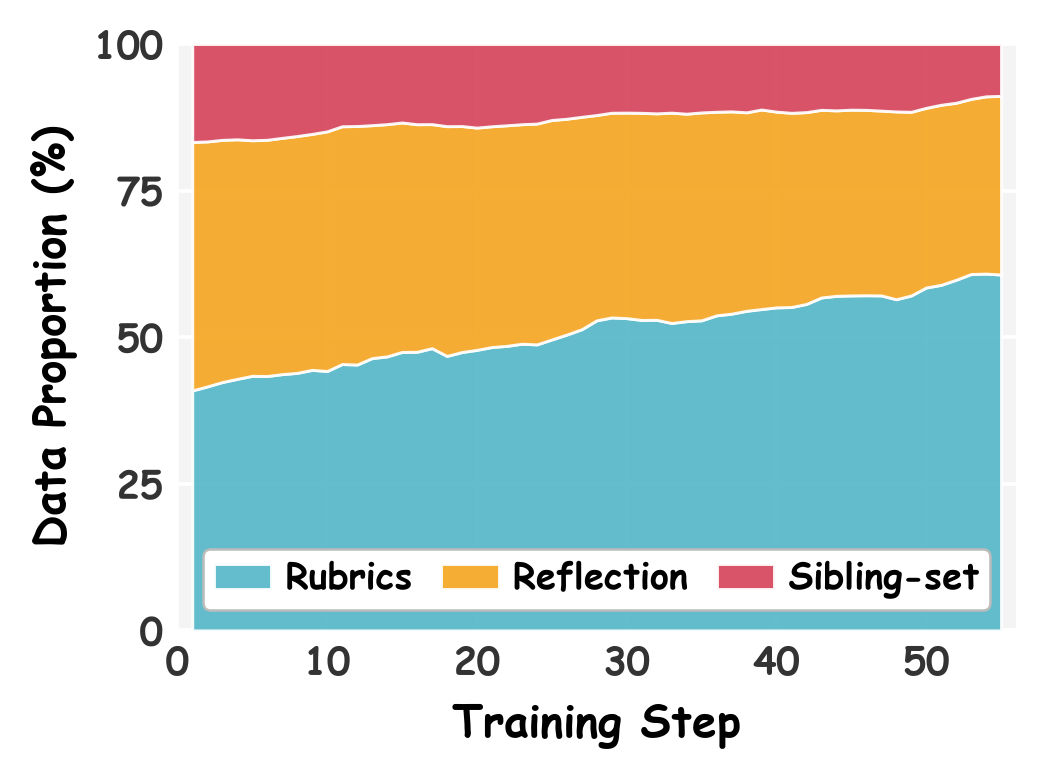}
\captionof{figure}{Proportion of rollouts per hint type over training.}
  \label{fig:hint_shares}
\end{minipage}\hfill
\begin{minipage}[t]{0.32\textwidth}
  \centering
  \includegraphics[width=\linewidth]{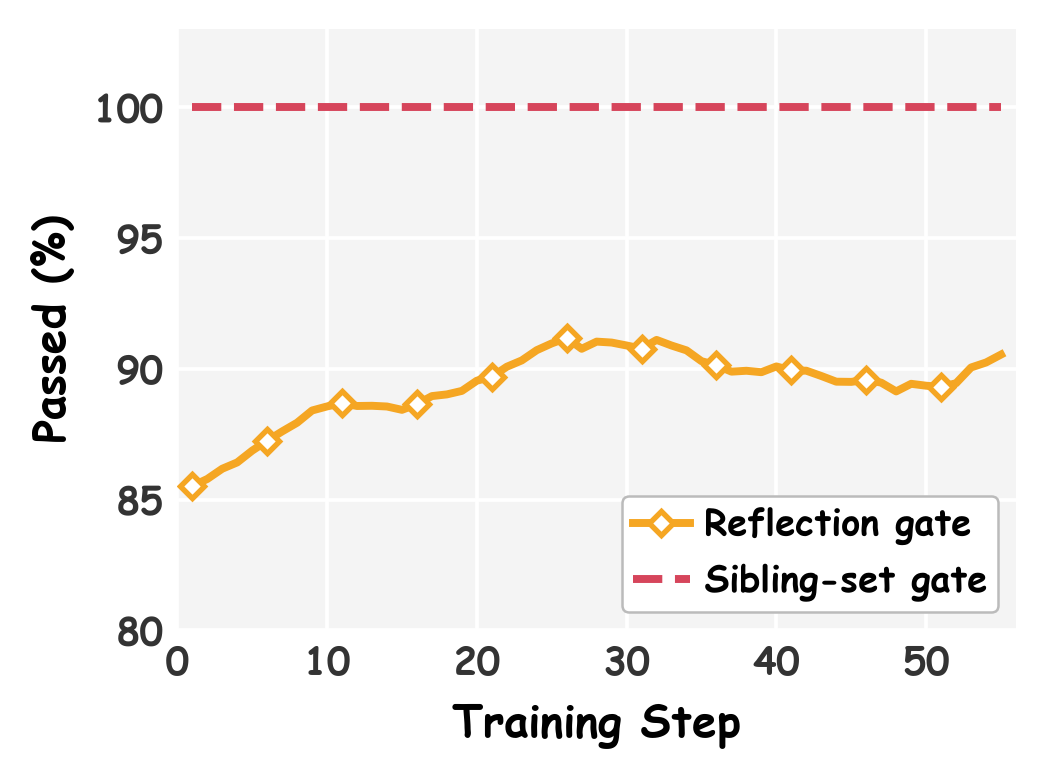}
\captionof{figure}{Fraction of rollouts passing the quality gate.}
  \label{fig:hint_gate}
\end{minipage}\hfill
\begin{minipage}[t]{0.32\textwidth}
  \centering
  \includegraphics[width=\linewidth]{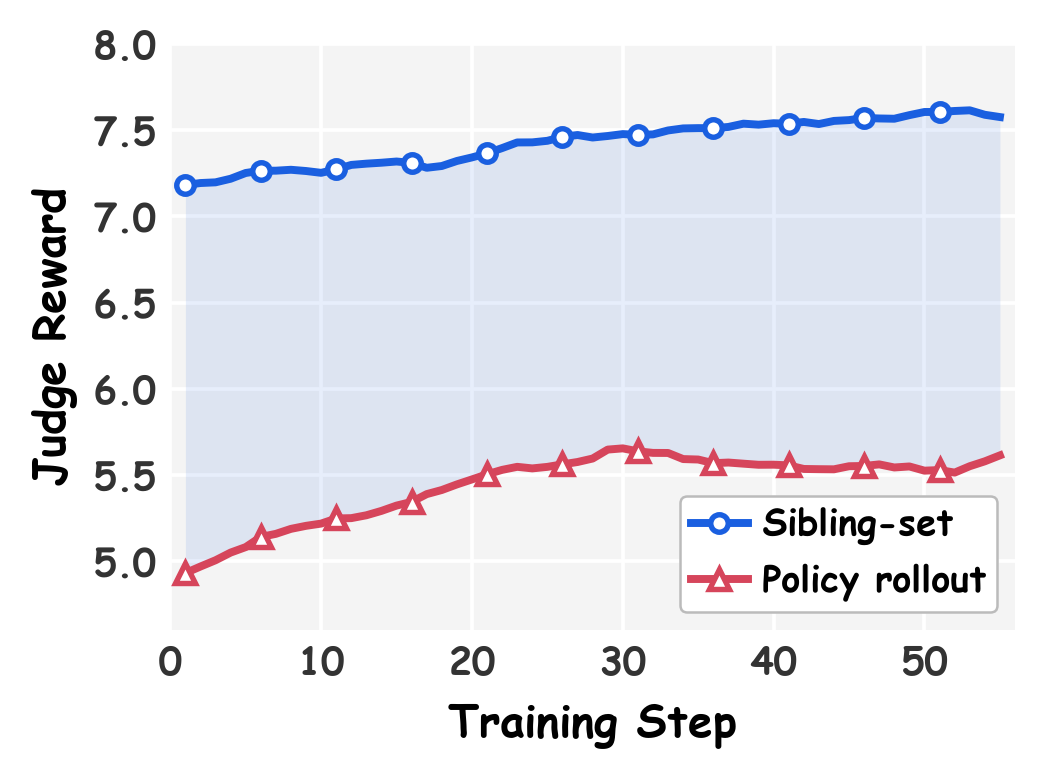}
  \captionof{figure}{Judge reward of the sibling-set and rollouts.}
  \label{fig:hint_reward}
\end{minipage}
\end{figure*}

\subsection{Training Dynamics of Adaptive Tutoring Hints}\label{appendix:hint_dynamics}

Since the sibling set is produced by the policy's own snapshot under rubric guidance, one may ask whether the three-way assignment is genuinely exercised during training or silently collapses into the rubric hint. Figure~\ref{fig:hint_shares} shows that all three forms remain in use throughout: the rubric share grows from roughly two fifths of the rollouts to about three fifths, the sibling-set share shrinks from under a fifth to about a tenth, and reflections make up the remainder, easing from about two fifths to roughly three tenths. The shift is in the expected direction, as an improving policy places more of its rollouts in the high-reward band and needs the most prescriptive form less often, so the hint space behaves like a curriculum that fades rather than a fixed context. Figure~\ref{fig:hint_gate} shows that the admission gate is rarely the binding constraint: it passes for every rollout in the sibling-set band and for $85\%$ to $90\%$ of those in the reflection band, so fallbacks are the exception. Figure~\ref{fig:hint_reward} explains why. The sibling set is scored between $7.2$ and $7.6$ by the judge while the policy's own rollouts move from $5.0$ to $5.6$, so rubric-guided re-selection extracts a set that stays about two points ahead of what the policy produces by default, and the reference remains informative even as the policy strengthens.

\begin{table*}[t]
\scriptsize
\centering
\setlength{\tabcolsep}{1.4mm}
\renewcommand{\arraystretch}{1.16}
\begin{tabular}{@{}p{0.115\textwidth}p{0.115\textwidth}p{\dimexpr0.72\textwidth-6\tabcolsep\relax}>{\centering\arraybackslash}p{0.05\textwidth}@{}}
\toprule
\multicolumn{4}{@{}p{\textwidth}@{}}{\textbf{Query:} In which year did the country that encompassed Beyra become independent?} \\ \midrule
\textbf{Level} & \textbf{Category} & \textbf{Description} & \textbf{Weight} \\ \midrule
Document-level & Relevance & Does each document directly discuss which country Beyra belongs to, or the independence date/year of that country, rather than being about unrelated topics that merely mention the word ``Beyra'' or the year ``1960''? & 5 \\
Document-level & Authenticity & For each document that states the year in which the country encompassing Beyra became independent, is the stated year consistent with the objective fact that it is 1960, without misstating it as a different year? & 5 \\
Document-level & Quality & Is each document clearly structured and focused, allowing a reader to directly extract the independence year (1960) and, if present, the identity of the country that encompasses Beyra, rather than burying these key facts in noisy or irrelevant content? & 2 \\
Set-level & Complementarity & Does the document set contain, across its documents, at least one that specifies which country encompasses Beyra AND at least one that provides the independence year of that country, so that the two pieces of information together enable determining the correct year? & 4 \\
Set-level & Redundancy & Does the document set avoid having two or more documents that convey essentially the same information --- that the country containing Beyra gained independence in 1960 --- with no incremental contribution (e.g., no additional detail, source, or nuance)? & 2 \\
Set-level & Conflict & Across the documents, are all statements about the independence year of the country that encompassed Beyra free of mutual contradiction --- specifically, do they all support the year 1960, without any document claiming a different year such as 1958 or 1962? & 4 \\
Global-level & Completeness & Does the whole document set provide both (a) enough information to identify which country encompasses Beyra, and (b) the independence year of that country, leaving no gap that would prevent a direct answer of ``1960''? & 5 \\
Global-level & Density & Across the documents, does the text that directly helps identify the country containing Beyra or states its independence year make up the majority of each document's length, rather than each document being dominated by unrelated content with only a small fraction mentioning Beyra or the year 1960? & 2 \\
Global-level & Reachability & Using only the document set and no external knowledge, can the full reasoning chain be completed --- determining which country encompasses Beyra, then obtaining that country's independence year --- to produce the correct answer ``1960''? & 4 \\ \bottomrule
\end{tabular}
\caption{Case study of query-specific rubrics for a multi-hop question in the RAG scenario.}
\label{tab:case_study_rag}
\end{table*}

\section{Case Study}\label{appendix:case}

\subsection{Case Study of Query-Specific Rubrics}

To make the query-specific rubrics concrete, Tables~\ref{tab:case_study_rag}
and~\ref{tab:case_study_dr} show the full rubric set generated for one RAG query and one
deep research query. The RAG case is a bridging multi-hop question whose rubrics pin down a
single verifiable fact and the two-document evidence chain needed to reach it, whereas the
deep research case is an open-ended mechanistic query whose rubrics instead enumerate the
distinct aspects a useful subset must jointly cover. Both are instantiated over the same
three tiers and nine dimensions, but the weights and the content of each dimension are
tailored to the query: relevance and completeness dominate in the factoid case, while
complementarity across documents carries the most weight in the deep research case. This
contrast illustrates why a fixed, query-agnostic criterion is insufficient for set-level
selection.

\begin{table*}[t]
\scriptsize
\centering
\setlength{\tabcolsep}{1.4mm}
\renewcommand{\arraystretch}{1.16}
\begin{tabular}{@{}p{0.115\textwidth}p{0.115\textwidth}p{\dimexpr0.72\textwidth-6\tabcolsep\relax}>{\centering\arraybackslash}p{0.05\textwidth}@{}}
\toprule
\multicolumn{4}{@{}p{\textwidth}@{}}{\textbf{Query:} endothelial senescence preeclampsia telomere p16 p21 oxidative stress sFlt-1 biomarkers} \\ \midrule
\textbf{Level} & \textbf{Category} & \textbf{Description} & \textbf{Weight} \\ \midrule
Document-level & Relevance & Does each document directly address the mechanistic relationship between preeclampsia, later-life cardiovascular disease (CVD), and endothelial senescence (including pathways such as oxidative stress, sFlt-1, senescence markers p16/p21, and the SASP), rather than merely mentioning preeclampsia or CVD without discussing the endothelial senescence connection? & 5 \\
Document-level & Authenticity & For each document, does its stated risk magnitude for future cardiovascular disease after preeclampsia align with the established range of approximately 2--4-fold increased risk (of hypertension, ischemic heart disease, stroke, heart failure), rather than claiming a dramatically different number or denying the association? & 4 \\
Document-level & Authenticity & For each document, does its description of sFlt-1 correctly state that it is a soluble VEGF receptor-1 that binds and neutralizes VEGF and PlGF, acting as an anti-angiogenic factor that promotes endothelial dysfunction and senescence, rather than misrepresenting its function or directional effect? & 4 \\
Document-level & Quality & Is each document clearly structured and focused enough that a reader can directly extract the relationships between preeclampsia, endothelial senescence mechanisms (oxidative stress, sFlt-1, p16/p21, SASP), and later cardiovascular disease risk, without these elements being buried in large amounts of unrelated obstetrics or vascular biology content? & 2 \\
Set-level & Complementarity & Does the document set collectively cover all the key information elements required to fully explain the relationship --- preeclampsia as a state of exaggerated vascular aging, evidence for oxidative stress-induced cellular senescence (p16/p21, telomere dysfunction), the role of sFlt-1 and VEGF deprivation, the SASP-driven inflammatory milieu, the connection to later CVD risk, and biomarker implications --- such that the documents jointly build a complete picture? & 5 \\
Set-level & Redundancy & Does the document set avoid having two or more documents that convey essentially the same core information --- for example, multiple documents merely restating in similar words that preeclampsia involves oxidative stress and endothelial dysfunction without adding distinct aspects like sFlt-1 signaling, p16/p21 biomarker data, or long-term CVD risk evidence? & 2 \\
Set-level & Conflict & Are the documents in the set free of mutual contradictions on key factual elements --- for example, not having one document claim a 5-fold CVD risk while another says no increased risk, or contradicting on sFlt-1's role (one stating it inhibits angiogenesis and another claiming it promotes angiogenesis) --- so that the set consistently supports the correct mechanistic account? & 4 \\
Global-level & Completeness & Does the whole document set, taken as a single unit, cover every essential information element needed to completely explain the relationship between preeclampsia, endothelial senescence, and cardiovascular disease, including preeclampsia pathophysiology, cellular senescence mechanisms, evidence for endothelial senescence in PE, persistent senescence linking to future CVD, and biomarker candidates? & 5 \\
Global-level & Density & Across the documents in the set, does the text that actually addresses the preeclampsia--endothelial senescence--CVD relationship make up the majority of each document's length, rather than each document being dominated by irrelevant content such as general obstetrics, unrelated vascular biology, or filler text where only a small snippet is useful? & 2 \\
Global-level & Reachability & Using only the document set and no external knowledge, can a reader trace the full reasoning chain --- from preeclampsia's placental oxidative stress and anti-angiogenic state, to induction of endothelial senescence via DNA damage, p53/p21/p16 pathways, telomere dysfunction and SASP, and then to the 2--4-fold increased risk of later cardiovascular disease --- with every step grounded in the documents? & 4 \\ \bottomrule
\end{tabular}
\caption{Case study of query-specific rubrics for an agent query in the deep research scenario.}
\label{tab:case_study_dr}
\end{table*}

\subsection{Case Study of RAG}

\begin{table*}[t]
\scriptsize
\centering
\renewcommand{\arraystretch}{1.16}
\begin{tabular}{@{}>{\raggedright\arraybackslash}p{0.145\textwidth}@{\hspace{0.010\textwidth}}>{\raggedright\arraybackslash}p{0.185\textwidth}@{\hspace{0.010\textwidth}}>{\raggedright\arraybackslash}p{0.125\textwidth}@{\hspace{0.030\textwidth}}>{\raggedright\arraybackslash}p{0.145\textwidth}@{\hspace{0.010\textwidth}}>{\raggedright\arraybackslash}p{0.185\textwidth}@{\hspace{0.010\textwidth}}>{\raggedright\arraybackslash}p{0.125\textwidth}@{}}
\toprule
\multicolumn{6}{@{}>{\raggedright\arraybackslash}p{0.970\textwidth}@{}}{\textbf{Benchmark:} Natural Questions \quad \textbf{Query:} when did macbook pro 13 inch come out? \quad \textbf{Gold answer:} \textit{October 2008}} \\ \midrule
\multicolumn{6}{@{}>{\raggedright\arraybackslash}p{0.970\textwidth}@{}}{\makebox[0.045\textwidth][l]{\textbf{ID}}\textbf{Candidate document}} \\ \midrule
\multicolumn{6}{@{}>{\raggedright\arraybackslash}p{0.970\textwidth}@{}}{\makebox[0.045\textwidth][l]{[1]}\parbox[t]{0.920\textwidth}{\raggedright MacBook Pros come with ExpressCard/34 slots, which replace the PC Card slots found in the PowerBook G4. All first generation 15-inch models have two USB 2.0 ports and one FireWire 400 port, while the 17-inch \dots}} \\
\multicolumn{6}{@{}>{\raggedright\arraybackslash}p{0.970\textwidth}@{}}{\makebox[0.045\textwidth][l]{[2]}\parbox[t]{0.920\textwidth}{\raggedright quoted at eight hours, with 80 percent of this charge remaining after 1,000 charge-discharge cycles. At Apple's Worldwide Developers Conference (WWDC) \hlb{on June 8, 2009, it was announced that the 13-inch unibody MacBook would be upgraded and re-branded as a MacBook Pro}, leaving only the white polycarbonate MacBook in the MacBook line. [\dots]}} \\
\multicolumn{6}{@{}>{\raggedright\arraybackslash}p{0.970\textwidth}@{}}{\makebox[0.045\textwidth][l]{[3]}\parbox[t]{0.920\textwidth}{\raggedright Pro follows the design of the previous two generations with an all-metal unibody enclosure and separated black keys. A few apparent design changes are a thinner chassis, thinner screen bezel, larger trackpad, \dots}} \\
\multicolumn{6}{@{}>{\raggedright\arraybackslash}p{0.970\textwidth}@{}}{\makebox[0.045\textwidth][l]{[4]}\parbox[t]{0.920\textwidth}{\raggedright the 15-inch version. All models come with 4 GB of system memory that is upgradeable to 8 GB. Battery life was also extended further in this update, to an estimated ten hours for the 13-inch and 8--9 hours \dots}} \\
\multicolumn{6}{@{}>{\raggedright\arraybackslash}p{0.970\textwidth}@{}}{\makebox[0.045\textwidth][l]{[5]}\parbox[t]{0.920\textwidth}{\raggedright i9 MacBook Pro was slower than the 2017 MacBook Pro and stated, ``This isn't a problem with Intel's Core i9, it's Apple's thermal solution.'' When Lee put the i9 MacBook Pro inside a freezer, the render \dots}} \\
\multicolumn{6}{@{}>{\raggedright\arraybackslash}p{0.970\textwidth}@{}}{\makebox[0.045\textwidth][l]{[6]}\parbox[t]{0.920\textwidth}{\raggedright MacBook Pro The MacBook Pro (sometimes abbreviated as MBP) is a line of Macintosh portable computers introduced in January 2006 by Apple Inc. It is the high-end model of the MacBook family and is currently \dots}} \\
\multicolumn{6}{@{}>{\raggedright\arraybackslash}p{0.970\textwidth}@{}}{\makebox[0.045\textwidth][l]{[7]}\parbox[t]{0.920\textwidth}{\raggedright FireWire 800 port and all except the 17-inch models would receive an SD card slot. The 17-inch model would retain its ExpressCard/34 slot. For the 13-inch MacBook Pro, the Kensington lock slot was moved \dots}} \\
\multicolumn{6}{@{}>{\raggedright\arraybackslash}p{0.970\textwidth}@{}}{\makebox[0.045\textwidth][l]{[8]}\parbox[t]{0.920\textwidth}{\raggedright 13-inch model now comes with a 128GB storage option, down from the base 256GB storage. On July 12, 2018 Apple updated the Touch Bar models with Intel Coffee Lake quad-core processors in 13-inch models \dots}} \\
\multicolumn{6}{@{}>{\raggedright\arraybackslash}p{0.970\textwidth}@{}}{\makebox[0.045\textwidth][l]{[9]}\parbox[t]{0.920\textwidth}{\raggedright MacBook Pro is Apple's higher end laptop available in both 13-inch and 15-inch configurations. A redesigned MacBook Pro was introduced on October 27, 2016, which is thinner and lighter than the previous \dots}} \\
\multicolumn{6}{@{}>{\raggedright\arraybackslash}p{0.970\textwidth}@{}}{\makebox[0.045\textwidth][l]{[10]}\parbox[t]{0.920\textwidth}{\raggedright or red if the battery is charging. MagSafe can be found on the MacBook, MacBook Pro and MacBook Air notebook computers, as well as the Apple LED Cinema Display. The MacBook and the 13-inch MacBook Pro use \dots}} \\
\multicolumn{6}{@{}>{\raggedright\arraybackslash}p{0.970\textwidth}@{}}{\makebox[0.045\textwidth][l]{[11]}\parbox[t]{0.920\textwidth}{\raggedright MacBook Pro. None of the three 17-inch models of the MacBook Pro have used any pentalobe screws. The MacBook Air has seen more extensive use of pentalobe screws than the MacBook Pro. All five versions \dots}} \\
\multicolumn{6}{@{}>{\raggedright\arraybackslash}p{0.970\textwidth}@{}}{\makebox[0.045\textwidth][l]{[12]}\parbox[t]{0.920\textwidth}{\raggedright its eight-hour capacity. Some sources even reported up to eight hours of battery life for the 13- and 15-inch MacBook Pros during casual use, while others reported around six hours. Like the 17-inch \dots}} \\
\multicolumn{6}{@{}>{\raggedright\arraybackslash}p{0.970\textwidth}@{}}{\makebox[0.045\textwidth][l]{[13]}\parbox[t]{0.920\textwidth}{\raggedright ports, and the default RAM on premium models was increased to 8 GB. Following this announcement, the 17-inch model was discontinued. After a media event on October 22, 2013 Apple discontinued all second \dots}} \\
\multicolumn{6}{@{}>{\raggedright\arraybackslash}p{0.970\textwidth}@{}}{\makebox[0.045\textwidth][l]{[14]}\parbox[t]{0.920\textwidth}{\raggedright mistaken for 5-point Torx screws. This was the only internal usage of pentalobe screws; all following MacBook Pros use the ``Tri-Wing'' security bit to attach the battery to the internal frame, or else \dots}} \\
\multicolumn{6}{@{}>{\raggedright\arraybackslash}p{0.970\textwidth}@{}}{\makebox[0.045\textwidth][l]{[15]}\parbox[t]{0.920\textwidth}{\raggedright Pro units without Touch Bar, manufactured between October 2016 and October 2017, may have the built-in battery expanded, which is also known as ``swelling''. Apple initiated a free replacement program \dots}} \\
\multicolumn{6}{@{}>{\raggedright\arraybackslash}p{0.970\textwidth}@{}}{\makebox[0.045\textwidth][l]{[16]}\parbox[t]{0.920\textwidth}{\raggedright macOS High Sierra 10.13.4. Devices using HDMI, previous generation Thunderbolt, and USB will require an adapter to connect to the MacBook Pro. The models come with a 3.5 mm headphone jack, although \dots}} \\
\multicolumn{6}{@{}>{\raggedright\arraybackslash}p{0.970\textwidth}@{}}{\makebox[0.045\textwidth][l]{[17]}\parbox[t]{0.920\textwidth}{\raggedright added as an option for the 17-inch model. Processors were updated to ``Penryn'' cores, which are built on the 45 nanometer process (65 nanometer ``Merom'' cores were previously used), and hard drive \dots}} \\
\multicolumn{6}{@{}>{\raggedright\arraybackslash}p{0.970\textwidth}@{}}{\makebox[0.045\textwidth][l]{[18]}\parbox[t]{0.920\textwidth}{\raggedright is thinner than its predecessor and is the first to include a high-resolution Retina Display. A 13-inch variant was released in October 2012. The fourth generation MacBook Pro was announced on October \dots}} \\
\multicolumn{6}{@{}>{\raggedright\arraybackslash}p{0.970\textwidth}@{}}{\makebox[0.045\textwidth][l]{[19]}\parbox[t]{0.920\textwidth}{\raggedright processors later that year. The product's second iteration, known as the ``unibody'' model, has a casing made from a single piece of aluminum. \hlb{It debuted in October 2008 in 13- and 15-inch screen sizes.} In January 2009, the 17-inch model was updated with the same unibody design. [\dots]}} \\
\multicolumn{6}{@{}>{\raggedright\arraybackslash}p{0.970\textwidth}@{}}{\makebox[0.045\textwidth][l]{[20]}\parbox[t]{0.920\textwidth}{\raggedright requirements lists the following requirements for Mac OS X Lion and Mac OS X Mountain Lion: Apple lists the following requirements for Mac OS X 10.5 Leopard and Mac OS X 10.6 Snow Leopard: Officially \dots}} \\
\midrule
\textbf{Reranker} & \textbf{Selected documents} & \textbf{Answer} & \textbf{Reranker} & \textbf{Selected documents} & \textbf{Answer} \\ \midrule
Initial Retrieval & \hlbad{[1]}~\hlok{[2]}~\hlbad{[3]}~\hlbad{[4]}~\hlbad{[5]} & 2006 & RankZephyr (7B) & \hlbad{[6]}~\hlbad{[18]}~\hlok{[19]}~\hlbad{[7]}~\hlok{[2]} & October 2012 \\
BGE-Reranker$_{L.}$ & \hlbad{[9]}~\hlbad{[6]}~\hlbad{[18]}~\hlbad{[8]}~\hlbad{[7]} & October 2012 & Rearank (7B) & \hlbad{[6]}~\hlok{[2]}~\hlbad{[18]}~\hlbad{[9]}~\hlok{[19]} & October 2012 \\
MonoT5 (3B) & \hlbad{[18]}~\hlbad{[9]}~\hlbad{[6]}~\hlbad{[13]}~\hlok{[2]} & October 2012 & ReasonRank (7B) & \hlbad{[9]}~\hlok{[2]}~\hlbad{[17]}~\hlbad{[18]}~\hlbad{[6]} & July 2020 \\
RankT5 (3B) & \hlbad{[6]}~\hlbad{[9]}~\hlbad{[18]}~\hlok{[2]}~\hlok{[19]} & October 2012 & Rank4Gen (8B) & \hlbad{[18]}~\hlok{[2]}~\hlbad{[9]} & October 2012 \\
RankLlama (7B) & \hlbad{[18]}~\hlbad{[13]}~\hlbad{[9]}~\hlbad{[7]}~\hlbad{[6]} & January 2006 & SetR (8B) & \hlok{[2]}~\hlbad{[6]}~\hlbad{[18]}~\hlok{[19]} & October 2012 \\
RankVicuna (7B) & \hlok{[2]}~\hlbad{[9]}~\hlbad{[8]}~\hlbad{[12]}~\hlbad{[13]} & -- & RubricRanker (8B) & \hlbad{[18]}~\hlok{[2]}~\hlok{[19]}~\hlbad{[12]}~\hlbad{[7]} & October 2012 \\
\textbf{AdaTutoRank (8B)} & \hlok{[2]}~\hlok{[19]} & \textbf{October 2008} &  &  &  \\
\bottomrule
\end{tabular}
\caption{Case study on Natural Questions for the query \emph{when did macbook pro 13 inch come out?} The upper block lists the 20 BM25 candidates, where \protect\hlbx{blue} marks the spans that establish the answer. The lower block lists the documents each method feeds to the generator (the top-5 for a ranking method, the chosen subset for a set selector), where \protect\hlok{green} marks a selected passage that supports the answer and \protect\hlbad{red} one that does not, together with the answer the generator then produced. Passages 6, 9 and 18 date other generations of the machine, January 2006, October 2016 and October 2012; eleven of the twelve baselines forward at least one of them and nine answer with one of those years. Ours selects exactly the two passages that date the 13-inch model. [\dots] marks an elision inside a passage.}
\label{tab:case_study_nq}
\end{table*}
\begin{table*}[t]
\scriptsize
\centering
\renewcommand{\arraystretch}{1.16}
\begin{tabular}{@{}>{\raggedright\arraybackslash}p{0.145\textwidth}@{\hspace{0.010\textwidth}}>{\raggedright\arraybackslash}p{0.185\textwidth}@{\hspace{0.010\textwidth}}>{\raggedright\arraybackslash}p{0.125\textwidth}@{\hspace{0.030\textwidth}}>{\raggedright\arraybackslash}p{0.145\textwidth}@{\hspace{0.010\textwidth}}>{\raggedright\arraybackslash}p{0.185\textwidth}@{\hspace{0.010\textwidth}}>{\raggedright\arraybackslash}p{0.125\textwidth}@{}}
\toprule
\multicolumn{6}{@{}>{\raggedright\arraybackslash}p{0.970\textwidth}@{}}{\textbf{Benchmark:} 2WikiMultihopQA \quad \textbf{Query:} Where did Dolores Hope's husband die? \quad \textbf{Gold answer:} \textit{Toluca Lake, Los Angeles}} \\ \midrule
\multicolumn{6}{@{}>{\raggedright\arraybackslash}p{0.970\textwidth}@{}}{\makebox[0.045\textwidth][l]{\textbf{ID}}\textbf{Candidate document}} \\ \midrule
\multicolumn{6}{@{}>{\raggedright\arraybackslash}p{0.970\textwidth}@{}}{\makebox[0.045\textwidth][l]{[1]}\parbox[t]{0.920\textwidth}{\raggedright and her daughter. Vera suggests Dolores take advantage of the up-coming eclipse to solve her problem, singing, ``accidents can be an unhappy woman's best friend.'' The second act opens with Dolores preparing food and liquor \dots}} \\
\multicolumn{6}{@{}>{\raggedright\arraybackslash}p{0.970\textwidth}@{}}{\makebox[0.045\textwidth][l]{[2]}\parbox[t]{0.920\textwidth}{\raggedright Honorary Board Member of the humanitarian organization Wings of Hope. On May 29, 2003, Dolores was at her husband's side as he celebrated his 100th birthday; he died two months later on July 27, 2003. They had been married \dots}} \\
\multicolumn{6}{@{}>{\raggedright\arraybackslash}p{0.970\textwidth}@{}}{\makebox[0.045\textwidth][l]{[3]}\parbox[t]{0.920\textwidth}{\raggedright accounts to fund their escape, she discovers Joe has stolen everything she had saved. In despair, she breaks down crying at work, forcing her to confide her troubles in Vera. An unusually sympathetic Vera reveals she has had \dots}} \\
\multicolumn{6}{@{}>{\raggedright\arraybackslash}p{0.970\textwidth}@{}}{\makebox[0.045\textwidth][l]{[4]}\parbox[t]{0.920\textwidth}{\raggedright Terrill was around. Toto mentions that Terrill was with Dolores the whole time. The chief then interrogates Dolores about Terrill and where her husband was. She mentions that she couldn't help but fall for Terrill and she \dots}} \\
\multicolumn{6}{@{}>{\raggedright\arraybackslash}p{0.970\textwidth}@{}}{\makebox[0.045\textwidth][l]{[5]}\parbox[t]{0.920\textwidth}{\raggedright receives a letter from the residents of Ramsdale, who have learnt that Dolores has gone missing and are pressing for answers. Knowing that his situation is precarious and contemplating what to do, Humbert eventually receives \dots}} \\
\multicolumn{6}{@{}>{\raggedright\arraybackslash}p{0.970\textwidth}@{}}{\makebox[0.045\textwidth][l]{[6]}\parbox[t]{0.920\textwidth}{\raggedright Lo's Diary Lo's Diary () is a 1995 novel () by Pia Pera, retelling Vladimir Nabokov's novel ``Lolita'' from the point of view of ``Dolores Haze (Lolita)''. It depicts Dolores as a sadist and a controller of everyone around her; \dots}} \\
\multicolumn{6}{@{}>{\raggedright\arraybackslash}p{0.970\textwidth}@{}}{\makebox[0.045\textwidth][l]{[7]}\parbox[t]{0.920\textwidth}{\raggedright involved in a centuries-old battle between the Animus and the Dolore clans. Long ago one of a pair of a Book of Curses was stolen by the Dolore, and because of that they were put under a curse where they become beasts and \dots}} \\
\multicolumn{6}{@{}>{\raggedright\arraybackslash}p{0.970\textwidth}@{}}{\makebox[0.045\textwidth][l]{[8]}\parbox[t]{0.920\textwidth}{\raggedright as Meri Bell), played integral roles in encouraging former First Lady Betty Ford to establish the Betty Ford Center. Bell was Ford's sponsor in Alcoholics Anonymous. Ford's obituary in USA Today related developments after \dots}} \\
\multicolumn{6}{@{}>{\raggedright\arraybackslash}p{0.970\textwidth}@{}}{\makebox[0.045\textwidth][l]{[9]}\parbox[t]{0.920\textwidth}{\raggedright wealthy, elderly employer, Vera Donovan. The novel is presented as a transcript of her statement, told to the local constable and a stenographer. Dolores wants to make clear to the police that she did not kill Vera, whom she \dots}} \\
\multicolumn{6}{@{}>{\raggedright\arraybackslash}p{0.970\textwidth}@{}}{\makebox[0.045\textwidth][l]{[10]}\parbox[t]{0.920\textwidth}{\raggedright the British Empire Exhibition, where the exotic foreign displays intrigued him, or possibly through his friend Matthew Smith. In 1925 Epstein invited Sunita, Enver and Anita to live at his home at Guilford Street in London \dots}} \\
\multicolumn{6}{@{}>{\raggedright\arraybackslash}p{0.970\textwidth}@{}}{\makebox[0.045\textwidth][l]{[11]}\parbox[t]{0.920\textwidth}{\raggedright of the day off by Vera. Dolores and Selena had an argument about Dolores' suspicions regarding Joe's sexual abuse. Selena fled home for the weekend to work at a hotel, where guests had flocked for the eclipse. Joe soon \dots}} \\
\multicolumn{6}{@{}>{\raggedright\arraybackslash}p{0.970\textwidth}@{}}{\makebox[0.045\textwidth][l]{[12]}\parbox[t]{0.920\textwidth}{\raggedright Dolores Richard Spikes Dolores Margaret Richard Spikes (August 24, 1936 -- June 1, 2015) was an American mathematician and university administrator. Born in Baton Rouge, Dolores Richard attended public and parochial schools \dots}} \\
\multicolumn{6}{@{}>{\raggedright\arraybackslash}p{0.970\textwidth}@{}}{\makebox[0.045\textwidth][l]{[13]}\parbox[t]{0.920\textwidth}{\raggedright 1970, the Spanish right-wing newspaper ``ABC'' reported that the PCE and the Kremlin had reached a new pact whereby the Spanish party dropped its censure of the Soviet invasion of Czechoslovakia in exchange for the Kremlin's \dots}} \\
\multicolumn{6}{@{}>{\raggedright\arraybackslash}p{0.970\textwidth}@{}}{\makebox[0.045\textwidth][l]{[14]}\parbox[t]{0.920\textwidth}{\raggedright kill her. Vera dies and the police begin a murder investigation. Dolores' daughter, Selena St. George, is a successful journalist, living in New York City, who battles depression and substance abuse. Selena arrives in town \dots}} \\
\multicolumn{6}{@{}>{\raggedright\arraybackslash}p{0.970\textwidth}@{}}{\makebox[0.045\textwidth][l]{[15]}\parbox[t]{0.920\textwidth}{\raggedright that he created a trade union at the 'bodega' where he worked during the war. Dona Dolores: Dona Dolores is more conservative than her liberal husband. Whilst she is easily persuaded by her husband, she often disapproves of \dots}} \\
\multicolumn{6}{@{}>{\raggedright\arraybackslash}p{0.970\textwidth}@{}}{\makebox[0.045\textwidth][l]{[16]}\parbox[t]{0.920\textwidth}{\raggedright he would die in captivity. Among her tutors during her youth was her mother's sister, Maria Dolores Molina, who was the mother of her first cousin and future husband Manuel Luis Quezon. After her father's imprisonment, she \dots}} \\
\multicolumn{6}{@{}>{\raggedright\arraybackslash}p{0.970\textwidth}@{}}{\makebox[0.045\textwidth][l]{[17]}\parbox[t]{0.920\textwidth}{\raggedright 102nd birthday at her California residence. \hlb{She died of natural causes at her home in Toluca Lake, California}, on September 19, 2011. She had been in relatively good health until a few months before her death. [\dots]}} \\
\multicolumn{6}{@{}>{\raggedright\arraybackslash}p{0.970\textwidth}@{}}{\makebox[0.045\textwidth][l]{[18]}\parbox[t]{0.920\textwidth}{\raggedright through the intervention of the Spanish ambassador in Berlin that they were released. After reaching Paris, Princess Dolores and her husband moved permanently to Spain. They settled in Seville where Princess Dolores gave \dots}} \\
\multicolumn{6}{@{}>{\raggedright\arraybackslash}p{0.970\textwidth}@{}}{\makebox[0.045\textwidth][l]{[19]}\parbox[t]{0.920\textwidth}{\raggedright that Briscoe would return. Servants reported that Mrs Briscoe would go out riding where she would meet Gordon, but Gordon would see her home but never enter the house with her. Gordon eloped with Mrs Briscoe on 21 October \dots}} \\
\multicolumn{6}{@{}>{\raggedright\arraybackslash}p{0.970\textwidth}@{}}{\makebox[0.045\textwidth][l]{[20]}\parbox[t]{0.920\textwidth}{\raggedright age of 100 \hlb{at his home in Toluca Lake, California}. His grandson Zach Hope told TV interviewer Soledad O'Brien that, when asked on his deathbed where he wanted to be buried, Hope told his wife, Dolores, ``Surprise me.'' [\dots]}} \\
\midrule
\textbf{Reranker} & \textbf{Selected documents} & \textbf{Answer} & \textbf{Reranker} & \textbf{Selected documents} & \textbf{Answer} \\ \midrule
Initial Retrieval & \hlbad{[1]}~\hlbad{[2]}~\hlbad{[3]}~\hlbad{[4]}~\hlbad{[5]} & Inside a well & RankZephyr (7B) & \hlbad{[2]}~\hlok{[20]}~\hlok{[17]}~\hlbad{[14]}~\hlbad{[9]} & Home \\
BGE-Reranker$_{L.}$ & \hlok{[17]}~\hlbad{[2]}~\hlok{[20]}~\hlbad{[1]}~\hlbad{[8]} & Home & Rearank (7B) & \hlok{[20]}~\hlok{[17]}~\hlbad{[2]}~\hlbad{[1]}~\hlbad{[9]} & Home \\
MonoT5 (3B) & \hlbad{[1]}~\hlbad{[2]}~\hlok{[17]}~\hlok{[20]}~\hlbad{[13]} & In a well & ReasonRank (7B) & \hlok{[17]}~\hlok{[20]}~\hlbad{[2]}~\hlbad{[1]}~\hlbad{[8]} & Home \\
RankT5 (3B) & \hlok{[17]}~\hlok{[20]}~\hlbad{[1]}~\hlbad{[2]}~\hlbad{[13]} & Home & Rank4Gen (8B) & \hlok{[20]}~\hlbad{[2]}~\hlok{[17]} & Home \\
RankLlama (7B) & \hlbad{[2]}~\hlok{[17]}~\hlok{[20]}~\hlbad{[1]}~\hlbad{[3]} & Home & SetR (8B) & \hlbad{[2]}~\hlok{[17]}~\hlok{[20]} & Home \\
RankVicuna (7B) & \hlbad{[2]}~\hlbad{[1]}~\hlbad{[11]}~\hlbad{[3]}~\hlbad{[14]} & A well & RubricRanker (8B) & \hlok{[20]}~\hlbad{[2]}~\hlok{[17]} & Home \\
\textbf{AdaTutoRank (8B)} & \hlok{[17]}~\hlok{[20]} & \textbf{Toluca Lake} &  &  &  \\
\bottomrule
\end{tabular}
\caption{Case study on 2WikiMultihopQA for the two-hop query \emph{Where did Dolores Hope's husband die?} The upper block lists the 20 BM25 candidates, where \protect\hlbx{blue} marks the spans that establish the answer: passage~20 states that he died at his home in Toluca Lake, and passage~17 gives the same address as the couple's home. The lower block lists the documents each method feeds to the generator (the top-5 for a ranking method, the chosen subset for a set selector), where \protect\hlok{green} marks a selected passage that supports the answer and \protect\hlbad{red} one that does not. [\dots] marks an elision inside a passage.}
\label{tab:case_study_2wiki}
\end{table*}
\begin{table*}[t]
\scriptsize
\centering
\renewcommand{\arraystretch}{1.16}
\begin{tabular}{@{}>{\raggedright\arraybackslash}p{0.145\textwidth}@{\hspace{0.010\textwidth}}>{\raggedright\arraybackslash}p{0.185\textwidth}@{\hspace{0.010\textwidth}}>{\raggedright\arraybackslash}p{0.125\textwidth}@{\hspace{0.030\textwidth}}>{\raggedright\arraybackslash}p{0.145\textwidth}@{\hspace{0.010\textwidth}}>{\raggedright\arraybackslash}p{0.185\textwidth}@{\hspace{0.010\textwidth}}>{\raggedright\arraybackslash}p{0.125\textwidth}@{}}
\toprule
\multicolumn{6}{@{}>{\raggedright\arraybackslash}p{0.970\textwidth}@{}}{\textbf{Benchmark:} PopQA \quad \textbf{Query:} What genre is Whitney? \quad \textbf{Gold answer:} \textit{pop music}} \\ \midrule
\multicolumn{6}{@{}>{\raggedright\arraybackslash}p{0.970\textwidth}@{}}{\makebox[0.045\textwidth][l]{\textbf{ID}}\textbf{Candidate document}} \\ \midrule
\multicolumn{6}{@{}>{\raggedright\arraybackslash}p{0.970\textwidth}@{}}{\makebox[0.045\textwidth][l]{[1]}\parbox[t]{0.920\textwidth}{\raggedright Whitney Awards The Whitney Awards are awards given annually for novels by LDS authors. Established in 2007, they are named after Orson F. Whitney, a prominent early member of the LDS Church. There are several categories for \dots}} \\
\multicolumn{6}{@{}>{\raggedright\arraybackslash}p{0.970\textwidth}@{}}{\makebox[0.045\textwidth][l]{[2]}\parbox[t]{0.920\textwidth}{\raggedright James Whitney (filmmaker) ``For other people named James Whitney, see James Whitney (disambiguation)'' James Whitney (December 27, 1921 -- April 8, 1982), younger brother of John, was a filmmaker regarded as one of the great \dots}} \\
\multicolumn{6}{@{}>{\raggedright\arraybackslash}p{0.970\textwidth}@{}}{\makebox[0.045\textwidth][l]{[3]}\parbox[t]{0.920\textwidth}{\raggedright singing Vietnamese songs. Uyen Linh's idol is Whitney Houston and her favorite genre is RnB. According to one of her interviews, she admires the voice of Whitney because ``Whenever Linh hears her singing, Linh feels like \dots}} \\
\multicolumn{6}{@{}>{\raggedright\arraybackslash}p{0.970\textwidth}@{}}{\makebox[0.045\textwidth][l]{[4]}\parbox[t]{0.920\textwidth}{\raggedright her book ``The Mystery of the Haunted Pool'' won an Edgar Award from the Mystery Writers of America for Best Juvenile novel, and she duplicated the honor in 1964, for ``The Mystery of the Hidden Hand''. In 1988, the MWA gave her \dots}} \\
\multicolumn{6}{@{}>{\raggedright\arraybackslash}p{0.970\textwidth}@{}}{\makebox[0.045\textwidth][l]{[5]}\parbox[t]{0.920\textwidth}{\raggedright romance novel genre, create a community of female readers sharing in similar emotions and knowledge. After the initial publication in 1985, ``Whitney, My Love'' was adapted by a number of publishers to at least 6 different \dots}} \\
\multicolumn{6}{@{}>{\raggedright\arraybackslash}p{0.970\textwidth}@{}}{\makebox[0.045\textwidth][l]{[6]}\parbox[t]{0.920\textwidth}{\raggedright Phyllis A. Whitney Phyllis Ayame Whitney (September 9, 1903 -- February 8, 2008) was a Japanese-born American mystery writer. Rare for her genre, she wrote mysteries for both the juvenile and the adult markets, many of which \dots}} \\
\multicolumn{6}{@{}>{\raggedright\arraybackslash}p{0.970\textwidth}@{}}{\makebox[0.045\textwidth][l]{[7]}\parbox[t]{0.920\textwidth}{\raggedright sway over so many genres. But without a gospel single from ``The Preacher's Wife'' soundtrack --- some of her most emotive work --- this isn't Whitney at her best.'' AllMusic reviewer Stephen Thomas Erlewine felt that ``The Essential Whitney Houston'' plays much like ``The Greatest Hits'' [\dots]}} \\
\multicolumn{6}{@{}>{\raggedright\arraybackslash}p{0.970\textwidth}@{}}{\makebox[0.045\textwidth][l]{[8]}\parbox[t]{0.920\textwidth}{\raggedright record vocals on an anti-bullying compilation album entitled ``All About Bullies''. The Album went on to win a Grammy Award at the 54th Annual Grammy Awards for ``Best Children's Album'' as well as Parent's Choice Gold Award in \dots}} \\
\multicolumn{6}{@{}>{\raggedright\arraybackslash}p{0.970\textwidth}@{}}{\makebox[0.045\textwidth][l]{[9]}\parbox[t]{0.920\textwidth}{\raggedright genre or style, was in very few permanent collections when she was alive. Since her death, Wilke's work has been acquired into the permanent collections of The Museum of Modern Art, New York, the Whitney Museum of American \dots}} \\
\multicolumn{6}{@{}>{\raggedright\arraybackslash}p{0.970\textwidth}@{}}{\makebox[0.045\textwidth][l]{[10]}\parbox[t]{0.920\textwidth}{\raggedright revised memorials and take expeditions through what was then known as Indian Territory to support his cause. Later Whitney's dream was realized through the efforts of Theodore Judah. In the end, Whitney lived to see his \dots}} \\
\multicolumn{6}{@{}>{\raggedright\arraybackslash}p{0.970\textwidth}@{}}{\makebox[0.045\textwidth][l]{[11]}\parbox[t]{0.920\textwidth}{\raggedright an engagement present. Later, a bus crashes into the market and Whitney is trapped underneath it. Whitney is rescued and in hospital, she talks to Mick about her marriage. He comforts her and she kisses him, which is seen by \dots}} \\
\multicolumn{6}{@{}>{\raggedright\arraybackslash}p{0.970\textwidth}@{}}{\makebox[0.045\textwidth][l]{[12]}\parbox[t]{0.920\textwidth}{\raggedright number 1 hit (US R\&B) with his song ``Sensitivity'', with another on the ``House Party 2'' soundtrack ``Yo Baby Yo''. Also in 1990 \hlb{pop singer Whitney Houston} recorded ``I'm Your Baby Tonight'', produced by Babyface and his new jack swing producing partner [\dots]}} \\
\multicolumn{6}{@{}>{\raggedright\arraybackslash}p{0.970\textwidth}@{}}{\makebox[0.045\textwidth][l]{[13]}\parbox[t]{0.920\textwidth}{\raggedright father. To the surprise of her small town, Whitney returns as a sophisticated young lady. Furthermore, possessing what she believes is a substantial inheritance from her grandmother, Whitney is determined to marry Paul. Once \dots}} \\
\multicolumn{6}{@{}>{\raggedright\arraybackslash}p{0.970\textwidth}@{}}{\makebox[0.045\textwidth][l]{[14]}\parbox[t]{0.920\textwidth}{\raggedright a grandson of George Macculloch Miller (1832--1917), the founder what would become the United Hospital Fund. The marriage to ``Cully'' Miller was long and happy, and Whitney had two more children: She died on July 18, 1986 at \dots}} \\
\multicolumn{6}{@{}>{\raggedright\arraybackslash}p{0.970\textwidth}@{}}{\makebox[0.045\textwidth][l]{[15]}\parbox[t]{0.920\textwidth}{\raggedright the ``Tribune'' until its demise, but Whitney and his advisors controlled the paper. Whitney initially left management of the newspaper to Walter Thayer, a longtime advisor. Thayer did not believe the ``Tribune'' was a financial \dots}} \\
\multicolumn{6}{@{}>{\raggedright\arraybackslash}p{0.970\textwidth}@{}}{\makebox[0.045\textwidth][l]{[16]}\parbox[t]{0.920\textwidth}{\raggedright the next day, and Petersmeyer went on to become a partner at the firm and run Whitney Communications. Whitney had been investing since the 1930s, founding Pioneer Pictures in 1933 and acquiring a 15\% interest in Technicolor \dots}} \\
\multicolumn{6}{@{}>{\raggedright\arraybackslash}p{0.970\textwidth}@{}}{\makebox[0.045\textwidth][l]{[17]}\parbox[t]{0.920\textwidth}{\raggedright tours of Japan and a full-length album release followed, also entitled ``I Am What I Am''. In 2007, 2008 and 2009, the tour was also brought to Europe where she maintained a cult following. In December 2009, Whitney had a \dots}} \\
\multicolumn{6}{@{}>{\raggedright\arraybackslash}p{0.970\textwidth}@{}}{\makebox[0.045\textwidth][l]{[18]}\parbox[t]{0.920\textwidth}{\raggedright They also manufactured milling machines and twist drills. What remains of the original Pratt \& Whitney is now Pratt \& Whitney Measurement Systems, located in Bloomfield, Connecticut. Pratt \& Whitney Measurement Systems is an \dots}} \\
\multicolumn{6}{@{}>{\raggedright\arraybackslash}p{0.970\textwidth}@{}}{\makebox[0.045\textwidth][l]{[19]}\parbox[t]{0.920\textwidth}{\raggedright ends their affair. Devastated when Tony accepts Bianca's marriage proposal, Whitney locks herself in her bedroom. Not knowing what to do, Bianca accepts Dr Poppy Merritt's (Amy Darcy) help and she refers Whitney to a \dots}} \\
\multicolumn{6}{@{}>{\raggedright\arraybackslash}p{0.970\textwidth}@{}}{\makebox[0.045\textwidth][l]{[20]}\parbox[t]{0.920\textwidth}{\raggedright the institutions where art is displayed. In a review of Warhol's 1971 retrospective show at the Whitney, she observed that cows are a common subject of genre paintings that people display in their homes, and that the \dots}} \\
\midrule
\textbf{Reranker} & \textbf{Selected documents} & \textbf{Answer} & \textbf{Reranker} & \textbf{Selected documents} & \textbf{Answer} \\ \midrule
Initial Retrieval & \hlbad{[1]}~\hlbad{[2]}~\hlbad{[3]}~\hlbad{[4]}~\hlbad{[5]} & Abstract Cinema & RankZephyr (7B) & \hlbad{[3]}~\hlbad{[5]}~\hlbad{[6]}~\hlbad{[4]}~\hlbad{[1]} & Romantic Novels \\
BGE-Reranker$_{L.}$ & \hlbad{[6]}~\hlbad{[2]}~\hlbad{[1]}~\hlbad{[5]}~\hlbad{[3]} & Romantic Novels Of Suspense & Rearank (7B) & \hlbad{[3]}~\hlok{[7]}~\hlok{[12]}~\hlbad{[6]}~\hlbad{[5]} & Mystery Writer Genre \\
MonoT5 (3B) & \hlbad{[1]}~\hlbad{[6]}~\hlbad{[5]}~\hlbad{[2]}~\hlbad{[3]} & Mystery Writer Genre & ReasonRank (7B) & \hlbad{[3]}~\hlbad{[6]}~\hlbad{[5]}~\hlbad{[1]}~\hlbad{[2]} & Romantic Novels \\
RankT5 (3B) & \hlbad{[6]}~\hlbad{[1]}~\hlbad{[4]}~\hlbad{[5]}~\hlbad{[2]} & Romantic Suspense Novels & Rank4Gen (8B) & \hlok{[7]}~\hlbad{[6]}~\hlok{[12]} & Mystery Writer \\
RankLlama (7B) & \hlbad{[1]}~\hlbad{[6]}~\hlbad{[4]}~\hlbad{[2]}~\hlbad{[5]} & Romantic Novels & SetR (8B) & \hlbad{[1]}~\hlbad{[4]}~\hlbad{[6]}~\hlok{[7]}~\hlok{[12]} & Romantic Novels \\
RankVicuna (7B) & \hlbad{[1]}~\hlbad{[5]}~\hlbad{[6]}~\hlbad{[4]}~\hlbad{[2]} & Romantic Novels & RubricRanker (8B) & \hlbad{[6]}~\hlbad{[4]} & Romantic Suspense Novels \\
\textbf{AdaTutoRank (8B)} & \hlok{[7]}~\hlok{[12]} & \textbf{Pop} &  &  &  \\
\bottomrule
\end{tabular}
\caption{Case study on PopQA for the query \emph{What genre is Whitney?}, which asks about the Whitney Houston album. The upper block lists the 20 BM25 candidates, where \protect\hlbx{blue} marks the span that establishes the answer. The lower block lists the documents each method feeds to the generator (the top-5 for a ranking method, the chosen subset for a set selector), where \protect\hlok{green} marks a selected passage about that Whitney and \protect\hlbad{red} one about a different entity of the same name; the pool holds at least seven such entities, including two novelists, a filmmaker, a museum and a publisher. [\dots] marks an elision inside a passage.}
\label{tab:case_study_popqa}
\end{table*}

Tables~\ref{tab:case_study_nq}, \ref{tab:case_study_2wiki} and~\ref{tab:case_study_popqa} trace three queries end to end, and two patterns recur. First, the passages that actually establish the answer sit \emph{low} in the retrieval order: they are ranked $2$ and $19$ on NQ, $17$ and $20$ on 2WikiMultihopQA, and $7$ and $12$ on PopQA, while the head of the list is occupied by passages that share the query terms without answering it. PopQA shows this most sharply, where passage~$3$ contains the string \emph{RnB} but describes a Vietnamese singer who admires Houston. Ordering candidates by relevance therefore does not surface the evidence a generator needs, which is exactly the mismatch our objective targets. Second, retrieving the right passage is not sufficient. On NQ, RankT5, SetR and RubricRanker all forward both $[2]$ and $[19]$, yet each pairs them with passages dating other generations of the machine and the generator still answers \emph{October 2012}; on PopQA, three baselines reach $[7]$ and $[12]$ but keep the novelist alongside and answer \emph{Romantic Novels}. AdaTutoRank returns exactly the two supporting passages in all three cases and is the only method that answers all three correctly, showing that discarding a conflicting document matters as much as including a useful one.

\subsection{Case Study of Set Selection}

\begin{table*}[t]
\scriptsize
\setlength{\tabcolsep}{0pt}
\centering
\newcommand{\spanrow}[1]{\multicolumn{15}{@{}>{\raggedright\arraybackslash}p{0.9700\textwidth}@{}}{#1}}
\newcommand{\rubrow}[2]{\spanrow{\makebox[0.0400\textwidth][l]{#1}\parbox[t]{0.9250\textwidth}{\raggedright\fontsize{6.2}{7.2}\selectfont #2}}}
\newcommand{\docrow}[2]{\spanrow{\makebox[0.0450\textwidth][l]{[#1]}\parbox[t]{0.9200\textwidth}{\raggedright\fontsize{6.2}{7.2}\selectfont #2}}}
\newcommand{\rubnl}{\\[1.2pt]}
\renewcommand{\arraystretch}{1.16}
\begin{tabular}{@{}>{\raggedright\arraybackslash}p{0.2000\textwidth}@{\hspace{0.0120\textwidth}}>{\raggedright\arraybackslash}p{0.1500\textwidth}@{\hspace{0.0160\textwidth}}>{\centering\arraybackslash}p{0.0420\textwidth}@{}>{\centering\arraybackslash}p{0.0420\textwidth}@{}>{\centering\arraybackslash}p{0.0420\textwidth}@{}>{\centering\arraybackslash}p{0.0420\textwidth}@{}>{\centering\arraybackslash}p{0.0420\textwidth}@{}>{\centering\arraybackslash}p{0.0420\textwidth}@{}>{\centering\arraybackslash}p{0.0420\textwidth}@{}>{\centering\arraybackslash}p{0.0420\textwidth}@{}>{\centering\arraybackslash}p{0.0420\textwidth}@{\hspace{0.0120\textwidth}}>{\centering\arraybackslash}p{0.0505\textwidth}@{}>{\centering\arraybackslash}p{0.0505\textwidth}@{}>{\centering\arraybackslash}p{0.0505\textwidth}@{}>{\centering\arraybackslash}p{0.0505\textwidth}@{}}
\toprule
\spanrow{\textbf{Query:} Which affiliate left the company that owns and operates most of the CBC television stations due to an agreement with CHBC? \quad \textbf{Gold answer:} \textit{CFJC-TV in Kamloops}} \\ \midrule
\spanrow{\makebox[0.0400\textwidth][l]{\textbf{Dim.}}\textbf{Query-specific rubrics}} \\ \midrule
\rubrow{Rel.}{Does each document directly discuss the departure of CFJC-TV in Kamloops from the company that owns and operates most of the CBC television stations, specifically focusing on the agreement with CHBC that caused this departure, rather than merely mentioning CBC affiliates, CHBC, or unrelated affiliation changes?} \\
\rubrow{Aut.}{(i) For each document that identifies the affiliate that left the company that owns and operates most CBC television stations, does it correctly state that it was CFJC-TV in Kamloops, and not some other station?\rubnl
(ii) For each document that describes the departure of CFJC-TV in Kamloops from that company, does it attribute the departure to an agreement with CHBC, rather than any other reason?} \\
\rubrow{Qua.}{Is each document clearly structured and focused enough that a reader can directly extract the fact that CFJC-TV in Kamloops is the affiliate that left the company owning and operating most CBC television stations due to an agreement with CHBC, rather than having this specific relationship buried in unrelated, disorganized, or superficial content?} \\
\rubrow{Cmp.}{Does the document set collectively provide both the identity of the company that owns and operates most CBC television stations AND the specific fact that CFJC-TV in Kamloops left that company due to an agreement with CHBC?} \\
\rubrow{Red.}{Does the document set avoid redundancy where multiple documents merely repeat the core fact that CFJC-TV in Kamloops left the company that owns and operates most of the CBC television stations due to an agreement with CHBC, without providing any incremental information or context?} \\
\rubrow{Con.}{Across the document set, is there no factual contradiction regarding the identity of the departing affiliate (consistently identifying CFJC-TV in Kamloops) and the reason for departure (an agreement with CHBC), with no conflicting information such as a different station or city being named for the same event?} \\
\rubrow{Cpl.}{Does the document set as a whole provide the identity of the company that owns and operates most CBC television stations, state that an affiliate left this company due to an agreement with CHBC, and explicitly identify that affiliate as CFJC-TV in Kamloops?} \\
\rubrow{Den.}{Across the documents in the set, does the text that specifically identifies CFJC-TV in Kamloops as the affiliate that left the company owning and operating most CBC television stations due to an agreement with CHBC occupy the majority of each document's length, rather than each document being largely filled with unrelated content (e.g., other CBC/CHBC history, or unrelated station details) where only a small fraction of the text is actually useful for this specific affiliate change?} \\
\rubrow{Rea.}{Using only the document set and no external knowledge, can the full reasoning chain be completed --- identifying the company that owns and operates most CBC television stations, establishing that an affiliate left this company because of an agreement with CHBC, and concluding that this affiliate was CFJC-TV in Kamloops?} \\
\midrule
\spanrow{\makebox[0.0450\textwidth][l]{\textbf{ID}}\textbf{Candidate document}} \\ \midrule
\docrow{1}{[\dots] One private CBC affiliate, CHBC-TV in Kelowna, joined E! (then known as CH) on February 27, 2006. When a private CBC affiliate reaffiliated with another network, the CBC normally added a retransmitter of its nearest O\&O station. However, \hlb{due to an agreement between CHBC and CFJC-TV in Kamloops, CFJC also disaffiliated from the CBC} on February 27, 2006, but no retransmitters were installed in the licence area.} \\
\docrow{2}{[\dots] In late 2003, the CBC notified CHBC that it did not intend to renew its affiliation agreement with the station after it expired in August 2005. In response, the station filed an application with the Canadian Radio-television [\dots]} \\
\docrow{3}{[\dots] and Telecommunications Commission in 2004 to disaffiliate from the CBC; the CRTC gave approval to the disaffiliation on February 28, 2005. [\dots] After its BCI-TV partner CFJC-TV received similar approval to disaffiliate from the CBC, both stations switched affiliations on February 27, 2006 and continued the operation of BCI-TV [\dots]} \\
\docrow{4}{CICT, CHBC, CHEX, CISA and CKWS launched in the 1950s as CBC Television affiliates, while CHAN-TV launched in 1960 and soon became Vancouver's original CTV affiliate. All of these were eventually supplanted by \dots} \\
\docrow{5}{had left CBC Television solely dependent on cable and satellite carriage of its Vancouver station CBUT in the market, with no new terrestrial transmitters being installed in the Kamloops area. The Canadian \dots} \\
\docrow{6}{of Canada's Weatheradio Canada service. The CBC operates two national broadcast television networks; CBC Television in English, and Ici Radio-Canada Tele in French. Like private broadcasters, both those \dots} \\
\docrow{7}{were produced live to air. Locally produced programs during the station's early days included ``Kids Bids'', ``The Three R's'', ``Romper Room'', ``Let's Visit'', ``Midday'', ``Focus'' and ``Okanagan Magazine''. In 1964, \dots} \\
\docrow{8}{its other E! owned-and-operated stations, stating that ``a second conventional TV network [was] no longer key to the long-term success'' of the company. Although for a time it was reported that CHBC might cease \dots} \\
\docrow{9}{via an aerial signal in the Hamilton-Toronto-Buffalo area on CHCH-DT Channel 18. CHCH retains this digital signal under its new ownership. In addition, CHBC, CHEK and CJNT have since converted to digital \dots} \\
\docrow{10}{is also a high definition feed available on Telus TV channel 692, and Shaw Direct Classic tier channel 7 and Advanced tier channel 507. The station first signed on the air on April 8, 1957 as CFCR-TV, \dots} \\
\docrow{11}{CHEX was originally a CBC affiliate, but after 60 years with the CBC it began airing CTV programming on August 31, 2015 under a programming supply agreement. On August 14, 2018, it was announced that CHEX's \dots} \\
\docrow{12}{[\dots] Kelowna's CHBC and \hlb{Kamloops's CFJC}, the latter owned by the Jim Pattison Group, \hlb{also disaffiliated from the CBC} in February 2006 and joined CH. Although CFJC was not owned by Canwest, \hlb{its joint sales agreement with CHBC necessitated its affiliation switch.}} \\
\docrow{13}{CKX-TV CKX-TV was a television station in Brandon, Manitoba, Canada, formerly affiliated with CBC Television. Owned and operated by CTVglobemedia, it was the first privately owned television station in \dots} \\
\docrow{14}{Durham Region. The station remained affiliated with CBC despite the fact its signal overlaps with that of the network's Toronto owned-and-operated station (O\&O) CBLT-DT; as a result, the Toronto market was \dots} \\
\docrow{15}{ownership of the station changed, beginning with the purchase of CKOK's one-third ownership by general manager Roy Chapman, which he later sold to CHAN. Selkirk Communications brought CJIB, and along with it, \dots} \\
\docrow{16}{[\dots] Most private affiliates produce their own local newscasts for a duration of at least 35 minutes. Some of the private affiliates later began adding CBC's overnight programming to their schedules since the network began broadcasting 24 hours a day in October 2006. Following the disaffiliation of the last privately-owned CBC affiliate CKSA-DT in Lloydminster on August 31, 2016, no more private stations operate as CBC affiliates [\dots]} \\
\docrow{17}{of media consolidation have always been allowed. In small markets where the population could not adequately support multiple television stations competing for advertising revenue, the CRTC began permitting \dots} \\
\docrow{18}{CHBC-DT CHBC-DT, virtual channel 2 (UHF digital channel 27), is a Global owned-and-operated television station located in Kelowna, British Columbia, Canada. The station is owned by Corus Entertainment. CHBC \dots} \\
\docrow{19}{Telecommunications Commission (CRTC) has significantly more lenient rules regarding media ownership. As such, most television stations, regardless of market size, are now O\&Os of their respective networks, \dots} \\
\docrow{20}{compete with CBC. The deal was approved by the Competition Bureau in March 2013, and by the CRTC in June 2013. On October 28, 2015, the CRTC made public an application by Bell to disaffiliate CJDC from CBC \dots} \\
\midrule
\textbf{Reranker} & \textbf{Selected documents} & \textbf{Rel.} & \textbf{Aut.} & \textbf{Qua.} & \textbf{Cmp.} & \textbf{Red.} & \textbf{Con.} & \textbf{Cpl.} & \textbf{Den.} & \textbf{Rea.} & \textbf{Doc.} & \textbf{Set} & \textbf{Glo.} & \textbf{Ovr.} \\ \midrule
RubricRanker (8B) & \hlok{[1]}~\hlbad{[3]}~\hlok{[12]}~\hlbad{[2]} & 2.5 & 2.5 & 2.0 & 1.0 & 2.0 & 4.0 & 4.0 & 2.0 & 4.0 & 2.33 & 2.33 & 3.33 & 2.67 \\
\textbf{AdaTutoRank (8B)} & \hlok{[1]}~\hlok{[12]} & \textbf{4.0} & \textbf{4.0} & \textbf{3.0} & \textbf{2.0} & \textbf{2.0} & \textbf{4.0} & \textbf{4.0} & \textbf{2.0} & \textbf{4.0} & \textbf{3.67} & \textbf{2.67} & \textbf{3.33} & \textbf{3.22} \\
\bottomrule
\end{tabular}
\vspace{-5pt}
\caption{Setwise-level case study on SetwiseEvalKit: the query-specific rubrics, the top-20 BM25 candidates, and each selector's set with its judge scores (0--4).}
\label{tab:case_study_rubric_musique}
\end{table*}

Table~\ref{tab:case_study_rubric_musique} inspects a single query under the full rubric. The two passages that establish the answer are $[1]$ and $[12]$, and both selectors retain them; the difference is that RubricRanker additionally forwards $[2]$ and $[3]$, which describe CHBC's own disaffiliation without naming CFJC-TV. These extras are free riders: the set stays complete and the answer stays reachable, so the two selectors receive an identical global-level score of $3.33$. What the extras do change is the other two levels. Document-level criteria are averaged over the set, so two passages that fail relevance and authenticity pull both scores from $4.0$ down to $2.5$, and since the extras contribute no distinct information, complementarity falls from $2.0$ to $1.0$. The overall score separates the two sets ($3.22$ against $2.67$) only because utility is scored at the document and set levels as well as globally; a judgement made on the final answer alone would have rated them equally.

\clearpage

\section{Details of Prompt} \label{appendix:prompt}

\prompt
{\method Prompt: RAG Setting}{
\systemprompt{
You are a high-quality document set selector for a retrieval-augmented QA system. Given a user question and a list of retrieved passages, select the best subset that jointly answers the question, then rank the selected passages from most to least useful.\\
Use these 9 rubrics to judge a candidate document set:\\
Relevance (doc-level, primary objective): Does the document directly address, at the topic level, the core information need of the query (naming the concrete subject/facet the query asks about), rather than merely matching surface keywords?\\
Authenticity (doc-level): Are the document's factual statements consistent with the correct reference facts (dates, numbers, names, causal conclusions) given by the query/answer? Drop documents that state falsehoods.\\
Quality (doc-level): Is the document clearly structured, focused, and information-dense enough that the target information can be directly extracted, rather than being empty, too shallow, noisy, or low-quality filler?\\
Complementarity (set-level): Do the retained documents complement one another so that, taken together, they cover the key information elements the answer requires more completely than any single document could? Prefer subsets whose documents jointly raise coverage.\\
Redundancy (set-level): Does the subset avoid two or more documents that convey essentially the same core information with no incremental value? When documents duplicate each other, keep only the single best one.\\
Conflict (set-level): Are the documents free of mutual factual contradiction on the same point, so the subset consistently supports the correct answer? Reject subsets that contain contradictory statements.\\
Completeness (global-level): Does the whole subset collectively supply every key information element the correct answer depends on, leaving no obvious gap?\\
Density (global-level): Across the subset, does the query-relevant content occupy the majority of each document's length, rather than the documents being bloated with irrelevant padding around a small useful snippet?\\
Reachability (global-level): Using only this subset and no external knowledge, can the full reasoning chain required to produce the correct answer be completed --- is every link grounded in the set?\\
Output format (STRICT): respond with only the selected document identifiers, each wrapped in square brackets and separated by a single space, ordered from most to least useful. Example: [4] [2] [7] [11] [1]. Do not output any other text or explanation.
}
\userprompt{
I will provide you with \prompttag{\textbf{\color{insert}<NUM>}} passages, each indicated by a numerical identifier []. Select and rank the passages based on their joint usefulness for the user question: \prompttag{\textbf{\color{insert}<QUESTION>}}.\\
$[$PASSAGES$]$\\
\prompttag{\textbf{\color{insert}<CONTEXT>}}\\
User Question: \prompttag{\textbf{\color{insert}<QUESTION>}}.\\
Select the best subset of the \prompttag{\textbf{\color{insert}<NUM>}} passages above so that, judged by the 9 dimensions rubrics, the selected set best answers the question. Then rank the selected passages from most to least useful. The number of selected passages is at most 10 and should not include redundant ones. The output must be only the selected identifiers in ranked order, e.g. [2] [1] [4]. Only respond with the selection; do not say any other word and do not explain.
}}%

\prompt
{\method Prompt: Deep-Research Setting}{
\systemprompt{
You are a high-quality document set selector for a deep-research agent. Given a search query from the agent, the agent's reasoning as query intent, and a list of retrieved documents, select the best subset that jointly answers the query, then rank the selected documents from most to least useful.\\
The query intent may clarify why this query was issued and what information it is trying to find, but it may also contain broader, exploratory, or irrelevant thoughts. Stay anchored on the query itself and only use the parts of the query intent that are actually relevant to this query.\\
Use these 9 rubrics to judge a candidate document set:\\
Relevance (doc-level, primary objective): Does the document directly address, at the topic level, the core information need of the query (naming the concrete subject/facet the query asks about), rather than merely matching surface keywords?\\
Authenticity (doc-level): Are the document's factual statements consistent with the correct reference facts (dates, numbers, names, causal conclusions) given by the query/answer? Drop documents that state falsehoods.\\
Quality (doc-level): Is the document clearly structured, focused, and information-dense enough that the target information can be directly extracted, rather than being empty, too shallow, noisy, or low-quality filler?\\
Complementarity (set-level): Do the retained documents complement one another so that, taken together, they cover the key information elements the answer requires more completely than any single document could? Prefer subsets whose documents jointly raise coverage.\\
Redundancy (set-level): Does the subset avoid two or more documents that convey essentially the same core information with no incremental value? When documents duplicate each other, keep only the single best one.\\
Conflict (set-level): Are the documents free of mutual factual contradiction on the same point, so the subset consistently supports the correct answer? Reject subsets that contain contradictory statements.\\
Completeness (global-level): Does the whole subset collectively supply every key information element the correct answer depends on, leaving no obvious gap?\\
Density (global-level): Across the subset, does the query-relevant content occupy the majority of each document's length, rather than the documents being bloated with irrelevant padding around a small useful snippet?\\
Reachability (global-level): Using only this subset and no external knowledge, can the full reasoning chain required to produce the correct answer be completed --- is every link grounded in the set?\\
Output format (STRICT): respond with only the selected document identifiers, each wrapped in square brackets and separated by a single space, ordered from most to least useful. Example: [4] [2] [7] [11] [1]. Do not output any other text or explanation.
}
\userprompt{
I will provide you with \prompttag{\textbf{\color{insert}<NUM>}} passages, each indicated by a numerical identifier []. Select and rank the passages based on their joint usefulness for the search query: \prompttag{\textbf{\color{insert}<QUERY>}}.\\
$[$QUERY INTENT$]$\\
\prompttag{\textbf{\color{insert}<QUERY\_INTENT>}}\\
$[$PASSAGES$]$\\
\prompttag{\textbf{\color{insert}<CONTEXT>}}\\
Search Query: \prompttag{\textbf{\color{insert}<QUERY>}}.\\
Select the best subset of the \prompttag{\textbf{\color{insert}<NUM>}} passages above so that, judged by the 9 dimensions rubrics, the selected set best answers the query (anchored on the query, with the query intent only as clarification). Then rank the selected passages from most to least useful. The number of selected passages is at most 10 and should not include redundant ones. The output must be only the selected identifiers in ranked order, e.g. [2] [1] [4]. Only respond with the selection; do not say any other word and do not explain.
}}%

\prompt
{Rubric Generation Prompt: Short-form Scenario}{
\systemprompt{
You are a senior rubric-generation expert. Given a (query, answer) pair, you generate binary (Yes/No) evaluation rubrics along nine dimensions. These rubrics are used to judge whether a candidate document set can effectively support answering the query and producing the correct answer.\\
The nine dimensions come in three families, and each family is scoped differently. Respect the scope of the family a dimension belongs to:\\
- Doc-level (Relevance, Authenticity, Quality): each rubric judges EACH individual document in the set. Phrase them as ``Does each document \dots''.\\
- Set-level (Complementarity, Redundancy, Conflict): each rubric judges the relationships BETWEEN documents in the set. Phrase them as ``Does the document set \dots'' / ``Across the documents, \dots''.\\
- Global-level (Completeness, Density, Reachability): each rubric judges the ENTIRE set as one unit --- its overall utility for answering the query. Phrase them as ``Does the whole document set \dots''.\\
\# Rubric requirements\\
Every rubric you generate MUST satisfy all of the following:\\
- Relevant: It must be directly tied to the core information need of the query and help produce the answer, so that it meaningfully reduces uncertainty when judging quality.\\
- Binary: It must be a Yes/No question. No degree-based phrasing (e.g., ``to what extent'').\\
- Qualitative \& query-specific: It must be purpose-built for this specific (query, answer). You MUST name the concrete entities, facts, dates, or numbers from the query/answer. Never use vague wording such as ``relevant content'' or ``important information''.\\
- Grounded: Its judgment criterion must be derivable from the given (query, answer); a judge should not need external knowledge to answer it.\\
- Salient: It should focus on the aspects an experienced information-retrieval expert would care about.\\
- Correctly scoped: It must match the scope (doc-level / set-level / global-level) of its own dimension, as defined above.\\
- Weighted: It must carry an integer weight from 1 to 5 stating how important this rubric is for judging whether the document set can support answering THIS query.\\
\# Weight scale\\
Assign the weight by asking: if the document set fails this rubric, how badly is the query damaged?\\
- 1 = marginally useful; minor added value only\\
- 2 = useful but secondary\\
- 3 = moderately important; clearly helpful for evaluating a good document set\\
- 4 = very important; strong effect on document set quality\\
- 5 = critical; if missing or violated, the document set is seriously inadequate for answering the query\\
Weight the rubric, not the dimension: two rubrics under the same dimension can differ in weight, and a rubric under a ``minor'' dimension can outweigh one under a ``major'' dimension if this particular query depends on it more. Spread the weights --- do not label everything 4 or 5. Reserve 5 for the few rubrics that are genuinely make-or-break for this query, and use 1-2 for the nice-to-have ones.\\
\# Dimensions\\
\#\# Doc-level\\
- Relevance: Measures how semantically related each document's content is to the core information need of the query --- whether the document directly addresses, at the topic level, what the query is asking, rather than merely matching surface keywords.\\
Requirement: The rubric must name the concrete subject and information facet the query asks about (e.g., specific people / works / events / time dimension) and judge whether the document truly discusses this specific question, not merely whether it is ``relevant''.\\
Goal: Keep documents that directly answer the user's question; discard off-topic or keyword-only, semantically irrelevant documents.\\
- Authenticity: Measures how well the facts stated in each document conform to objective real-world information.\\
Requirement: The rubric must spell out the correct reference fact (taken from the answer or the deterministic information in the query) and judge whether the document's statements are consistent with it, rather than asking whether the document ``contains correct information''. Only generate this dimension when the query/answer involves concrete, verifiable facts (e.g., dates, numbers, names, causal conclusions); otherwise return an empty list for it.\\
Goal: Discard documents with false statements or factual errors.\\
- Quality: Measures how usable each document is in terms of information organization and presentation --- whether it is clearly structured, focused, and lets the reader directly extract the target information, rather than being poorly formatted, too short / shallow, noisy, or low-quality filler.\\
Requirement: The rubric must reference the specific information a reader needs to extract in order to answer this query (name the concrete entities, not generic terms), and judge whether the document makes that information clearly accessible. This dimension evaluates presentation quality only, not factual correctness.\\
Goal: Discard low-quality documents (empty content, low information density, unclear phrasing, etc.).\\
\#\# Set-level\\
- Complementarity: Measures the degree to which the documents in the set complement one another --- whether, taken together, they cover all the key information elements the answer requires and jointly build a more complete, well-rounded answer than any single document could (e.g., one document gives an overview while another gives the details; one covers 70\% of the must-have points while another covers the remaining 30\%).\\
Requirement: The rubric must name the concrete key information elements the query/answer requires, and judge whether the set as a whole covers them across its documents. Distinguish complementarity (documents combine to increase information value) from conflict (documents contradict each other).\\
Goal: Reward sets whose documents jointly raise coverage of the query's information need when no single document fully satisfies it.\\
- Redundancy: Measures the degree of content redundancy among the documents in the set --- whether two or more documents convey essentially the same core information, data, or conclusions without any incremental contribution (common when documents merely reprint, summarize, rewrite, or re-report the same source).\\
Requirement: The rubric must name the concrete query-specific information, and judge whether the set avoids having two or more documents that carry essentially identical information with no incremental value (in which case only the single best --- highest information density, most concise, clearest --- should be kept).\\
Goal: Discourage sets that pad in duplicated documents instead of adding new information.\\
- Conflict: Measures the degree of factual consistency across the documents in the set --- whether different documents give mutually contradictory factual statements about the same point (e.g., two different values for the same fact, one correct and one wrong).\\
Requirement: The rubric must name the concrete fact at issue and judge whether the documents' statements are free of mutual contradiction, so that the set consistently supports the correct answer. A high-quality set must never contain factual contradictions. Only generate this dimension when the query/answer involves concrete, verifiable facts that documents could contradict; otherwise return an empty list for it.\\
Goal: Reject sets containing factual contradictions or documents with erroneous information.\\
\#\# Global-level\\
- Completeness: Measures whether the document set, as a whole, can COMPLETELY answer the query with no obvious information gaps --- every key information element the correct answer depends on is present somewhere in the set.\\
Requirement: The rubric must enumerate the concrete key information elements the query/answer requires and judge whether the whole set collectively supplies all of them, leaving no obvious gap that would prevent producing the correct answer.\\
Goal: Reward sets that fully cover the query's information need; reject sets that miss essential pieces.\\
- Density: Measures, in terms of LENGTH / span, how much of each document is actually useful for answering this query --- i.e., the ratio of useful content to the document's total length. It is NOT about whether a document is removable, and NOT about whether the set as a whole is complete. It focuses on the internal ``signal-to-length'' ratio of the documents.\\
- What we WANT (high Density): each document in the set is compact and on-point, so that the text carrying the query-relevant information (name the concrete query-specific facts/entities/dates) occupies the MAJORITY of that document's length.\\
- What we DO NOT want (low Density): a document that is long but mostly filler --- the bulk of its length is off-topic / unrelated content, and only a small fraction of the text actually contributes useful information for this query.\\
Requirement: The rubric must name the concrete query-specific useful information, and judge whether --- across the documents in the set --- the useful content takes up the main portion of each document's length, rather than the documents being padded with large amounts of irrelevant text where only a small snippet is useful.\\
Goal: Reward sets whose documents are dense with useful content; penalize sets containing bloated documents that are mostly irrelevant padding with only a sliver of useful information.\\
- Reachability: Measures whether a model WITHOUT any external knowledge could complete the full reasoning chain required to produce the correct answer using ONLY the document set. Every link of the inference chain must be grounded in the set.\\
Requirement: The rubric must name the concrete reasoning steps / intermediate facts the query requires (e.g., the bridge entities and the final comparison), and judge whether the whole set self-containedly provides every link so the correct answer is derivable without outside knowledge.\\
Goal: Reward sets that are self-contained for the full reasoning chain; reject sets with a missing link that forces reliance on external knowledge.\\
\# Additional criteria\\
- Knowledge-cutoff: Do not ask for or rely on information that would violate a knowledge cutoff date.\\
- Coverage \& non-redundancy: The rubrics together should cover the important aspects of the query without redundant or duplicate questions. In particular, do not restate the same question under two different dimensions --- each dimension must contribute a distinct judgment.\\
- Generate at most 3 questions per dimension (fewer is fine). Every question must be important, useful, and non-redundant.\\
\# Example\\
Query: Which was founded earlier, Arthur's Magazine or First for Women?\\
Answer: Arthur's Magazine\\
{\ttfamily
\{\\
\hspace*{1em}"Relevance": [\\
\hspace*{2em}\{"rubric": "Does each document directly discuss the founding/launch date or early publication history of Arthur's Magazine or First for Women, rather than merely containing words like 'Arthur', 'magazine', or 'women' without addressing founding dates?", "weight": 5\}\\
\hspace*{1em}],\\
\hspace*{1em}"Authenticity": [\\
\hspace*{2em}\{"rubric": "For each document, are its founding-date statements consistent with objective fact - namely Arthur's Magazine = 1844 and First for Women = 1989 - without misstating the years or attributing them to the wrong magazine?", "weight": 5\},\\
\hspace*{2em}\{"rubric": "For each document that states which was founded first, is that conclusion consistent with the fact that Arthur's Magazine predates First for Women, rather than claiming First for Women is older?", "weight": 3\}\\
\hspace*{1em}],\\
\hspace*{1em}"Quality": [\\
\hspace*{2em}\{"rubric": "Is each document clearly structured and focused enough that a reader can directly extract the founding year of Arthur's Magazine and/or First for Women (or the conclusion that Arthur's Magazine came first), rather than having these key dates buried under large amounts of irrelevant content?", "weight": 2\}\\
\hspace*{1em}],\\
\hspace*{1em}"Complementarity": [\\
\hspace*{2em}\{"rubric": "Does the document set contain at least one document that explicitly records the founding year of Arthur's Magazine AND at least one that explicitly records the founding year of First for Women, so that the two can be compared to determine which was founded first?", "weight": 5\}\\
\hspace*{1em}],\\
\hspace*{1em}"Redundancy": [\\
\hspace*{2em}\{"rubric": "Does the document set avoid having two or more documents that convey essentially the same founding-date information (e.g., all merely stating 'Arthur's Magazine was founded in 1844') without any incremental information contribution?", "weight": 2\}\\
\hspace*{1em}],\\
\hspace*{1em}"Conflict": [\\
\hspace*{2em}\{"rubric": "Are the statements across the document set about the two magazines' founding dates free of mutual contradiction (e.g., one saying 1844 and another saying 1850), so that they consistently support the conclusion that Arthur's Magazine was founded earlier?", "weight": 4\}\\
\hspace*{1em}],\\
\hspace*{1em}"Completeness": [\\
\hspace*{2em}\{"rubric": "Does the document set as a whole provide the founding year of BOTH Arthur's Magazine and First for Women, leaving no gap that would prevent determining which was founded earlier?", "weight": 5\}\\
\hspace*{1em}],\\
\hspace*{1em}"Density": [\\
\hspace*{2em}\{"rubric": "Across the documents in the set, does the text that actually states the founding date(s) of Arthur's Magazine and First for Women make up the majority of each document's length, rather than each document being dominated by unrelated content (e.g., editorial history, staff, or coverage of other magazines) with only a small fraction of its length mentioning those founding dates?", "weight": 2\}\\
\hspace*{1em}],\\
\hspace*{1em}"Reachability": [\\
\hspace*{2em}\{"rubric": "Using only the document set and no external knowledge, can the full reasoning chain be completed --- obtaining the founding year of Arthur's Magazine, obtaining the founding year of First for Women, and comparing them --- to reach the conclusion that Arthur's Magazine was founded earlier?", "weight": 4\}\\
\hspace*{1em}]\\
\}
}\\
\# Output format\\
Output ONLY a valid JSON object (no markdown code fences, no extra commentary) with exactly these nine keys: ``Relevance'', ``Authenticity'', ``Quality'', ``Complementarity'', ``Redundancy'', ``Conflict'', ``Completeness'', ``Density'', ``Reachability''. Each key maps to a list of objects, each object having exactly two fields: ``rubric'' (the Yes/No question string) and ``weight'' (an integer from 1 to 5). A dimension may have multiple rubrics. If a dimension does not apply (e.g., Authenticity or Conflict for a query with no verifiable facts), use an empty list [] --- but still include the key.
}
\userprompt{
Query: \prompttag{\textbf{\color{insert}<QUESTION>}}\\
Answer: \prompttag{\textbf{\color{insert}<ANSWER>}}\\
Rubrics (JSON):
}}%
\prompt
{Rubric Generation Prompt: Long-form Scenario}{
\systemprompt{
You are a senior rubric-generation expert. Given a (query, answer) pair, you generate binary (Yes/No) evaluation rubrics along nine dimensions. These rubrics are used to judge whether a candidate document set can effectively support answering the query and producing the correct answer.\\
The nine dimensions come in three families, and each family is scoped differently. Respect the scope of the family a dimension belongs to:\\
- Doc-level (Relevance, Authenticity, Quality): each rubric judges EACH individual document in the set. Phrase them as ``Does each document \dots''.\\
- Set-level (Complementarity, Redundancy, Conflict): each rubric judges the relationships BETWEEN documents in the set. Phrase them as ``Does the document set \dots'' / ``Across the documents, \dots''.\\
- Global-level (Completeness, Density, Reachability): each rubric judges the ENTIRE set as one unit --- its overall utility for answering the query. Phrase them as ``Does the whole document set \dots''.\\
\# Rubric requirements\\
Every rubric you generate MUST satisfy all of the following:\\
- Relevant: It must be directly tied to the core information need of the query and help produce the answer, so that it meaningfully reduces uncertainty when judging quality.\\
- Binary: It must be a Yes/No question. No degree-based phrasing (e.g., ``to what extent'').\\
- Qualitative \& query-specific: It must be purpose-built for this specific (query, answer). You MUST name the concrete entities, facts, dates, or numbers from the query/answer. Never use vague wording such as ``relevant content'' or ``important information''.\\
- Grounded: Its judgment criterion must be derivable from the given (query, answer); a judge should not need external knowledge to answer it.\\
- Salient: It should focus on the aspects an experienced information-retrieval expert would care about.\\
- Correctly scoped: It must match the scope (doc-level / set-level / global-level) of its own dimension, as defined above.\\
- Weighted: It must carry an integer weight from 1 to 5 stating how important this rubric is for judging whether the document set can support answering THIS query.\\
\# Weight scale\\
Assign the weight by asking: if the document set fails this rubric, how badly is the query damaged?\\
- 1 = marginally useful; minor added value only\\
- 2 = useful but secondary\\
- 3 = moderately important; clearly helpful for evaluating a good document set\\
- 4 = very important; strong effect on document set quality\\
- 5 = critical; if missing or violated, the document set is seriously inadequate for answering the query\\
Weight the rubric, not the dimension: two rubrics under the same dimension can differ in weight, and a rubric under a ``minor'' dimension can outweigh one under a ``major'' dimension if this particular query depends on it more. Spread the weights --- do not label everything 4 or 5. Reserve 5 for the few rubrics that are genuinely make-or-break for this query, and use 1-2 for the nice-to-have ones.\\
\# Dimensions\\
\#\# Doc-level\\
- Relevance: Measures how semantically related each document's content is to the core information need of the query --- whether the document directly addresses, at the topic level, what the query is asking, rather than merely matching surface keywords.\\
Requirement: The rubric must name the concrete subject and information facet the query asks about (e.g., specific people / works / events / time dimension) and judge whether the document truly discusses this specific question, not merely whether it is ``relevant''.\\
Goal: Keep documents that directly answer the user's question; discard off-topic or keyword-only, semantically irrelevant documents.\\
- Authenticity: Measures how well the facts stated in each document conform to objective real-world information.\\
Requirement: The rubric must spell out the correct reference fact (taken from the answer or the deterministic information in the query) and judge whether the document's statements are consistent with it, rather than asking whether the document ``contains correct information''. Only generate this dimension when the query/answer involves concrete, verifiable facts (e.g., dates, numbers, names, causal conclusions); otherwise return an empty list for it.\\
Goal: Discard documents with false statements or factual errors.\\
- Quality: Measures how usable each document is in terms of information organization and presentation --- whether it is clearly structured, focused, and lets the reader directly extract the target information, rather than being poorly formatted, too short / shallow, noisy, or low-quality filler.\\
Requirement: The rubric must reference the specific information a reader needs to extract in order to answer this query (name the concrete entities, not generic terms), and judge whether the document makes that information clearly accessible. This dimension evaluates presentation quality only, not factual correctness.\\
Goal: Discard low-quality documents (empty content, low information density, unclear phrasing, etc.).\\
\#\# Set-level\\
- Complementarity: Measures the degree to which the documents in the set complement one another --- whether, taken together, they cover all the key information elements the answer requires and jointly build a more complete, well-rounded answer than any single document could (e.g., one document gives an overview while another gives the details; one covers 70\% of the must-have points while another covers the remaining 30\%).\\
Requirement: The rubric must name the concrete key information elements the query/answer requires, and judge whether the set as a whole covers them across its documents. Distinguish complementarity (documents combine to increase information value) from conflict (documents contradict each other).\\
Goal: Reward sets whose documents jointly raise coverage of the query's information need when no single document fully satisfies it.\\
- Redundancy: Measures the degree of content redundancy among the documents in the set --- whether two or more documents convey essentially the same core information, data, or conclusions without any incremental contribution (common when documents merely reprint, summarize, rewrite, or re-report the same source).\\
Requirement: The rubric must name the concrete query-specific information, and judge whether the set avoids having two or more documents that carry essentially identical information with no incremental value (in which case only the single best --- highest information density, most concise, clearest --- should be kept).\\
Goal: Discourage sets that pad in duplicated documents instead of adding new information.\\
- Conflict: Measures the degree of factual consistency across the documents in the set --- whether different documents give mutually contradictory factual statements about the same point (e.g., two different values for the same fact, one correct and one wrong).\\
Requirement: The rubric must name the concrete fact at issue and judge whether the documents' statements are free of mutual contradiction, so that the set consistently supports the correct answer. A high-quality set must never contain factual contradictions. Only generate this dimension when the query/answer involves concrete, verifiable facts that documents could contradict; otherwise return an empty list for it.\\
Goal: Reject sets containing factual contradictions or documents with erroneous information.\\
\#\# Global-level\\
- Completeness: Measures whether the document set, as a whole, can COMPLETELY answer the query with no obvious information gaps --- every key information element the correct answer depends on is present somewhere in the set.\\
Requirement: The rubric must enumerate the concrete key information elements the query/answer requires and judge whether the whole set collectively supplies all of them, leaving no obvious gap that would prevent producing the correct answer.\\
Goal: Reward sets that fully cover the query's information need; reject sets that miss essential pieces.\\
- Density: Measures, in terms of LENGTH / span, how much of each document is actually useful for answering this query --- i.e., the ratio of useful content to the document's total length. It is NOT about whether a document is removable, and NOT about whether the set as a whole is complete. It focuses on the internal ``signal-to-length'' ratio of the documents.\\
- What we WANT (high Density): each document in the set is compact and on-point, so that the text carrying the query-relevant information (name the concrete query-specific facts/entities/dates) occupies the MAJORITY of that document's length.\\
- What we DO NOT want (low Density): a document that is long but mostly filler --- the bulk of its length is off-topic / unrelated content, and only a small fraction of the text actually contributes useful information for this query.\\
Requirement: The rubric must name the concrete query-specific useful information, and judge whether --- across the documents in the set --- the useful content takes up the main portion of each document's length, rather than the documents being padded with large amounts of irrelevant text where only a small snippet is useful.\\
Goal: Reward sets whose documents are dense with useful content; penalize sets containing bloated documents that are mostly irrelevant padding with only a sliver of useful information.\\
- Reachability: Measures whether a model WITHOUT any external knowledge could complete the full reasoning chain required to produce the correct answer using ONLY the document set. Every link of the inference chain must be grounded in the set.\\
Requirement: The rubric must name the concrete reasoning steps / intermediate facts the query requires (e.g., the bridge entities and the final comparison), and judge whether the whole set self-containedly provides every link so the correct answer is derivable without outside knowledge.\\
Goal: Reward sets that are self-contained for the full reasoning chain; reject sets with a missing link that forces reliance on external knowledge.\\
\# Additional criteria\\
- Knowledge-cutoff: Do not ask for or rely on information that would violate a knowledge cutoff date.\\
- Coverage \& non-redundancy: The rubrics together should cover the important aspects of the query without redundant or duplicate questions. In particular, do not restate the same question under two different dimensions --- each dimension must contribute a distinct judgment.\\
- Generate at most 3 questions per dimension (fewer is fine). Every question must be important, useful, and non-redundant.\\
\# Example\\
Query: How does strain-softening behavior in sensitive clays differ from that in non-sensitive clays in terms of pore pressure development, shear strength, and strain effects?\\
Answer: In sensitive (e.g., quick) clays, undrained shearing produces POSITIVE excess pore pressures due to their contractive, collapsible structure, driving rapid strength loss through pore-pressure-induced softening, with strain tending to localize into shear bands. In contrast, non-sensitive / overconsolidated clays may develop NEGATIVE excess pore pressures (dilative tendency), show a more gradual shear-strength change, and deform more diffusely.\\
{\ttfamily
\{\\
\hspace*{1em}"Relevance": [\\
\hspace*{2em}\{"rubric": "Does each document directly address the strain-softening behavior of clays in terms of pore pressure development, shear-strength degradation, or strain localization (for sensitive and/or non-sensitive clays), rather than merely mentioning 'clay' or general soil mechanics without discussing strain-softening?", "weight": 5\}\\
\hspace*{1em}],\\
\hspace*{1em}"Authenticity": [\\
\hspace*{2em}\{"rubric": "For each document, are its stated mechanisms consistent with established soil mechanics - namely that sensitive/quick clays generate positive excess pore pressure (contractive) under undrained shear while overconsolidated non-sensitive clays can develop negative excess pore pressure (dilative) - rather than asserting the opposite relationship?", "weight": 5\},\\
\hspace*{2em}\{"rubric": "For each document that describes strain effects, is its account consistent with strain localizing into shear bands in sensitive clays versus more diffuse deformation in non-sensitive clays, rather than reversing the two?", "weight": 3\}\\
\hspace*{1em}],\\
\hspace*{1em}"Quality": [\\
\hspace*{2em}\{"rubric": "Is each document clearly structured and focused enough that a reader can directly extract the comparison of pore-pressure behavior, shear-strength change, and strain effects between sensitive and non-sensitive clays, rather than having these buried under large amounts of unrelated content?", "weight": 2\}\\
\hspace*{1em}],\\
\hspace*{1em}"Complementarity": [\\
\hspace*{2em}\{"rubric": "Does the document set, taken together, cover all three required aspects - pore pressure development, shear-strength degradation, and strain/localization effects - for BOTH sensitive and non-sensitive clays (e.g., different documents supplying different aspects), so that the documents jointly build the full comparison rather than any single aspect being missing across the set?", "weight": 5\}\\
\hspace*{1em}],\\
\hspace*{1em}"Redundancy": [\\
\hspace*{2em}\{"rubric": "Does the document set avoid having two or more documents that convey essentially the same finding (e.g., several documents merely restating that sensitive clays develop positive excess pore pressure under undrained shear) without adding any incremental aspect such as shear-strength behavior, strain localization, or the non-sensitive-clay contrast?", "weight": 2\}\\
\hspace*{1em}],\\
\hspace*{1em}"Conflict": [\\
\hspace*{2em}\{"rubric": "Are the documents in the set free of mutual contradiction on the key mechanisms - e.g., not one stating that sensitive clays develop positive excess pore pressure under undrained shear while another asserts negative excess pore pressure for the same condition - so that the set consistently supports the correct sensitive-vs-non-sensitive comparison?", "weight": 4\}\\
\hspace*{1em}],\\
\hspace*{1em}"Completeness": [\\
\hspace*{2em}\{"rubric": "Does the document set as a whole cover every key information element the answer requires - pore pressure development, shear-strength degradation, and strain/localization effects for BOTH sensitive and non-sensitive clays - leaving no obvious gap that would prevent producing the full sensitive-vs-non-sensitive comparison?", "weight": 5\}\\
\hspace*{1em}],\\
\hspace*{1em}"Density": [\\
\hspace*{2em}\{"rubric": "Across the documents in the set, does the text that actually describes the strain-softening mechanisms (pore pressure, shear strength, and strain effects for sensitive vs non-sensitive clays) make up the majority of each document's length, rather than each document being dominated by unrelated content with only a small fraction of its length addressing these mechanisms?", "weight": 2\}\\
\hspace*{1em}],\\
\hspace*{1em}"Reachability": [\\
\hspace*{2em}\{"rubric": "Using only the document set and no external knowledge, can the full reasoning chain be completed --- establishing that sensitive clays develop positive excess pore pressure and rapid, localized strength loss, establishing the contrasting behavior of non-sensitive/overconsolidated clays (negative excess pore pressure, gradual strength change, diffuse deformation), and contrasting the two --- to reach the correct differentiation?", "weight": 4\}\\
\hspace*{1em}]\\
\}
}\\
\# Output format\\
Output ONLY a valid JSON object (no markdown code fences, no extra commentary) with exactly these nine keys: ``Relevance'', ``Authenticity'', ``Quality'', ``Complementarity'', ``Redundancy'', ``Conflict'', ``Completeness'', ``Density'', ``Reachability''. Each key maps to a list of objects, each object having exactly two fields: ``rubric'' (the Yes/No question string) and ``weight'' (an integer from 1 to 5). A dimension may have multiple rubrics. If a dimension does not apply (e.g., Authenticity or Conflict for a query with no verifiable facts), use an empty list [] --- but still include the key.
}
\userprompt{
Query: \prompttag{\textbf{\color{insert}<QUESTION>}}\\
Answer: \prompttag{\textbf{\color{insert}<ANSWER>}}\\
Rubrics (JSON):
}}%

\prompt
{Rubric-based Judge Scoring Prompt: Short-form Scenario}{
\systemprompt{
You are a senior search-result quality assessor. Given a user query, a candidate document set (doc set, variable size) and a list of rubrics (each labeled with its dimension type and judgment content), you must score every rubric on a 0--10 scale, strictly grounded in the document content.\\
This round evaluates all 9 dimensions at once, grouped into three categories:\\
{[}A. Doc-Level dimensions] --- judged for EACH individual document in the doc set: Relevance, Authenticity, Quality.\\
{[}B. Set-Level dimensions] --- judged for the ENTIRE doc set as a whole; give ONE overall score, not per-document: Complementarity, Redundancy, Conflict.\\
{[}C. Global-Level dimensions] --- also judged for the ENTIRE doc set as a whole; give ONE overall score, not per-document: Completeness, Density, Reachability.\\
\# Absolute prerequisite: you MUST read the documents first\\
- Every score must be grounded in the actual document text. Judge only based on what ACTUALLY appears in the documents; do not use your own external knowledge to fill in information the documents do not state.\\
- The rubric text is itself query-specific (it already contains the concrete entities, facts, and numbers of the question); the user query is also provided above for context. Judge directly against the rubric, grounded in the documents.\\
\# Evaluation procedure --- RELEVANCE GATE (read carefully, follow the order)\\
Think step by step in this order:\\
1. FIRST evaluate the Relevance dimension for EVERY document in the doc set (give each document its per\_doc Relevance score).\\
2. Define the RELEVANT SUBSET = the documents whose Relevance score >= 1 (i.e. documents that actually touch the query's key information point(s)). Documents with Relevance = 0 are OFF-TOPIC and are EXCLUDED from EVERY other dimension: they NEVER count as providing coverage, as a distinct information point, as useful length, or as a comparable statement, and they NEVER count toward ``no redundancy'' or ``no conflict''.\\
3. Then check the relevance outcome:\\
\hspace*{1em}- CASE A --- the RELEVANT SUBSET is EMPTY (every document's Relevance score = 0): the documents this ranker selected are all useless for this question, so the other 8 dimensions (Authenticity, Quality, Complementarity, Redundancy, Conflict, Completeness, Density, Reachability) carry NO evaluation meaning. In this case DO NOT score them --- output each of those 8 dimensions as null (see output format). Only the Relevance dimension is scored.\\
\hspace*{1em}- CASE B --- the RELEVANT SUBSET is non-empty (at least ONE document has Relevance >= 1): evaluate the other 8 dimensions ONLY over the RELEVANT SUBSET, and apply each dimension's own APPLICABILITY PREREQUISITE below. A relevance-grounded dimension whose prerequisite is not met over the relevant subset (e.g. it needs a comparison / division of labour among relevant documents but there are not enough relevant documents, or no relevant document actually carries the key-information-point content it measures) MUST be output as null --- NOT as a high score. Never reward a dimension merely because off-topic documents trivially ``don't conflict / don't overlap / don't dilute''.\\
4. Never leave Relevance itself null --- Relevance is always scored.\\
\# Scoring scale (0--10)\\
Overall meaning (higher = better satisfies the rubric). Use the following anchors, and you MAY use intermediate integers (0--10) for in-between cases:\\
- 10 = Completely: fully satisfied, with sufficient detail and evidence.\\
- 8 = Mostly: largely satisfied, but missing some detail or support.\\
- 5 = Moderately: relevant content is mentioned, but key details are missing.\\
- 3 = Barely: not explicitly stated, only weakly inferable.\\
- 0 = Not at all: not addressed at all / irrelevant to the rubric / contradicts the facts.\\
\#\# I. Doc-Level dimensions: score EACH document individually\\
- Relevance: to what extent the document DIRECTLY discusses the specific information facet the rubric points to.\\
\hspace*{1em}- 10: the document's core content directly and fully discusses this specific question and can directly support the answer.\\
\hspace*{1em}- 8: directly discusses the topic, but the information is incomplete or unfocused.\\
\hspace*{1em}- 5: mentions the related topic only in passing, missing key information.\\
\hspace*{1em}- 3: barely discusses it directly, relevance is only weakly inferable.\\
\hspace*{1em}- 0: entirely off-topic, merely matching keywords or unrelated to the question.\\
- Authenticity: how well the facts stated in the document CONFORM to the reference fact given by the rubric. Only assess whether the STATEMENTS that appear are accurate; independent of relevance.\\
\hspace*{1em}- 10: the document explicitly states the relevant fact, fully consistent with the reference fact.\\
\hspace*{1em}- 8: the statement is consistent with the fact, but phrased unclearly or somewhat vaguely.\\
\hspace*{1em}- 5: partially correct, or the fact is right but wrapped in easily-misleading phrasing.\\
\hspace*{1em}- 3: only faintly touched upon; correctness is hard to judge or slightly off.\\
\hspace*{1em}- 0: the statement contradicts the reference fact, has an obvious factual error, or the document makes no statement related to that fact at all.\\
- Quality: how easy the document makes it for a reader to EXTRACT the target information, in terms of information organization and presentation. Assess presentation quality only, not factual correctness.\\
\hspace*{1em}- 10: clearly structured and focused; the target information is immediately apparent.\\
\hspace*{1em}- 8: mostly clear; the target information is extractable but mixed with some irrelevant content.\\
\hspace*{1em}- 5: the target information exists but is diluted by lots of irrelevant content and takes effort to find.\\
\hspace*{1em}- 3: messy layout, high noise; the target information is very hard to extract.\\
\hspace*{1em}- 0: empty, extremely noisy, or nearly unparseable content, OR the document is entirely off-topic and contains no target information at all.\\
\#\# II. Set-Level dimensions: give ONE overall score for the WHOLE doc set\\
- Complementarity: whether the key information points are covered through DIVISION OF LABOUR across DIFFERENT documents --- i.e., different documents each contribute a DIFFERENT needed piece (one gives an overview, another the details; one supplies point P1, another supplies point P2), so that the whole is more valuable than any single document. APPLICABILITY PREREQUISITE (judge this FIRST): Complementarity is judged ONLY over the RELEVANT SUBSET (Relevance >= 1) and ONLY on the query-required key information points; off-topic documents (Relevance = 0) NEVER count as a contributing part. If FEWER THAN TWO relevant documents each contribute a DIFFERENT required point (there is at most one relevant document, or all relevant documents only carry the SAME single point), there is no possible cross-document division of labour on the key information points --- Complementarity is NOT APPLICABLE: output it as null (NOT a high score). IMPORTANT: this dimension judges only the complementary STRUCTURE among the RELEVANT documents; it does NOT judge whether the set is complete. Do NOT lower the score merely because some required point is missing (that is Completeness's job) --- an incomplete set can still be highly complementary if the points it DOES cover come from different documents. Conversely, if a SINGLE document already covers everything on its own, complementarity is LOW even though coverage may be perfect (there is no division of labour).\\
\hspace*{1em}- 10: strong complementarity --- multiple documents clearly divide the labour, each contributing a DIFFERENT needed point, so the whole is far more valuable than any single document.\\
\hspace*{1em}- 8: mostly complementary --- most of the covered points are supplied by different documents; only an occasional point is carried alone by one document.\\
\hspace*{1em}- 5: partial complementarity --- there is some division of labour, but a large share of the covered content is concentrated in a single document.\\
\hspace*{1em}- 3: very weak complementarity --- almost all useful content comes from one document, or several documents merely restate the same single point.\\
\hspace*{1em}- 0: no complementarity at all --- only one document contributes any useful content (the others add nothing), OR a single document already covers everything by itself so no cross-document complementarity exists.\\
- Redundancy: judged against the MINIMAL document subset needed to cover the key information points the answer requires (the SAME key information points as Completeness). It checks whether each document brings INCREMENTAL coverage of those required information points, or merely repeats a point already covered by another document. If two or more documents cover the SAME required information point while adding no new required point, that overlap is redundancy --- those documents could be dropped from the minimal covering set without losing any required point. (Example: if answering the query needs 2 information points, and both doc {[}1] and doc {[}5] only answer the SAME single point, that is redundancy.) Higher score = less redundancy (each document adds a new required point; close to a minimal covering set); lower = many documents duplicating the same point with no increment. APPLICABILITY PREREQUISITE (judge this FIRST): Redundancy is judged ONLY over the RELEVANT SUBSET (Relevance >= 1), against the required key information points; off-topic documents (Relevance = 0) are IGNORED and NEVER count as ``a distinct point'' nor improve the score. If FEWER THAN TWO relevant documents actually cover a key information point (there is nothing that could possibly overlap), Redundancy is NOT APPLICABLE: output it as null --- do NOT output 10, and never reward ``nothing to overlap''.\\
\hspace*{1em}- 10: no redundancy --- every document contributes a DISTINCT required information point; the set is essentially a minimal covering set with no duplicated point.\\
\hspace*{1em}- 8: nearly no redundancy --- at most a negligible overlap; almost every document still adds a new required point.\\
\hspace*{1em}- 5: some redundancy --- a few documents repeat an already-covered information point and add no new required point.\\
\hspace*{1em}- 3: heavy redundancy --- many documents cover the same information point(s), giving little incremental coverage of the required points.\\
\hspace*{1em}- 0: severe redundancy --- most documents duplicate the same one/few information points and bring no new required point (e.g., several documents all only answer the same single point).\\
- Conflict: whether different documents in the set are FREE OF mutual contradiction on the same factual point. APPLICABILITY PREREQUISITE (judge this FIRST): Conflict is judged ONLY over the RELEVANT SUBSET (Relevance >= 1); off-topic documents (Relevance = 0) never count as a comparable statement. Conflict can only be scored when the RELEVANT SUBSET contains AT LEAST TWO comparable statements about the SAME query-relevant key-information-point factual point (i.e., two or more documents each make a checkable claim about the same fact, so that agreement or contradiction is even possible). If FEWER THAN TWO comparable statements exist about the same fact (the relevant fact is asserted by at most one document, or is not asserted at all), then there is nothing that could agree or contradict --- Conflict is NOT APPLICABLE. In that case DO NOT give a 0--10 score; output this dimension as null (see output format). Do NOT give a high score merely because ``nothing contradicts'': absence of comparable statements is null, not 10. When the prerequisite is met (>=2 comparable statements about the same fact), score 0--10 by how consistent they are. Higher score = less conflict, more consistently supporting the correct answer. (Only judge whether contradictions exist within the doc set; do not factor in relevance here.)\\
\hspace*{1em}- 10: the two-or-more statements about the same fact are fully consistent, with no contradiction.\\
\hspace*{1em}- 8: overall consistent; only a minor non-key wording difference that can be ignored.\\
\hspace*{1em}- 5: slight inconsistency exists, but it does not affect reaching the correct conclusion.\\
\hspace*{1em}- 3: clear disagreement on a key fact, potentially misleading the conclusion.\\
\hspace*{1em}- 0: a direct factual contradiction exists (e.g., two mutually exclusive values for the same fact), undermining the set's support for the correct answer.\\
\#\# III. Global-Level dimensions: give ONE overall score for the WHOLE doc set\\
- Completeness: whether the doc set as a whole covers ALL key information points the correct answer depends on, with no obvious information gap. Coverage counts ONLY from the RELEVANT SUBSET (Relevance >= 1): an off-topic document (Relevance = 0) NEVER counts as covering any key information point. (This dimension measures the information GAP, so if the relevant subset covers none of the required points, Completeness = 0 --- it is 0, not null.)\\
\hspace*{1em}- 10: fully covers all key information points the rubric lists, with no gap, sufficient to derive the correct answer.\\
\hspace*{1em}- 8: covers the vast majority of key points, missing only a minor piece; largely answerable.\\
\hspace*{1em}- 5: covers some key points but misses important information; hard to answer completely.\\
\hspace*{1em}- 3: only sporadically covers a few points; key information largely missing.\\
\hspace*{1em}- 0: covers almost no key point; impossible to answer from it.\\
- Density: judged in terms of LENGTH / span --- whether the content that is actually useful for answering the query (the text that touches the required information points) occupies the MAJORITY of each document's length, rather than the documents being padded with large amounts of irrelevant / filler text where only a small fraction of the length is useful. Higher score = the documents are compact and on-point (high useful-content-to-length ratio); lower = the documents are long but mostly filler, with only a small useful snippet. This concerns the internal signal-to-length ratio of the documents, NOT whether a document is removable. APPLICABILITY PREREQUISITE (judge this FIRST): Density is judged ONLY over the RELEVANT SUBSET (Relevance >= 1), and ``useful content'' means ONLY the text that actually carries the query-required key-information-point content. Off-topic documents (Relevance = 0) are NOT counted at all (neither their length nor as useful content). If NO relevant document carries any key-information-point content, Density is NOT APPLICABLE: output it as null (do NOT give a score just because documents look short/clean).\\
\hspace*{1em}- 10: across the documents, useful content occupies the vast majority of each document's length; the documents are compact and on-point with almost no filler.\\
\hspace*{1em}- 8: useful content occupies most of the length in most documents; only minor filler / padding.\\
\hspace*{1em}- 5: useful content and irrelevant filler are roughly balanced; a fair share of the length is not useful.\\
\hspace*{1em}- 3: in most documents only a small fraction of the length is useful; the bulk is filler or off-topic padding.\\
\hspace*{1em}- 0: the documents are almost entirely filler; only a tiny sliver of the length is useful for this query.\\
- Reachability: whether, using ONLY this doc set and NO external knowledge, the COMPLETE reasoning chain required to reach the correct answer can be completed (every link grounded in the doc set). Only content from the RELEVANT SUBSET (Relevance >= 1) may ground a link; off-topic documents (Relevance = 0) contribute NOTHING to the chain. (This dimension measures derivability, so if the relevant subset grounds no link of the required key-information-point reasoning chain, Reachability = 0 --- it is 0, not null.)\\
\hspace*{1em}- 10: every link of the reasoning chain (bridge entities, intermediate facts, final comparison, etc.) is grounded in the set; the correct answer is self-containedly derivable.\\
\hspace*{1em}- 8: the vast majority of links are supported by the set; only a minor link needs slight guessing.\\
\hspace*{1em}- 5: a key link is partially missing; some external knowledge is needed to complete the chain.\\
\hspace*{1em}- 3: the reasoning chain breaks in multiple places; many links lack document grounding.\\
\hspace*{1em}- 0: a key link is missing; it is almost impossible to derive the correct answer from the doc set alone.\\
\# Scoring example (Case)\\
Suppose the rubrics are:\\
\hspace*{1em}{[}Relevance] (Doc-Level) ``Does each document directly discuss the 1999 regular-season win-loss record of the team that won Super Bowl XXXIV (the St. Louis Rams)?''\\
\hspace*{1em}{[}Authenticity] (Doc-Level) ``Is each document's statement about that team's 1999 regular-season record consistent with the objective fact (13--3)?''\\
\hspace*{1em}{[}Quality] (Doc-Level) ``Is each document clearly structured so a reader can directly extract the 13--3 record?''\\
\hspace*{1em}{[}Complementarity] (Set-Level) ``Does the doc set, through complementary documents, jointly cover both key facets --- `the identity of the champion team' and `the 1999 regular-season 13--3 record'?''\\
\hspace*{1em}{[}Redundancy] (Set-Level) ``Does the doc set avoid multiple documents redundantly covering the same required information point (the champion identity or the 13--3 record) without adding a new required point?''\\
\hspace*{1em}{[}Conflict] (Set-Level) ``Are the statements across the doc set about the Rams' 1999 regular-season record consistent, with no mutual contradiction?''\\
\hspace*{1em}{[}Completeness] (Global-Level) ``Does the doc set fully provide the champion team and its 1999 regular-season 13--3 record, sufficient to answer the question?''\\
\hspace*{1em}{[}Density] (Global-Level) ``Across the documents, does the text useful for the champion team and its 13--3 record occupy the majority of each document's length, rather than being buried in filler?''\\
\hspace*{1em}{[}Reachability] (Global-Level) ``Using only this doc set, can the full reasoning chain `confirm the champion is the Rams $\rightarrow$ find their 1999 record 13--3' be completed?''\\
Suppose the answer requires 2 key information points: (P1) the champion is the St. Louis Rams; (P2) their 1999 regular-season record is 13--3.\\
Doc-Level dimensions (score each document):\\
- Doc A (text: ``\dots the Rams finished 1999 at 13--3 and won Super Bowl XXXIV\dots''): Relevance = 10: directly and explicitly gives the champion team's 1999 record. Authenticity = 10: 13--3 fully matches the fact. Quality = 10: the record is clear and directly extractable.\\
- Doc B (text: about the Titans ``finishing 1999 at 13--3, losing to the Rams''): Relevance = 5: discusses the same game's record, but the subject is the losing Titans rather than the champion Rams, so only indirectly relevant. Authenticity = 5: the Titans indeed went 13--3, but using it to answer ``the champion's record'' is misleading (both teams coincidentally went 13--3). Quality = 8: the record is clearly stated, but the reader must distinguish which team.\\
- Doc C (text: about a different Super Bowl XLI, some team 12--4): Relevance = 0: an entirely different edition, unrelated to this question. Authenticity = 0: makes no statement about the Rams' 1999 record. Quality = 0: unrelated to the query, contains no target information.\\
Set-Level dimensions (one overall score):\\
- Complementarity = 3: Doc A ALONE already covers both required points (P1 champion = Rams, P2 record = 13--3), so there is essentially no division of labour; Doc B only adds non-required Titans context (not a new required point) and Doc C is unrelated --- no genuine cross-document complementarity, even though coverage (Completeness) is perfect.\\
- Redundancy = 5: Doc A already covers both required points (P1 and P2); Doc B mainly repeats the 13--3 point (P2) from the Titans' angle without adding a NEW required point, so relative to the minimal covering set (Doc A alone) there is some redundancy; Doc C is unrelated and covers no required point.\\
- Conflict = null (NOT APPLICABLE): the rubric points to the RAMS' 1999 record; only Doc A actually states the Rams' record (13--3). Doc B states the TITANS' record, not the Rams', and Doc C is unrelated --- so there are FEWER THAN TWO comparable statements about the same fact (the Rams' record). Since nothing could agree or contradict, Conflict is not applicable and is output as null (NOT 10).\\
Global-Level dimensions (one overall score):\\
- Completeness = 10: Doc A already gives ``the Rams won + 1999 record 13--3'' (both P1 and P2); the whole set has no information gap.\\
- Density = 8: judged ONLY over the relevant subset (Doc A and Doc B; Doc C has Relevance = 0 and is EXCLUDED --- it does NOT count as filler here, since Density ignores off-topic documents). Doc A is compact and on-point; Doc B carries some extra non-essential Titans narrative, so across the RELEVANT documents the useful key-information-point text still occupies most of the length.\\
- Reachability = 10: Doc A alone completes the full reasoning chain ``champion = Rams $\rightarrow$ record = 13--3'', with no external knowledge needed.\\
\# Output format\\
Output ONLY a single valid JSON object (no markdown code fences, no extra commentary). The object has exactly ONE top-level field:\\
1. ``score'': the atomic scores per dimension. Doc-level dimensions use per\_doc (one entry per document, with only doc\_id and score); set-level / global-level dimensions use set\_level (one overall score). You do NOT need to compute rubric\_avg or dimension\_score (those are computed by downstream code). Do NOT output any reasoning, explanation, or ``reason'' field --- output the ``score'' object only.\\
Structure:\\
{\ttfamily
\{\\
\hspace*{1em}"score": \{\\
\hspace*{2em}"Relevance": \{\\
\hspace*{3em}"level": "doc",\\
\hspace*{3em}"rubrics": [\\
\hspace*{4em}\{ "rubric": "<the original rubric text>",\\
\hspace*{5em}"per\_doc": [ \{"doc\_id": 9, "score": 10\}, \{"doc\_id": 18, "score": 5\} ] \}\\
\hspace*{3em}]\\
\hspace*{2em}\},\\
\hspace*{2em}"Authenticity": \{ "level": "doc", "rubrics": [ ... ] \},\\
\hspace*{2em}"Quality": \{ "level": "doc", "rubrics": [ ... ] \},\\
\hspace*{2em}"Complementarity": \{\\
\hspace*{3em}"level": "set",\\
\hspace*{3em}"rubrics": [ \{ "rubric": "<the original rubric text>", "set\_level": \{"score": 8\} \} ]\\
\hspace*{2em}\},\\
\hspace*{2em}"Redundancy": \{ "level": "set", "rubrics": [ ... ] \},\\
\hspace*{2em}"Conflict": \{ "level": "set", "rubrics": [ ... ] \},\\
\hspace*{2em}"Completeness": \{ "level": "set", "rubrics": [ ... ] \},\\
\hspace*{2em}"Density": \{ "level": "set", "rubrics": [ ... ] \},\\
\hspace*{2em}"Reachability": \{ "level": "set", "rubrics": [ ... ] \}\\
\hspace*{1em}\}\\
\}
}\\
If CASE A applies (ALL documents have Relevance = 0), output the 8 non-Relevance dimensions as null, like: ``Authenticity'': null, ``Quality'': null, ``Complementarity'': null, ``Redundancy'': null, ``Conflict'': null, ``Completeness'': null, ``Density'': null, ``Reachability'': null. (Only ``Relevance'' is fully scored in CASE A.)\\
Not-applicable dimensions (independent of CASE A): even in CASE B, a relevance-grounded set/global dimension whose APPLICABILITY PREREQUISITE is not met over the RELEVANT SUBSET must be output as a bare JSON null for the WHOLE dimension (NOT a high score), while the other dimensions are still scored:\\
\hspace*{1em}- ``Conflict'': null when fewer than two RELEVANT documents make comparable statements about the same query-relevant key-information-point fact.\\
\hspace*{1em}- ``Redundancy'': null when fewer than two RELEVANT documents cover a key information point (nothing could possibly overlap).\\
\hspace*{1em}- ``Complementarity'': null when fewer than two RELEVANT documents each contribute a DIFFERENT key information point (no possible division of labour).\\
\hspace*{1em}- ``Density'': null when NO relevant document carries any key-information-point content.\\
(Completeness and Reachability are NEVER null in CASE B --- they measure information GAP / derivability, so when the relevant subset covers nothing they are 0, not null.)\\
Rules:\\
- Output exactly ONE top-level field: ``score''. Do NOT output any ``reason'', per-item evidence, or per-item reason fields.\\
- Only output dimensions actually provided in the input; under each dimension, the number and order of rubrics must match the input.\\
- Doc-level dimensions: per\_doc must cover EVERY document in the doc set. CRITICAL: doc\_id MUST be exactly the id shown in square brackets before each document in the ``Doc set'' below (that is the document's original id, e.g. {[}9] $\rightarrow$ doc\_id 9). Do NOT renumber the documents as 1,2,3,\dots; use their bracketed original ids.\\
- set-level / global-level dimensions: give only one overall set\_level score, not per-document.\\
- RELEVANCE GATE: always score Relevance. If ALL documents have Relevance = 0 (CASE A), output the other 8 dimensions as null. Otherwise (CASE B) evaluate the other dimensions ONLY over the RELEVANT SUBSET (Relevance >= 1), and output as null any relevance-grounded dimension whose applicability prerequisite is not met.\\
- RELEVANCE-GROUNDED dimensions: Complementarity, Redundancy, Conflict, Density --- and the coverage / derivability judged by Completeness / Reachability --- are assessed ONLY over the RELEVANT SUBSET (documents with Relevance >= 1) and ONLY on the query-required key information points. Off-topic documents (Relevance = 0) are EXCLUDED: they never count as coverage, a distinct point, useful length, or a comparable statement, and they NEVER raise these scores. When a relevance-grounded SET-LEVEL dimension has too few relevant documents to be meaningful (see each prerequisite: Complementarity / Redundancy / Conflict / Density), output it as null --- never reward it just because off-topic documents trivially ``don't conflict / don't overlap / don't dilute''.\\
- Conflict is CONDITIONAL: only score it (0--10) when at least two documents make comparable statements about the SAME query-relevant fact; if there are fewer than two such comparable statements, output ``Conflict'': null (do NOT output 10). Never treat ``nothing to contradict'' as a high Conflict score.\\
- Complementarity vs Completeness are DIFFERENT: Complementarity judges cross-document DIVISION OF LABOUR (if a single document covers everything by itself, Complementarity is LOW even when nothing is missing), and must NOT be lowered just because some point is missing; Completeness judges only whether any required information point is MISSING (an information gap), regardless of how many documents supply it. Do not conflate the two.\\
- score can only be an integer 0--10 (or null for a whole dimension under CASE A, or when a relevance-grounded dimension's applicability prerequisite is not met over the RELEVANT SUBSET --- i.e. Conflict / Redundancy / Complementarity / Density); no N/A or empty per-item values allowed.
}
\userprompt{
User query:\\
\prompttag{\textbf{\color{insert}<QUERY>}}\\
Doc set:\\
\prompttag{\textbf{\color{insert}<DOCS>}}\\
Rubrics (each labeled with its type and content):\\
\prompttag{\textbf{\color{insert}<RUBRICS>}}\\
Now begin outputting the result (JSON with only the ``score'' object):
}}%
\prompt
{Rubric-based Judge Scoring Prompt: Long-form Scenario}{
\systemprompt{
You are a senior search-result quality assessor. Given a user query, a candidate document set (doc set, variable size) and a list of rubrics (each labeled with its dimension type and judgment content), you must score every rubric on a 0--10 scale, strictly grounded in the document content.\\
The query intent may clarify why this query was issued and what information it is trying to find, but it may also contain broader, exploratory, or irrelevant thoughts. Stay anchored on the query itself and only use the parts of the query intent that are actually relevant.\\
This round evaluates all 9 dimensions at once, grouped into three categories:\\
{[}A. Doc-Level dimensions] --- judged for EACH individual document in the doc set: Relevance, Authenticity, Quality.\\
{[}B. Set-Level dimensions] --- judged for the ENTIRE doc set as a whole; give ONE overall score, not per-document: Complementarity, Redundancy, Conflict.\\
{[}C. Global-Level dimensions] --- also judged for the ENTIRE doc set as a whole; give ONE overall score, not per-document: Completeness, Density, Reachability.\\
\# Absolute prerequisite: you MUST read the documents first\\
- Every score must be grounded in the actual document text. Judge only based on what ACTUALLY appears in the documents; do not use your own external knowledge to fill in information the documents do not state.\\
- The rubric text is itself query-specific (it already contains the concrete entities, facts, and numbers of the question); the user query is also provided above for context. Judge directly against the rubric, grounded in the documents.\\
\# Evaluation procedure --- RELEVANCE GATE (read carefully, follow the order)\\
Think step by step in this order:\\
1. FIRST evaluate the Relevance dimension for EVERY document in the doc set (give each document its per\_doc Relevance score).\\
2. Define the RELEVANT SUBSET = the documents whose Relevance score >= 1 (i.e. documents that actually touch the query's key information point(s)). Documents with Relevance = 0 are OFF-TOPIC and are EXCLUDED from EVERY other dimension: they NEVER count as providing coverage, as a distinct information point, as useful length, or as a comparable statement, and they NEVER count toward ``no redundancy'' or ``no conflict''.\\
3. Then check the relevance outcome:\\
\hspace*{1em}- CASE A --- the RELEVANT SUBSET is EMPTY (every document's Relevance score = 0): the documents this ranker selected are all useless for this question, so the other 8 dimensions (Authenticity, Quality, Complementarity, Redundancy, Conflict, Completeness, Density, Reachability) carry NO evaluation meaning. In this case DO NOT score them --- output each of those 8 dimensions as null (see output format). Only the Relevance dimension is scored.\\
\hspace*{1em}- CASE B --- the RELEVANT SUBSET is non-empty (at least ONE document has Relevance >= 1): evaluate the other 8 dimensions ONLY over the RELEVANT SUBSET, and apply each dimension's own APPLICABILITY PREREQUISITE below. A relevance-grounded dimension whose prerequisite is not met over the relevant subset (e.g. it needs a comparison / division of labour among relevant documents but there are not enough relevant documents, or no relevant document actually carries the key-information-point content it measures) MUST be output as null --- NOT as a high score. Never reward a dimension merely because off-topic documents trivially ``don't conflict / don't overlap / don't dilute''.\\
4. Never leave Relevance itself null --- Relevance is always scored.\\
\# Scoring scale (0--10)\\
Overall meaning (higher = better satisfies the rubric). Use the following anchors, and you MAY use intermediate integers (0--10) for in-between cases:\\
- 10 = Completely: fully satisfied, with sufficient detail and evidence.\\
- 8 = Mostly: largely satisfied, but missing some detail or support.\\
- 5 = Moderately: relevant content is mentioned, but key details are missing.\\
- 3 = Barely: not explicitly stated, only weakly inferable.\\
- 0 = Not at all: not addressed at all / irrelevant to the rubric / contradicts the facts.\\
\#\# I. Doc-Level dimensions: score EACH document individually\\
- Relevance: to what extent the document DIRECTLY discusses the specific information facet the rubric points to.\\
\hspace*{1em}- 10: the document's core content directly and fully discusses this specific question and can directly support the answer.\\
\hspace*{1em}- 8: directly discusses the topic, but the information is incomplete or unfocused.\\
\hspace*{1em}- 5: mentions the related topic only in passing, missing key information.\\
\hspace*{1em}- 3: barely discusses it directly, relevance is only weakly inferable.\\
\hspace*{1em}- 0: entirely off-topic, merely matching keywords or unrelated to the question.\\
- Authenticity: how well the facts stated in the document CONFORM to the reference fact given by the rubric. Only assess whether the STATEMENTS that appear are accurate; independent of relevance.\\
\hspace*{1em}- 10: the document explicitly states the relevant fact, fully consistent with the reference fact.\\
\hspace*{1em}- 8: the statement is consistent with the fact, but phrased unclearly or somewhat vaguely.\\
\hspace*{1em}- 5: partially correct, or the fact is right but wrapped in easily-misleading phrasing.\\
\hspace*{1em}- 3: only faintly touched upon; correctness is hard to judge or slightly off.\\
\hspace*{1em}- 0: the statement contradicts the reference fact, has an obvious factual error, or the document makes no statement related to that fact at all.\\
- Quality: how easy the document makes it for a reader to EXTRACT the target information, in terms of information organization and presentation. Assess presentation quality only, not factual correctness.\\
\hspace*{1em}- 10: clearly structured and focused; the target information is immediately apparent.\\
\hspace*{1em}- 8: mostly clear; the target information is extractable but mixed with some irrelevant content.\\
\hspace*{1em}- 5: the target information exists but is diluted by lots of irrelevant content and takes effort to find.\\
\hspace*{1em}- 3: messy layout, high noise; the target information is very hard to extract.\\
\hspace*{1em}- 0: empty, extremely noisy, or nearly unparseable content, OR the document is entirely off-topic and contains no target information at all.\\
\#\# II. Set-Level dimensions: give ONE overall score for the WHOLE doc set\\
- Complementarity: whether the key information points are covered through DIVISION OF LABOUR across DIFFERENT documents --- i.e., different documents each contribute a DIFFERENT needed piece (one gives an overview, another the details; one supplies point P1, another supplies point P2), so that the whole is more valuable than any single document. APPLICABILITY PREREQUISITE (judge this FIRST): Complementarity is judged ONLY over the RELEVANT SUBSET (Relevance >= 1) and ONLY on the query-required key information points; off-topic documents (Relevance = 0) NEVER count as a contributing part. If FEWER THAN TWO relevant documents each contribute a DIFFERENT required point (there is at most one relevant document, or all relevant documents only carry the SAME single point), there is no possible cross-document division of labour on the key information points --- Complementarity is NOT APPLICABLE: output it as null (NOT a high score). IMPORTANT: this dimension judges only the complementary STRUCTURE among the RELEVANT documents; it does NOT judge whether the set is complete. Do NOT lower the score merely because some required point is missing (that is Completeness's job) --- an incomplete set can still be highly complementary if the points it DOES cover come from different documents. Conversely, if a SINGLE document already covers everything on its own, complementarity is LOW even though coverage may be perfect (there is no division of labour).\\
\hspace*{1em}- 10: strong complementarity --- multiple documents clearly divide the labour, each contributing a DIFFERENT needed point, so the whole is far more valuable than any single document.\\
\hspace*{1em}- 8: mostly complementary --- most of the covered points are supplied by different documents; only an occasional point is carried alone by one document.\\
\hspace*{1em}- 5: partial complementarity --- there is some division of labour, but a large share of the covered content is concentrated in a single document.\\
\hspace*{1em}- 3: very weak complementarity --- almost all useful content comes from one document, or several documents merely restate the same single point.\\
\hspace*{1em}- 0: no complementarity at all --- only one document contributes any useful content (the others add nothing), OR a single document already covers everything by itself so no cross-document complementarity exists.\\
- Redundancy: judged against the MINIMAL document subset needed to cover the key information points the answer requires (the SAME key information points as Completeness). It checks whether each document brings INCREMENTAL coverage of those required information points, or merely repeats a point already covered by another document. If two or more documents cover the SAME required information point while adding no new required point, that overlap is redundancy --- those documents could be dropped from the minimal covering set without losing any required point. (Example: if answering the query needs 2 information points, and both doc {[}1] and doc {[}5] only answer the SAME single point, that is redundancy.) Higher score = less redundancy (each document adds a new required point; close to a minimal covering set); lower = many documents duplicating the same point with no increment. APPLICABILITY PREREQUISITE (judge this FIRST): Redundancy is judged ONLY over the RELEVANT SUBSET (Relevance >= 1), against the required key information points; off-topic documents (Relevance = 0) are IGNORED and NEVER count as ``a distinct point'' nor improve the score. If FEWER THAN TWO relevant documents actually cover a key information point (there is nothing that could possibly overlap), Redundancy is NOT APPLICABLE: output it as null --- do NOT output 10, and never reward ``nothing to overlap''.\\
\hspace*{1em}- 10: no redundancy --- every document contributes a DISTINCT required information point; the set is essentially a minimal covering set with no duplicated point.\\
\hspace*{1em}- 8: nearly no redundancy --- at most a negligible overlap; almost every document still adds a new required point.\\
\hspace*{1em}- 5: some redundancy --- a few documents repeat an already-covered information point and add no new required point.\\
\hspace*{1em}- 3: heavy redundancy --- many documents cover the same information point(s), giving little incremental coverage of the required points.\\
\hspace*{1em}- 0: severe redundancy --- most documents duplicate the same one/few information points and bring no new required point (e.g., several documents all only answer the same single point).\\
- Conflict: whether different documents in the set are FREE OF mutual contradiction on the same factual point. APPLICABILITY PREREQUISITE (judge this FIRST): Conflict is judged ONLY over the RELEVANT SUBSET (Relevance >= 1); off-topic documents (Relevance = 0) never count as a comparable statement. Conflict can only be scored when the RELEVANT SUBSET contains AT LEAST TWO comparable statements about the SAME query-relevant key-information-point factual point (i.e., two or more documents each make a checkable claim about the same fact, so that agreement or contradiction is even possible). If FEWER THAN TWO comparable statements exist about the same fact (the relevant fact is asserted by at most one document, or is not asserted at all), then there is nothing that could agree or contradict --- Conflict is NOT APPLICABLE. In that case DO NOT give a 0--10 score; output this dimension as null (see output format). Do NOT give a high score merely because ``nothing contradicts'': absence of comparable statements is null, not 10. When the prerequisite is met (>=2 comparable statements about the same fact), score 0--10 by how consistent they are. Higher score = less conflict, more consistently supporting the correct answer. (Only judge whether contradictions exist within the doc set; do not factor in relevance here.)\\
\hspace*{1em}- 10: the two-or-more statements about the same fact are fully consistent, with no contradiction.\\
\hspace*{1em}- 8: overall consistent; only a minor non-key wording difference that can be ignored.\\
\hspace*{1em}- 5: slight inconsistency exists, but it does not affect reaching the correct conclusion.\\
\hspace*{1em}- 3: clear disagreement on a key fact, potentially misleading the conclusion.\\
\hspace*{1em}- 0: a direct factual contradiction exists (e.g., two mutually exclusive values for the same fact), undermining the set's support for the correct answer.\\
\#\# III. Global-Level dimensions: give ONE overall score for the WHOLE doc set\\
- Completeness: whether the doc set as a whole covers ALL key information points the correct answer depends on, with no obvious information gap. Coverage counts ONLY from the RELEVANT SUBSET (Relevance >= 1): an off-topic document (Relevance = 0) NEVER counts as covering any key information point. (This dimension measures the information GAP, so if the relevant subset covers none of the required points, Completeness = 0 --- it is 0, not null.)\\
\hspace*{1em}- 10: fully covers all key information points the rubric lists, with no gap, sufficient to derive the correct answer.\\
\hspace*{1em}- 8: covers the vast majority of key points, missing only a minor piece; largely answerable.\\
\hspace*{1em}- 5: covers some key points but misses important information; hard to answer completely.\\
\hspace*{1em}- 3: only sporadically covers a few points; key information largely missing.\\
\hspace*{1em}- 0: covers almost no key point; impossible to answer from it.\\
- Density: judged in terms of LENGTH / span --- whether the content that is actually useful for answering the query (the text that touches the required information points) occupies the MAJORITY of each document's length, rather than the documents being padded with large amounts of irrelevant / filler text where only a small fraction of the length is useful. Higher score = the documents are compact and on-point (high useful-content-to-length ratio); lower = the documents are long but mostly filler, with only a small useful snippet. This concerns the internal signal-to-length ratio of the documents, NOT whether a document is removable. APPLICABILITY PREREQUISITE (judge this FIRST): Density is judged ONLY over the RELEVANT SUBSET (Relevance >= 1), and ``useful content'' means ONLY the text that actually carries the query-required key-information-point content. Off-topic documents (Relevance = 0) are NOT counted at all (neither their length nor as useful content). If NO relevant document carries any key-information-point content, Density is NOT APPLICABLE: output it as null (do NOT give a score just because documents look short/clean).\\
\hspace*{1em}- 10: across the documents, useful content occupies the vast majority of each document's length; the documents are compact and on-point with almost no filler.\\
\hspace*{1em}- 8: useful content occupies most of the length in most documents; only minor filler / padding.\\
\hspace*{1em}- 5: useful content and irrelevant filler are roughly balanced; a fair share of the length is not useful.\\
\hspace*{1em}- 3: in most documents only a small fraction of the length is useful; the bulk is filler or off-topic padding.\\
\hspace*{1em}- 0: the documents are almost entirely filler; only a tiny sliver of the length is useful for this query.\\
- Reachability: whether, using ONLY this doc set and NO external knowledge, the COMPLETE reasoning chain required to reach the correct answer can be completed (every link grounded in the doc set). Only content from the RELEVANT SUBSET (Relevance >= 1) may ground a link; off-topic documents (Relevance = 0) contribute NOTHING to the chain. (This dimension measures derivability, so if the relevant subset grounds no link of the required key-information-point reasoning chain, Reachability = 0 --- it is 0, not null.)\\
\hspace*{1em}- 10: every link of the reasoning chain (bridge entities, intermediate facts, final comparison, etc.) is grounded in the set; the correct answer is self-containedly derivable.\\
\hspace*{1em}- 8: the vast majority of links are supported by the set; only a minor link needs slight guessing.\\
\hspace*{1em}- 5: a key link is partially missing; some external knowledge is needed to complete the chain.\\
\hspace*{1em}- 3: the reasoning chain breaks in multiple places; many links lack document grounding.\\
\hspace*{1em}- 0: a key link is missing; it is almost impossible to derive the correct answer from the doc set alone.\\
\# Scoring example (Case)\\
Suppose the rubrics are:\\
\hspace*{1em}{[}Relevance] (Doc-Level) ``Does each document directly discuss the 1999 regular-season win-loss record of the team that won Super Bowl XXXIV (the St. Louis Rams)?''\\
\hspace*{1em}{[}Authenticity] (Doc-Level) ``Is each document's statement about that team's 1999 regular-season record consistent with the objective fact (13--3)?''\\
\hspace*{1em}{[}Quality] (Doc-Level) ``Is each document clearly structured so a reader can directly extract the 13--3 record?''\\
\hspace*{1em}{[}Complementarity] (Set-Level) ``Does the doc set, through complementary documents, jointly cover both key facets --- `the identity of the champion team' and `the 1999 regular-season 13--3 record'?''\\
\hspace*{1em}{[}Redundancy] (Set-Level) ``Does the doc set avoid multiple documents redundantly covering the same required information point (the champion identity or the 13--3 record) without adding a new required point?''\\
\hspace*{1em}{[}Conflict] (Set-Level) ``Are the statements across the doc set about the Rams' 1999 regular-season record consistent, with no mutual contradiction?''\\
\hspace*{1em}{[}Completeness] (Global-Level) ``Does the doc set fully provide the champion team and its 1999 regular-season 13--3 record, sufficient to answer the question?''\\
\hspace*{1em}{[}Density] (Global-Level) ``Across the documents, does the text useful for the champion team and its 13--3 record occupy the majority of each document's length, rather than being buried in filler?''\\
\hspace*{1em}{[}Reachability] (Global-Level) ``Using only this doc set, can the full reasoning chain `confirm the champion is the Rams $\rightarrow$ find their 1999 record 13--3' be completed?''\\
Suppose the answer requires 2 key information points: (P1) the champion is the St. Louis Rams; (P2) their 1999 regular-season record is 13--3.\\
Doc-Level dimensions (score each document):\\
- Doc A (text: ``\dots the Rams finished 1999 at 13--3 and won Super Bowl XXXIV\dots''): Relevance = 10: directly and explicitly gives the champion team's 1999 record. Authenticity = 10: 13--3 fully matches the fact. Quality = 10: the record is clear and directly extractable.\\
- Doc B (text: about the Titans ``finishing 1999 at 13--3, losing to the Rams''): Relevance = 5: discusses the same game's record, but the subject is the losing Titans rather than the champion Rams, so only indirectly relevant. Authenticity = 5: the Titans indeed went 13--3, but using it to answer ``the champion's record'' is misleading (both teams coincidentally went 13--3). Quality = 8: the record is clearly stated, but the reader must distinguish which team.\\
- Doc C (text: about a different Super Bowl XLI, some team 12--4): Relevance = 0: an entirely different edition, unrelated to this question. Authenticity = 0: makes no statement about the Rams' 1999 record. Quality = 0: unrelated to the query, contains no target information.\\
Set-Level dimensions (one overall score):\\
- Complementarity = 3: Doc A ALONE already covers both required points (P1 champion = Rams, P2 record = 13--3), so there is essentially no division of labour; Doc B only adds non-required Titans context (not a new required point) and Doc C is unrelated --- no genuine cross-document complementarity, even though coverage (Completeness) is perfect.\\
- Redundancy = 5: Doc A already covers both required points (P1 and P2); Doc B mainly repeats the 13--3 point (P2) from the Titans' angle without adding a NEW required point, so relative to the minimal covering set (Doc A alone) there is some redundancy; Doc C is unrelated and covers no required point.\\
- Conflict = null (NOT APPLICABLE): the rubric points to the RAMS' 1999 record; only Doc A actually states the Rams' record (13--3). Doc B states the TITANS' record, not the Rams', and Doc C is unrelated --- so there are FEWER THAN TWO comparable statements about the same fact (the Rams' record). Since nothing could agree or contradict, Conflict is not applicable and is output as null (NOT 10).\\
Global-Level dimensions (one overall score):\\
- Completeness = 10: Doc A already gives ``the Rams won + 1999 record 13--3'' (both P1 and P2); the whole set has no information gap.\\
- Density = 8: judged ONLY over the relevant subset (Doc A and Doc B; Doc C has Relevance = 0 and is EXCLUDED --- it does NOT count as filler here, since Density ignores off-topic documents). Doc A is compact and on-point; Doc B carries some extra non-essential Titans narrative, so across the RELEVANT documents the useful key-information-point text still occupies most of the length.\\
- Reachability = 10: Doc A alone completes the full reasoning chain ``champion = Rams $\rightarrow$ record = 13--3'', with no external knowledge needed.\\
\# Output format\\
Output ONLY a single valid JSON object (no markdown code fences, no extra commentary). The object has exactly ONE top-level field:\\
1. ``score'': the atomic scores per dimension. Doc-level dimensions use per\_doc (one entry per document, with only doc\_id and score); set-level / global-level dimensions use set\_level (one overall score). You do NOT need to compute rubric\_avg or dimension\_score (those are computed by downstream code). Do NOT output any reasoning, explanation, or ``reason'' field --- output the ``score'' object only.\\
Structure:\\
{\ttfamily
\{\\
\hspace*{1em}"score": \{\\
\hspace*{2em}"Relevance": \{\\
\hspace*{3em}"level": "doc",\\
\hspace*{3em}"rubrics": [\\
\hspace*{4em}\{ "rubric": "<the original rubric text>",\\
\hspace*{5em}"per\_doc": [ \{"doc\_id": 9, "score": 10\}, \{"doc\_id": 18, "score": 5\} ] \}\\
\hspace*{3em}]\\
\hspace*{2em}\},\\
\hspace*{2em}"Authenticity": \{ "level": "doc", "rubrics": [ ... ] \},\\
\hspace*{2em}"Quality": \{ "level": "doc", "rubrics": [ ... ] \},\\
\hspace*{2em}"Complementarity": \{\\
\hspace*{3em}"level": "set",\\
\hspace*{3em}"rubrics": [ \{ "rubric": "<the original rubric text>", "set\_level": \{"score": 8\} \} ]\\
\hspace*{2em}\},\\
\hspace*{2em}"Redundancy": \{ "level": "set", "rubrics": [ ... ] \},\\
\hspace*{2em}"Conflict": \{ "level": "set", "rubrics": [ ... ] \},\\
\hspace*{2em}"Completeness": \{ "level": "set", "rubrics": [ ... ] \},\\
\hspace*{2em}"Density": \{ "level": "set", "rubrics": [ ... ] \},\\
\hspace*{2em}"Reachability": \{ "level": "set", "rubrics": [ ... ] \}\\
\hspace*{1em}\}\\
\}
}\\
If CASE A applies (ALL documents have Relevance = 0), output the 8 non-Relevance dimensions as null, like: ``Authenticity'': null, ``Quality'': null, ``Complementarity'': null, ``Redundancy'': null, ``Conflict'': null, ``Completeness'': null, ``Density'': null, ``Reachability'': null. (Only ``Relevance'' is fully scored in CASE A.)\\
Not-applicable dimensions (independent of CASE A): even in CASE B, a relevance-grounded set/global dimension whose APPLICABILITY PREREQUISITE is not met over the RELEVANT SUBSET must be output as a bare JSON null for the WHOLE dimension (NOT a high score), while the other dimensions are still scored:\\
\hspace*{1em}- ``Conflict'': null when fewer than two RELEVANT documents make comparable statements about the same query-relevant key-information-point fact.\\
\hspace*{1em}- ``Redundancy'': null when fewer than two RELEVANT documents cover a key information point (nothing could possibly overlap).\\
\hspace*{1em}- ``Complementarity'': null when fewer than two RELEVANT documents each contribute a DIFFERENT key information point (no possible division of labour).\\
\hspace*{1em}- ``Density'': null when NO relevant document carries any key-information-point content.\\
(Completeness and Reachability are NEVER null in CASE B --- they measure information GAP / derivability, so when the relevant subset covers nothing they are 0, not null.)\\
Rules:\\
- Output exactly ONE top-level field: ``score''. Do NOT output any ``reason'', per-item evidence, or per-item reason fields.\\
- Only output dimensions actually provided in the input; under each dimension, the number and order of rubrics must match the input.\\
- Doc-level dimensions: per\_doc must cover EVERY document in the doc set. CRITICAL: doc\_id MUST be exactly the id shown in square brackets before each document in the ``Doc set'' below (that is the document's original id, e.g. {[}9] $\rightarrow$ doc\_id 9). Do NOT renumber the documents as 1,2,3,\dots; use their bracketed original ids.\\
- set-level / global-level dimensions: give only one overall set\_level score, not per-document.\\
- RELEVANCE GATE: always score Relevance. If ALL documents have Relevance = 0 (CASE A), output the other 8 dimensions as null. Otherwise (CASE B) evaluate the other dimensions ONLY over the RELEVANT SUBSET (Relevance >= 1), and output as null any relevance-grounded dimension whose applicability prerequisite is not met.\\
- RELEVANCE-GROUNDED dimensions: Complementarity, Redundancy, Conflict, Density --- and the coverage / derivability judged by Completeness / Reachability --- are assessed ONLY over the RELEVANT SUBSET (documents with Relevance >= 1) and ONLY on the query-required key information points. Off-topic documents (Relevance = 0) are EXCLUDED: they never count as coverage, a distinct point, useful length, or a comparable statement, and they NEVER raise these scores. When a relevance-grounded SET-LEVEL dimension has too few relevant documents to be meaningful (see each prerequisite: Complementarity / Redundancy / Conflict / Density), output it as null --- never reward it just because off-topic documents trivially ``don't conflict / don't overlap / don't dilute''.\\
- Conflict is CONDITIONAL: only score it (0--10) when at least two documents make comparable statements about the SAME query-relevant fact; if there are fewer than two such comparable statements, output ``Conflict'': null (do NOT output 10). Never treat ``nothing to contradict'' as a high Conflict score.\\
- Complementarity vs Completeness are DIFFERENT: Complementarity judges cross-document DIVISION OF LABOUR (if a single document covers everything by itself, Complementarity is LOW even when nothing is missing), and must NOT be lowered just because some point is missing; Completeness judges only whether any required information point is MISSING (an information gap), regardless of how many documents supply it. Do not conflate the two.\\
- score can only be an integer 0--10 (or null for a whole dimension under CASE A, or when a relevance-grounded dimension's applicability prerequisite is not met over the RELEVANT SUBSET --- i.e. Conflict / Redundancy / Complementarity / Density); no N/A or empty per-item values allowed.
}
\userprompt{
User query:\\
\prompttag{\textbf{\color{insert}<QUERY>}}\\
Query intent:\\
\prompttag{\textbf{\color{insert}<QUERY\_INTENT>}}\\
Doc set:\\
\prompttag{\textbf{\color{insert}<DOCS>}}\\
Rubrics (each labeled with its type and content):\\
\prompttag{\textbf{\color{insert}<RUBRICS>}}\\
Now begin outputting the result (JSON with only the ``score'' object):
}}%

\end{document}